\documentclass[runningheads]{llncs}

\usepackage{eccv}

\usepackage{eccvabbrv}

\usepackage{graphicx}
\usepackage{booktabs}
\usepackage{sidecap}

\usepackage[accsupp]{axessibility}  

\usepackage[pagebackref,breaklinks,colorlinks]{hyperref}

\usepackage{orcidlink}
\usepackage{url}
\usepackage{graphicx}
\usepackage{multicol}
\usepackage{multirow}
\usepackage{colortbl}

\definecolor{purple}{rgb}{0.56,0.27,0.68}
\definecolor{red}{rgb}{0.95,0.4,0.4}
\definecolor{purered}{rgb}{1,0,0}
\definecolor{teal}{RGB}{0, 150, 136}
\definecolor{blue}{rgb}{0.4,0.4,0.95}
\definecolor{darkblue}{rgb}{0,0,0.8}
\definecolor{grey}{rgb}{0.6,0.6,0.6}
\definecolor{col1}{RGB}{232, 161, 148}
\definecolor{col11}{RGB}{255, 228, 228}
\definecolor{col2}{RGB}{148, 187, 232}
\definecolor{col33}{RGB}{206, 239, 255}
\definecolor{col3}{RGB}{233, 255, 245}
\definecolor{lightgrey}{rgb}{0.85,0.85,0.85}
\definecolor{lightlightgrey}{rgb}{0.9,0.9,0.9}
\definecolor{verylightBG}{rgb}{0.9,0.99,0.99}
\definecolor{darkgreen}{rgb}{0., 0.85, 0.5}
\definecolor{lightgray}{gray}{0.75}

\definecolor{gtred}{RGB}{204, 0, 0}
\definecolor{predgreen}{RGB}{31, 237, 31}
\definecolor{figGreen}{RGB}{56, 118, 29}
\definecolor{lightgray}{gray}{0.75}

\definecolor{TableBlue}{rgb}{0.17,0.49,0.75}
\definecolor{TableRed}{rgb}{0.80,0.10,0.10}

\newcommand{\tabblueabs}[1]{$_{\color{TableBlue}\downarrow #1}$}
\newcommand{\tabredabs}[1]{$_{\color{TableRed}\uparrow #1}$}
\newcommand{\figtwosidecaption}[2]{%
    \rotatebox[origin=c]{90}{\shortstack[c]{\scriptsize\textbf{#1}\\\scriptsize\textbf{#2}}}%
}

\definecolor{TableBlue}{rgb}{0.17,0.49,0.75}
\newcommand{\tabblue}[1]{$_{\color{TableBlue}\downarrow #1}$}

\usepackage{algorithm2e}
\usepackage{pythonhighlight} 
\usepackage{fontawesome5}
\lstnewenvironment{pythonic}[1][]{\lstset{style=mypython, frame=none, #1}}{}

\usepackage[capitalize]{cleveref}
\usepackage{float}
\crefname{section}{Sec.}{Secs.}
\Crefname{section}{Section}{Sections}
\Crefname{table}{Table}{Tables}
\crefname{table}{Tab.}{Tabs.}

\newcommand{\method}{UniFlow}

\begin{document}

\title{\method: Zero-Shot LiDAR Scene Flow for Autonomous Driving} 

\titlerunning{\method: Zero-Shot LiDAR Scene Flow for Autonomous Driving}

\author{Siyi Li$^1$ \and Qingwen Zhang$^2$ \and Ishan Khatri$^{3,4}$ \and Kyle Vedder$^1$ \and \\ Eric Eaton$^1$ \and Deva Ramanan$^4$ \and Neehar Peri$^4$}

\authorrunning{Li et al.}

\institute{University of Pennsylvania \and KTH Royal Institute of Technology \and StackAV \and Carnegie Mellon University
}

\maketitle

\begin{abstract}
LiDAR scene flow is the task of estimating per-point 3D motion between consecutive point clouds. Recent methods achieve centimeter-level accuracy on popular autonomous vehicle (AV) datasets, but are typically only trained and evaluated on a single sensor. In this paper, we aim to learn general motion priors that transfer to diverse and unseen LiDAR sensors. However, prior work in LiDAR semantic segmentation and 3D object detection demonstrate that naively training on multiple datasets yields worse performance than single dataset models. Interestingly, we find that this conventional wisdom does {\em not} hold for motion estimation, and that state-of-the-art scene flow methods greatly benefit from cross-dataset training {\em without architectural modification}. We posit that low-level tasks such as motion estimation may be less sensitive to sensor configuration; indeed, our analysis shows that models trained on fast-moving objects (e.g., from highway datasets) perform well on fast-moving objects, even across different datasets. Informed by our analysis, we propose \method, a feedforward model that unifies and trains on multiple large-scale LiDAR scene flow datasets with diverse sensor placements and point cloud densities. Our frustratingly simple solution establishes a new state-of-the-art on Waymo and nuScenes, improving over prior work by 5.1\% and 35.2\% respectively. Moreover, \method \ achieves state-of-the-art accuracy on {\em unseen datasets} like TruckScenes and AEVAScenes, outperforming prior dataset-specific models by 30.1\% and 22.5\% respectively. See our \href{https://lisiyi777.github.io/UniFlow/}{project page} for additional visuals.
\end{abstract}
\vspace{-4em}

\begin{figure*}[t]
\centering
\setlength{\tabcolsep}{2pt}
\setlength{\fboxsep}{0pt}
\setlength{\fboxrule}{0.2pt}
\begin{tabular}{@{}c@{\hspace{0.6em}}c@{}}
{\figtwosidecaption{RGB}{}} &
\begin{minipage}{0.95\linewidth}\centering
  \begin{tabular}{@{}c c c c@{}}
    \fbox{\includegraphics[width=0.23\linewidth]{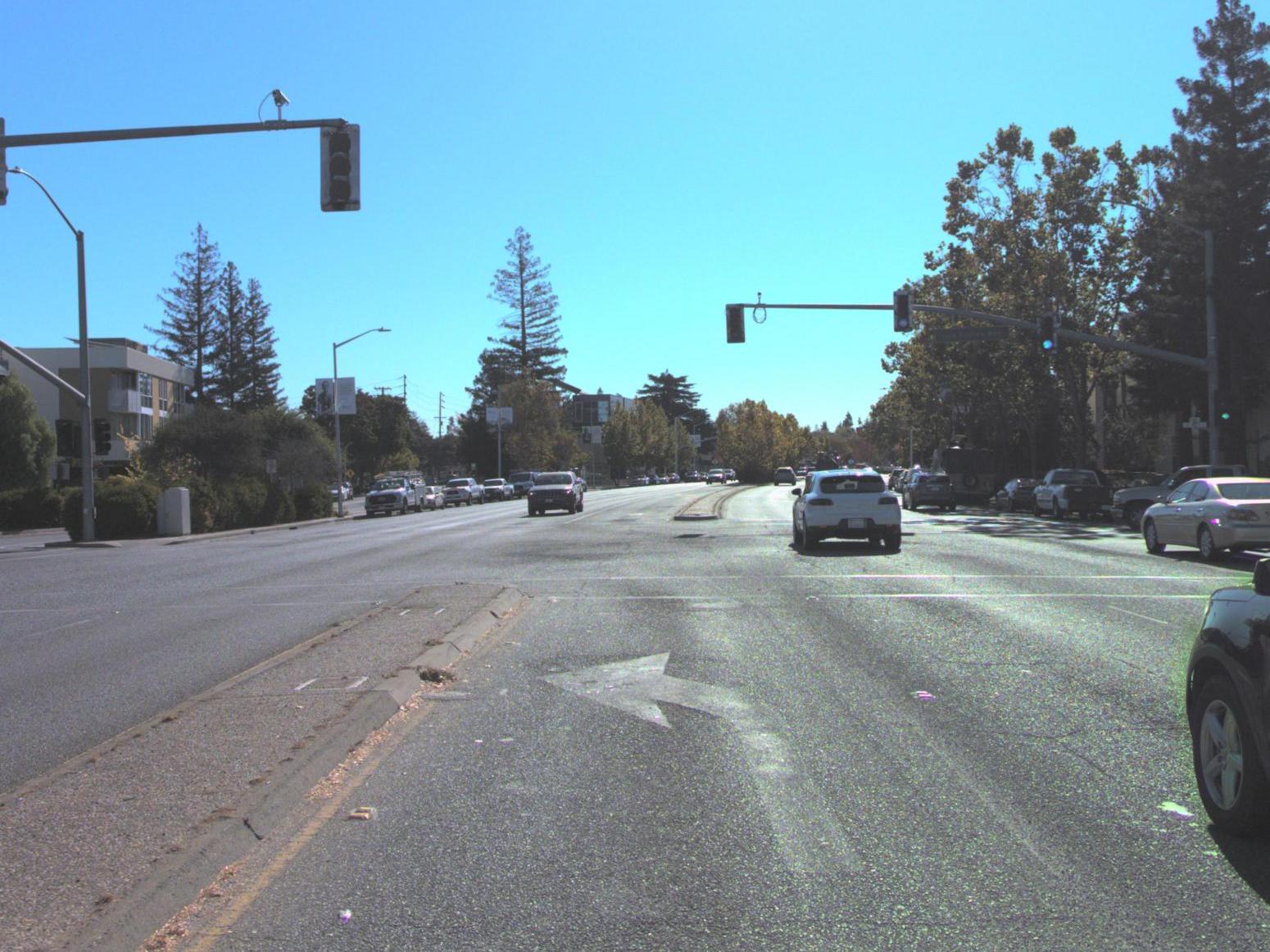}} &
    \fbox{\includegraphics[width=0.23\linewidth]{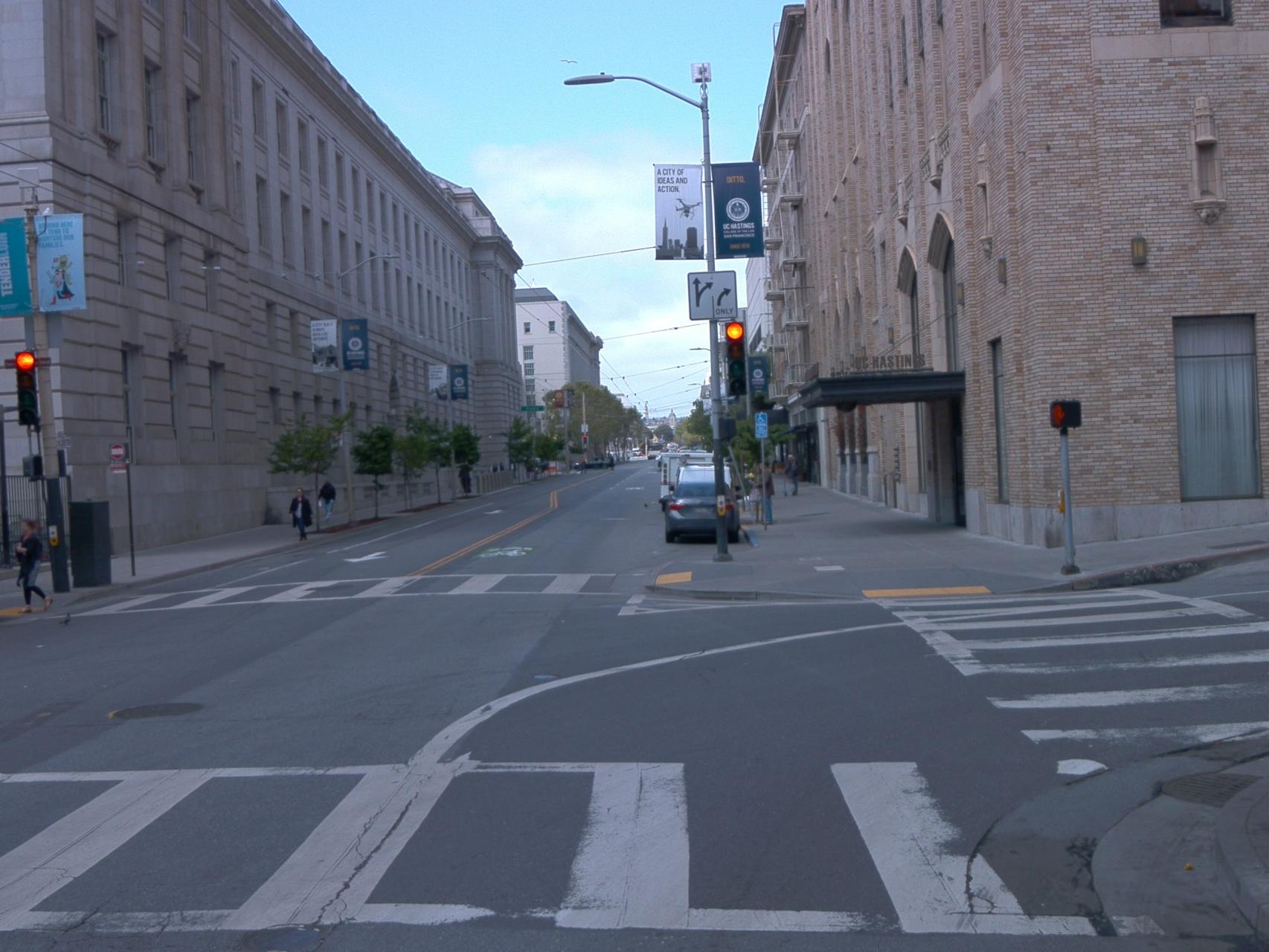}} &
    \fbox{\includegraphics[width=0.23\linewidth]{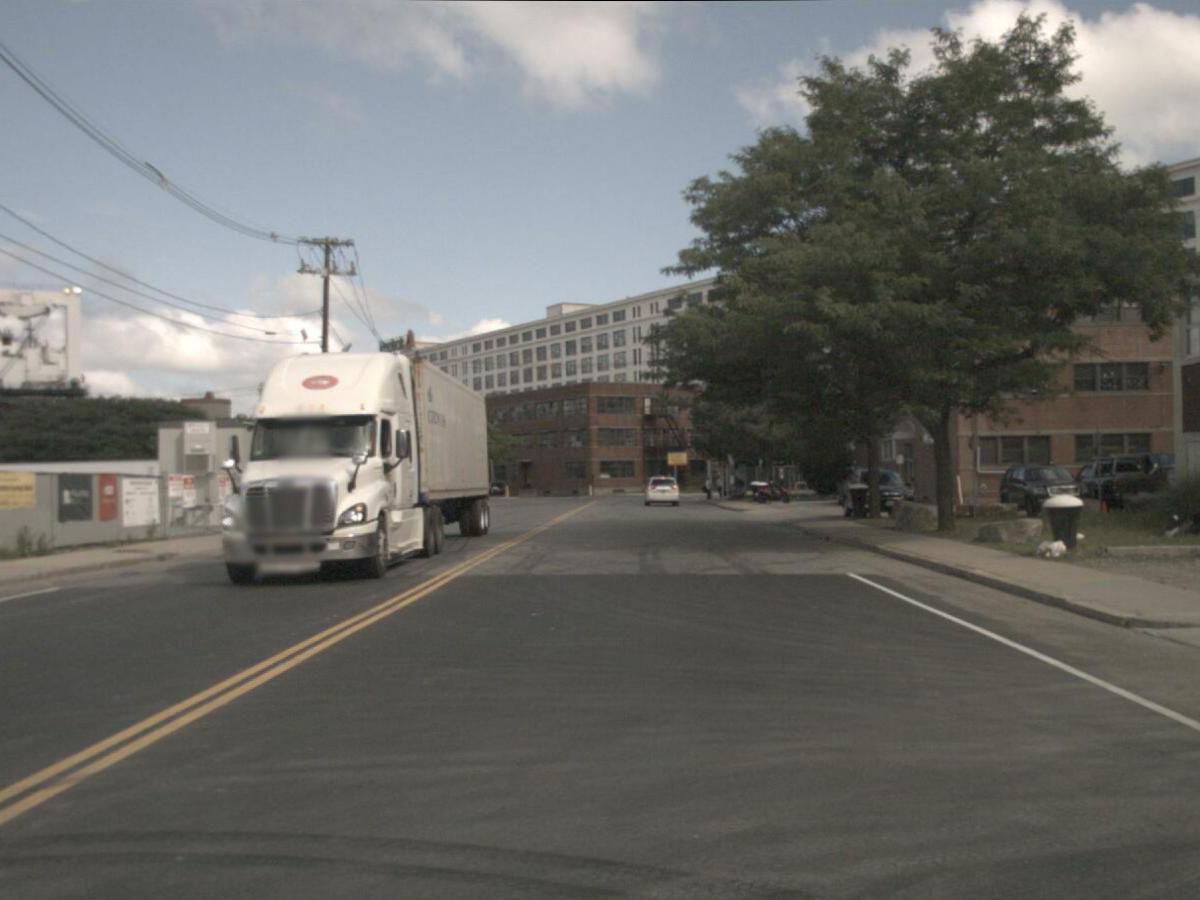}} &
    \fbox{\includegraphics[width=0.23\linewidth]{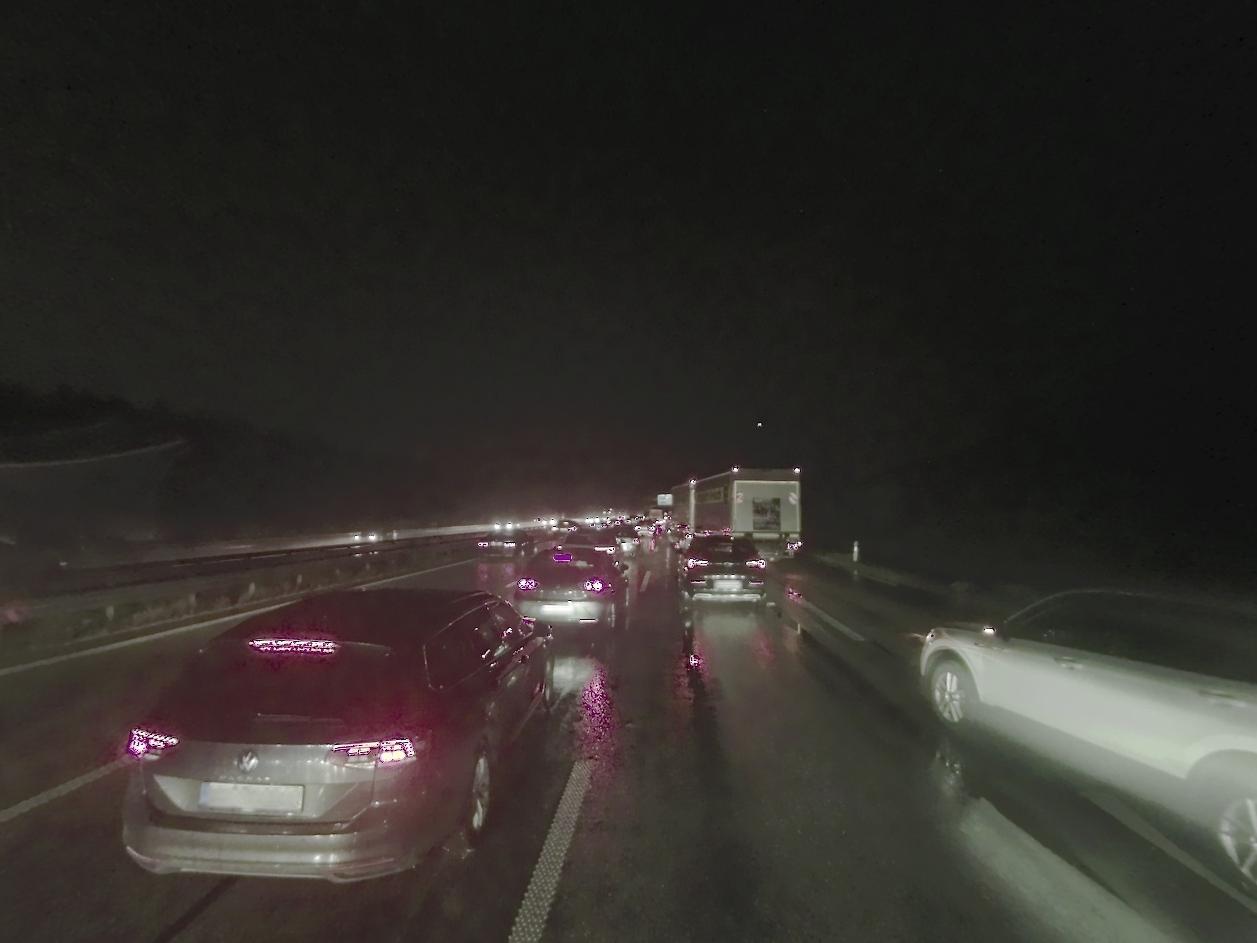}}
  \end{tabular}
\end{minipage} \\
{\figtwosidecaption{Sensor}{}} &
\begin{minipage}{0.95\linewidth}\centering
  \begin{tabular}{@{}c c c c@{}}
    \fbox{\includegraphics[width=0.23\linewidth]{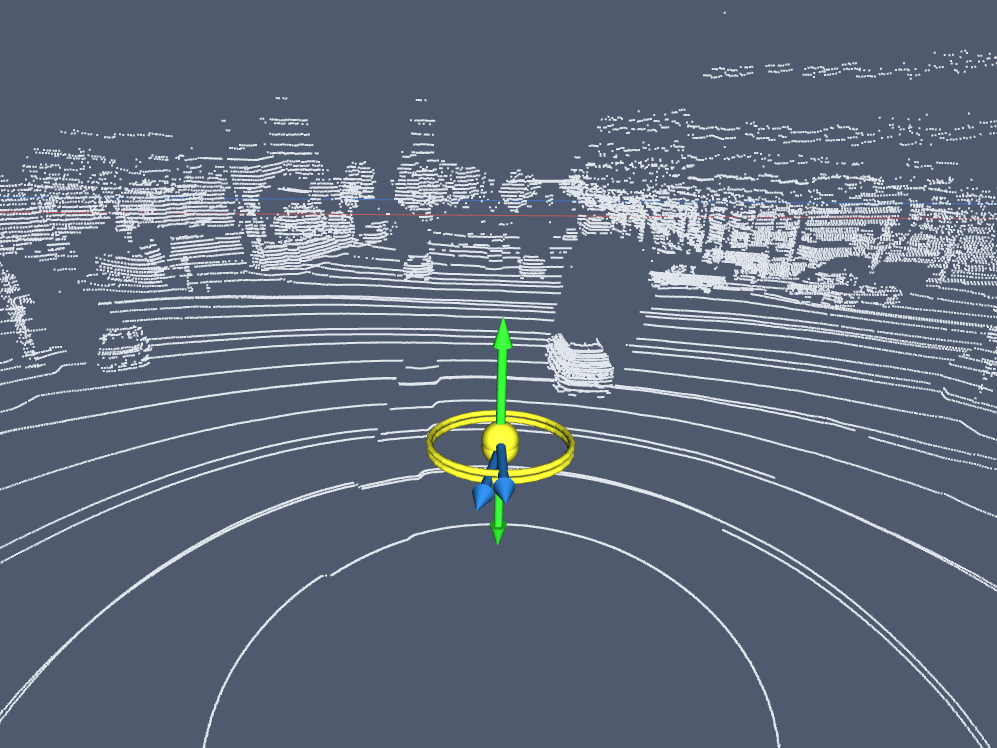}} &
    \fbox{\includegraphics[width=0.23\linewidth]{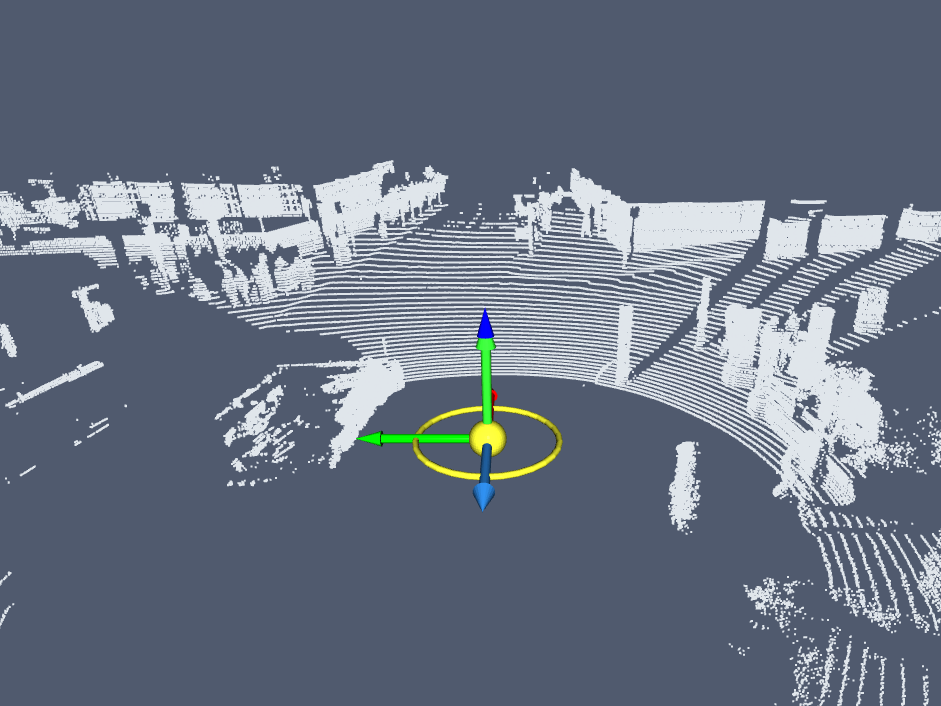}} &
    \fbox{\includegraphics[width=0.23\linewidth]{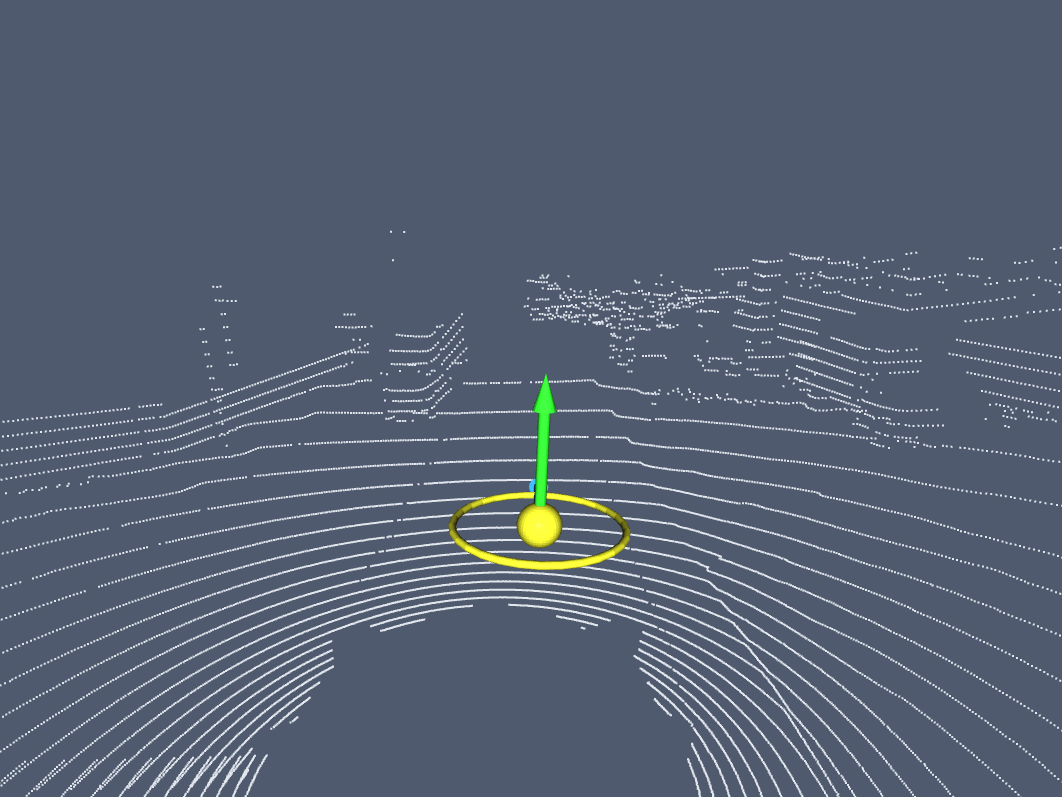}} &
    \fbox{\includegraphics[width=0.23\linewidth]{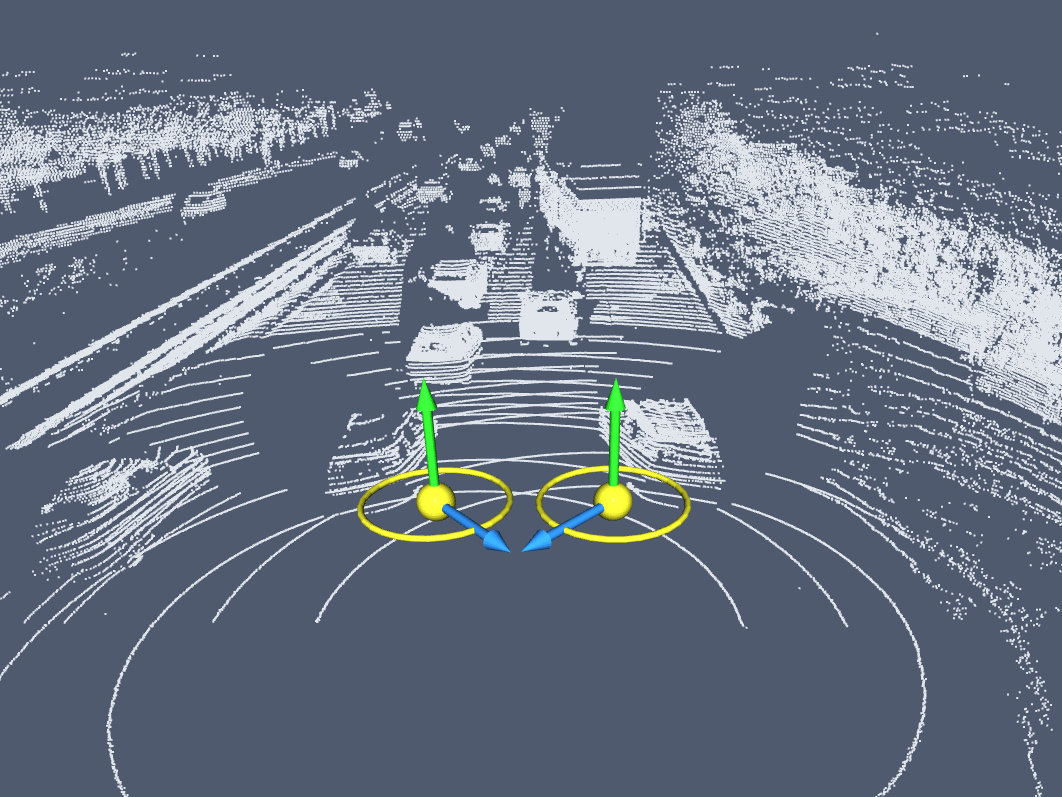}}
  \end{tabular}
\end{minipage} \\
{\figtwosidecaption{BEV}{}} &
\begin{minipage}{0.95\linewidth}\centering
  \begin{tabular}{@{}c c c c@{}}
    \fbox{\includegraphics[width=0.23\linewidth]{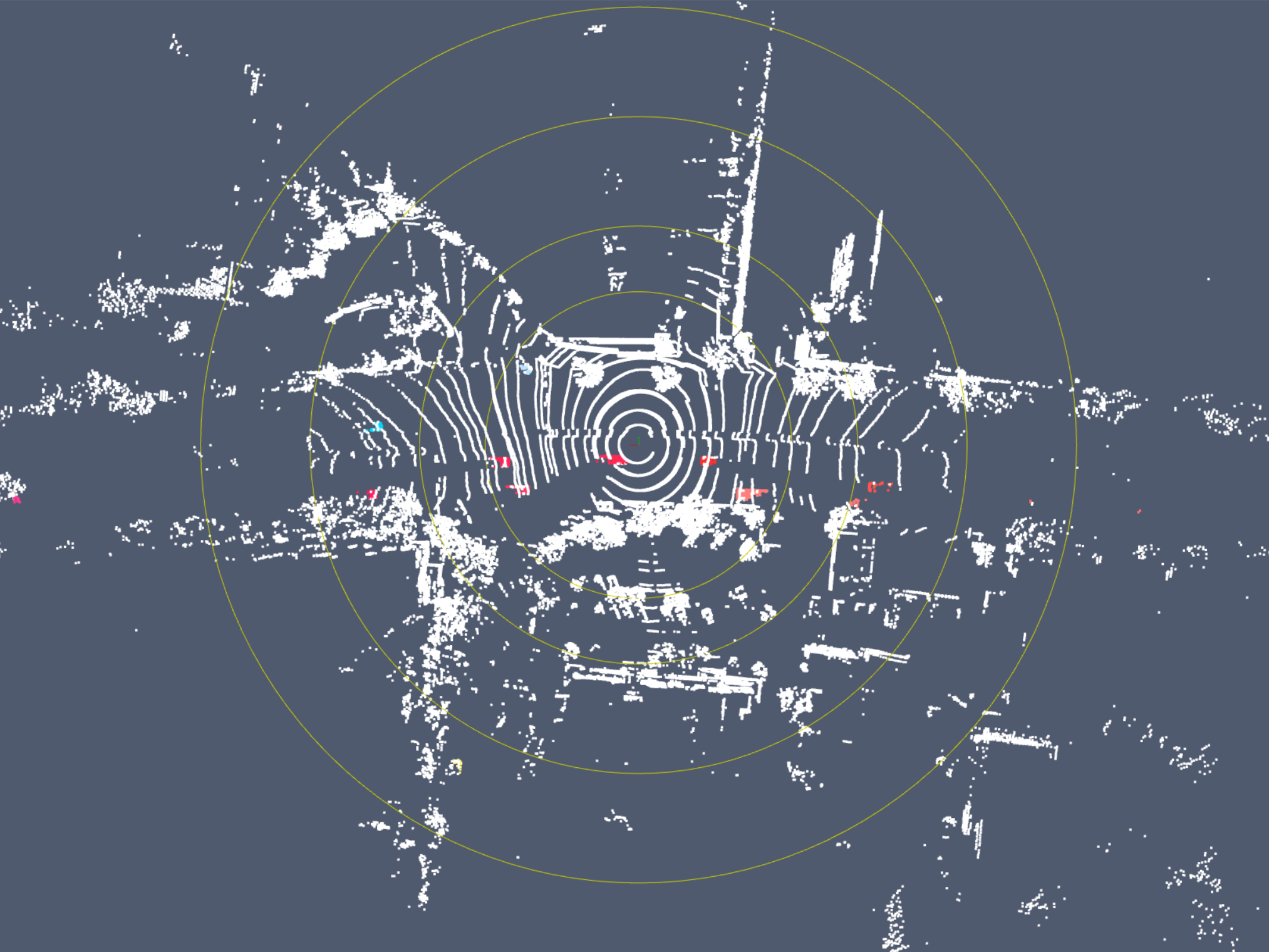}} &
    \fbox{\includegraphics[width=0.23\linewidth]{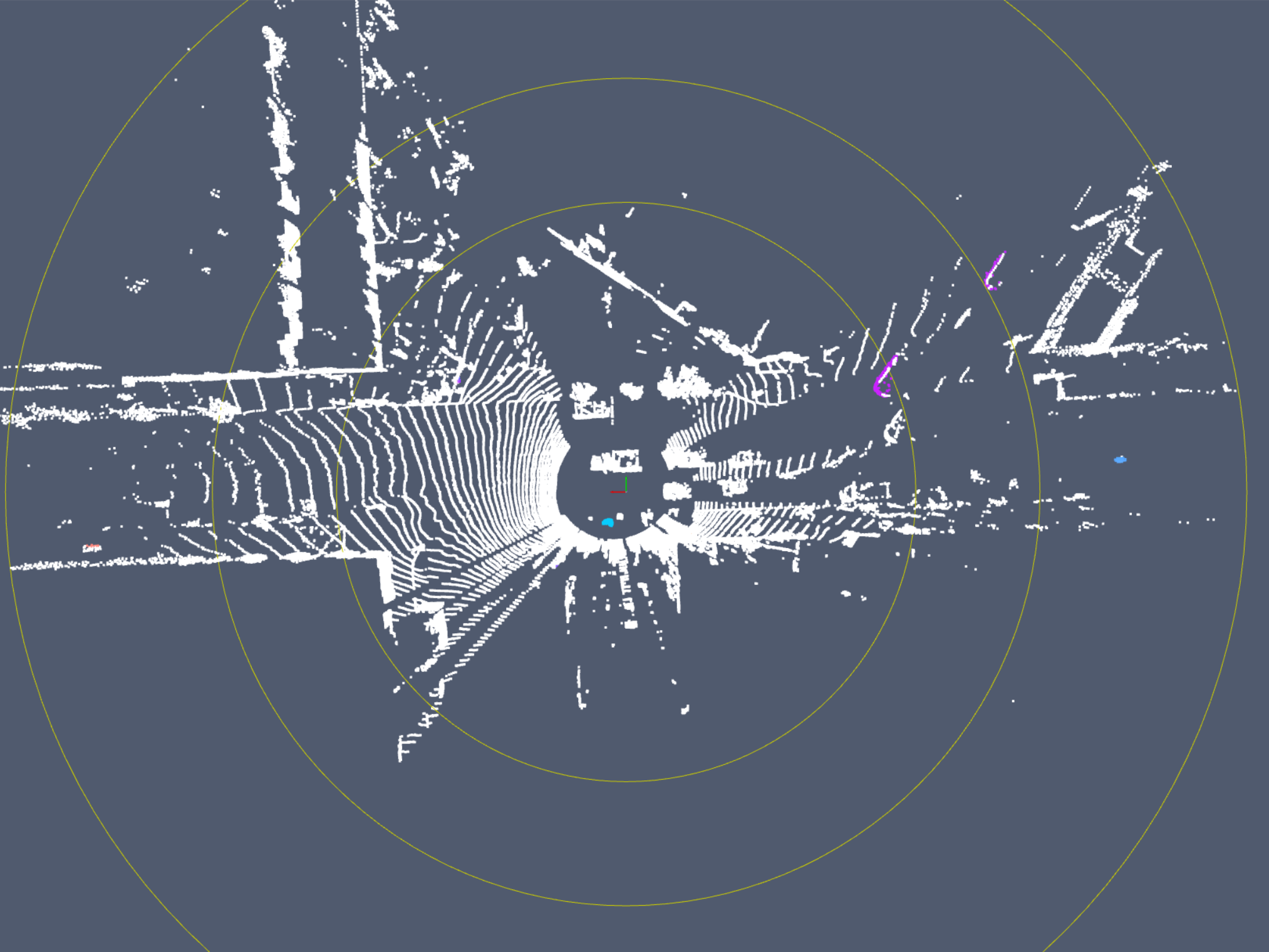}} &
    \fbox{\includegraphics[width=0.23\linewidth]{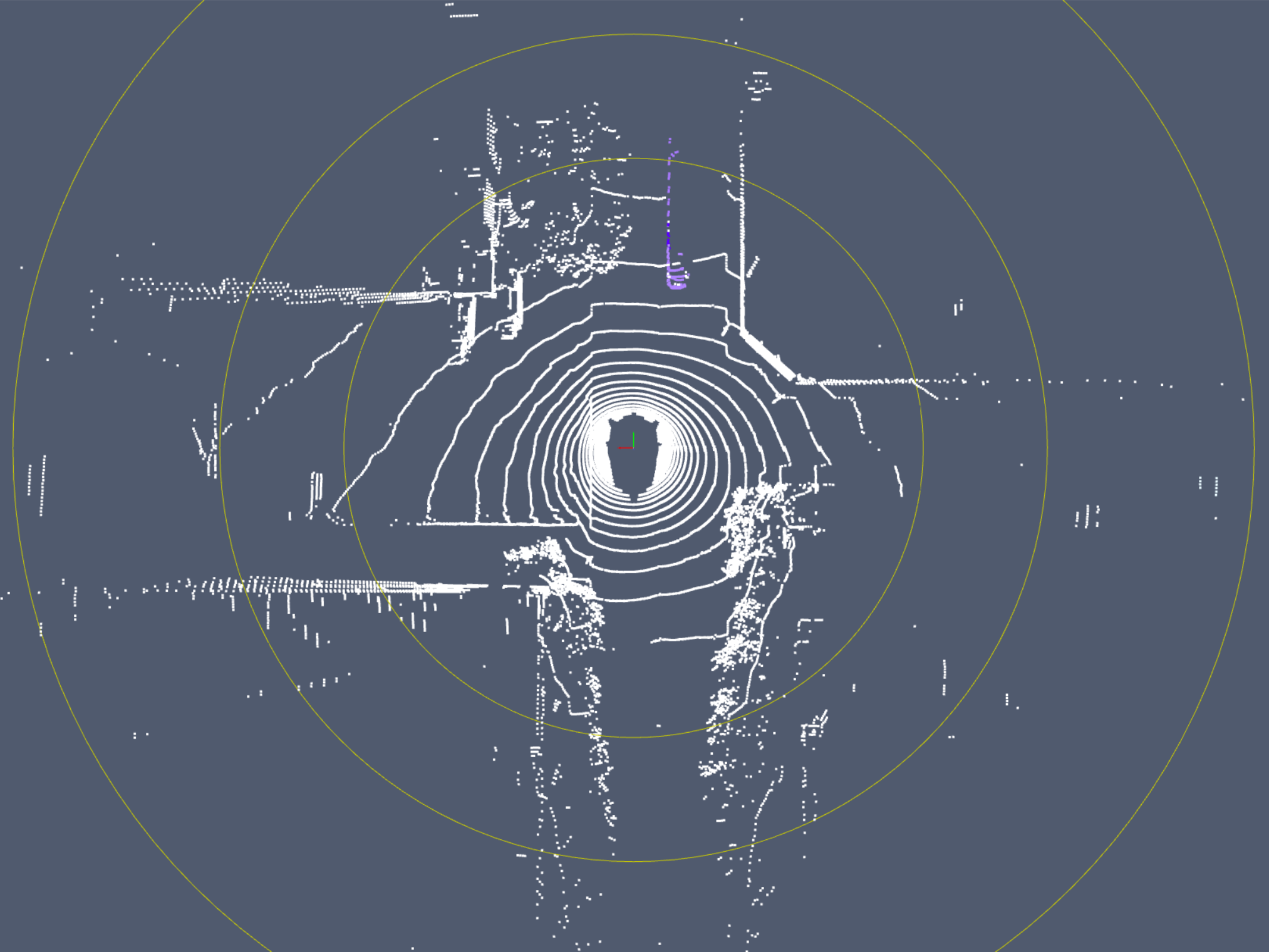}} &
    \fbox{\includegraphics[width=0.23\linewidth]{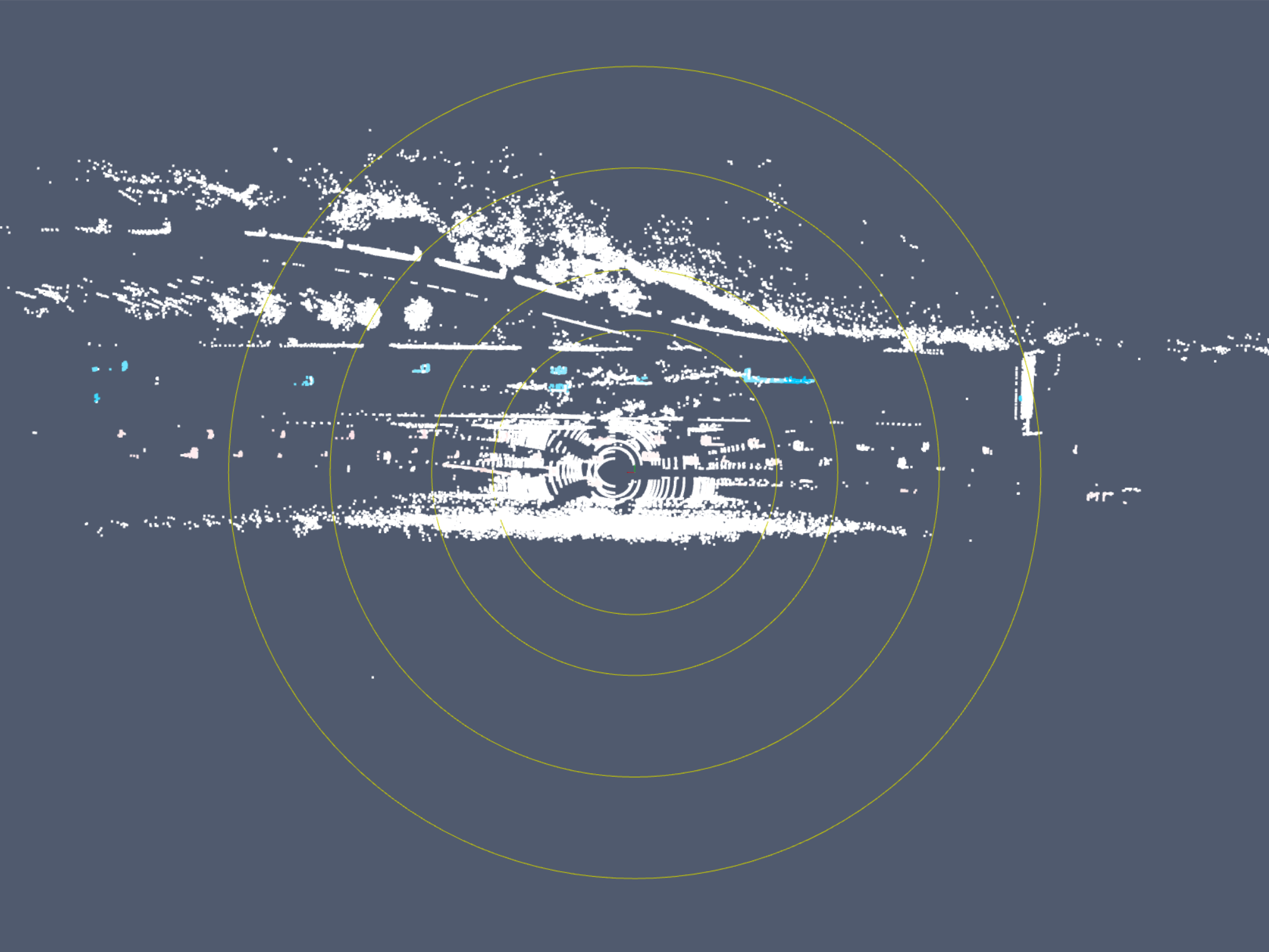}} \\

    \begin{subfigure}{0.23\linewidth}\centering
      \caption*{\textbf{Argoverse 2}}
    \end{subfigure} &
    \begin{subfigure}{0.23\linewidth}\centering
      \caption*{\textbf{Waymo}}
    \end{subfigure} &
    \begin{subfigure}{0.23\linewidth}\centering
      \caption*{\textbf{nuScenes}}
    \end{subfigure} &
    \begin{subfigure}{0.23\linewidth}\centering
      \caption*{\textbf{TruckScenes}}
    \end{subfigure}
  \end{tabular}
\end{minipage} \\
\end{tabular}
\caption{\textbf{Dataset Diversity.} We visualize the front-center RGB (top), LiDAR sensor positions (middle) and BEV LiDAR point clouds (bottom) for Argoverse 2, Waymo, nuScenes and TruckScenes. Notably, all four datasets use different sensors, and collect data in different environments. Specifically, Argoverse 2, Waymo, and nuScenes collect data in urban city centers with sedans, while TruckScenes primarily collects data on highways with a truck. Due to the diversity of environments and sensor configurations, contemporary LiDAR scene flow methods typically only train and evaluate on each dataset independently. However, we find that multi-dataset training significantly improves both in-domain and out-of-domain generalization. Note that RGB images are shown for visualization purposes only; we address LiDAR-only scene flow for AVs.}
\label{fig:teaser}
\end{figure*}

\section{Introduction}
Recent LiDAR scene flow methods achieve remarkable performance on popular autonomous vehicle (AV) datasets~\cite{vedder2024neural,zeroflow,zhang2025deltaflow,khatri2024can} like Argoverse 2, nuScenes, and Waymo \cite{sun2020scalability, nuscenes, Argoverse2_2021}. Notably, current state-of-the-art approaches train and evaluate on each dataset independently and do not attempt to generalize to unseen sensor configurations. Intuitively, since each dataset has different sensor placements and point cloud densities, training separate models for each domain \textit{seems} reasonable (Figure \ref{fig:teaser}). For example, Argoverse 2 (AV2) \cite{Argoverse2_2021} uses two out-of-phase 32 beam LiDARs, nuScenes \cite{nuscenes} uses one 32 beam sensor, and Waymo \cite{sun2020scalability} uses a custom 64 beam sensor. Indeed, it is notoriously challenging to train multi-dataset models for LiDAR-based 3D object detection \cite{wang2020train, soum2023mdt3d} and semantic segmentation \cite{kim2024rethinking, saltori2023walking, liu2024multi} that improve over single dataset models. Notably, prior works that do train on multiple datasets introduce bespoke architectures to explicitly address the problem of cross-domain alignment. 
Our key contribution is an empirical analysis that shows this does {\em not} hold for LiDAR-based scene flow; we simply expose standard architectures {\em without modification} to different sensor data and see a consistent improvement. We posit that ``high-level'' tasks like LiDAR semantic segmentation or 3D object detection are challenging to train across domains because they rely on  human-defined taxonomies \cite{lambert2020mseg} that may be inconsistent across datasets \cite{madan2023revisiting}. Instead, scene flow models appear to capture properties of the underlying physical 4D world that hold regardless of LiDAR sensors observing it.

 \textbf{Cross-Domain Generalization for Low-Level Vision.} We observe similar trends in RGB-based low-level vision tasks, where models learn generalizable geometric representations that transfer well across datasets. For example, FlowNet \cite{dosovitskiy2015flownet, ilg2017flownet} and RAFT \cite{teed2020raft} show that optical flow models trained on synthetic datasets (e.g. FlyingChairs, FlyingThings3D, Sintel) generalize surprisingly well to casually captured videos. More recently, zero-shot monocular depth estimators like MoGe \cite{wang2025moge} and monocular scene flow estimators like ZeroMSF \cite{liang2025zero} demonstrate that training on diverse datasets significantly improves zero-shot performance. Therefore, we posit that low-level LiDAR-based scene flow may similarly generalize to different sensors if trained at scale. 

\begin{figure*}[t]
  \centering
  \includegraphics[width=\linewidth]{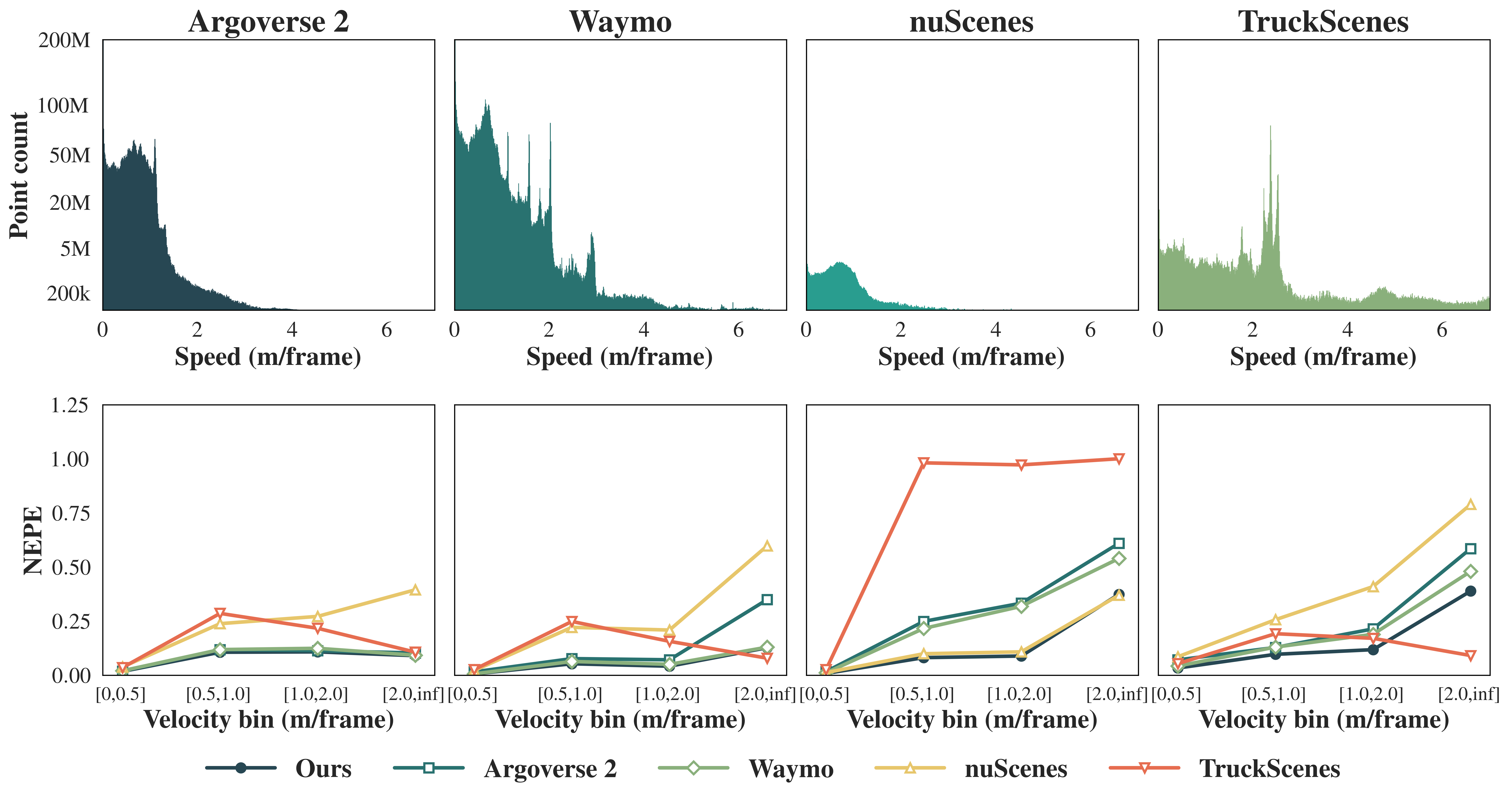}
  \caption{\textbf{Cross-Dataset Generalization Correlates with Velocity Distribution.} We plot the velocity distributions for the AV2, Waymo, nuScenes, and TruckScenes train sets (top) and the Dynamic Mean EPE per velocity bin of Flow4D trained on AV2, Waymo, nuScenes, TruckScenes, and \method \ (bottom). Notably, Flow4D trained on TruckScenes outperforms Flow4D trained on any other dataset for fast moving objects (2.0, $\infty$) across all datasets because it has the largest number of fast moving objects.}
  \label{fig:velocity}
\end{figure*}

To better understand this problem, we first train dataset-specific models for AV2 \cite{Argoverse2_2021}, Waymo \cite{sun2020scalability} and nuScenes \cite{nuscenes} and evaluate zero-shot cross-domain performance. Surprisingly, Flow4D \cite{kim2024flow4d} trained on Waymo achieves similar Dynamic Mean EPE on AV2 as Flow4D trained on AV2. Importantly, Waymo's 64 beam LiDAR is considerably more dense than AV2's two out-of-phase 32-beam LiDARs (Figure \ref{fig:teaser}), indicating that point density does not significantly impact LiDAR scene flow generalization. Further, Flow4D (Waymo) achieves lower EPE (i.e. better performance) than Flow4D (AV2) on fast moving objects in AV2 since there are more fast movers in Waymo than in AV2 (Table \ref{tab:metrics_analysis}). 
This suggests that the accuracy of motion prediction is highly correlated with object velocity (Figure \ref{fig:velocity}); in order to accurately predict fast motion, one should train on fast motion. This suggests that the key to LiDAR scene flow generalization is training on a diverse combination of datasets with a wide range of object velocities.

{
\setlength{\tabcolsep}{0.9em}
\begin{table}[t]
\centering
\caption{\textbf{Analysis on Cross-Domain Generalization.}
We train Flow4D \cite{kim2024flow4d} on AV2, Waymo, and nuScenes and evaluate each dataset-specific model across datasets. We further break down performance by velocity speed buckets (meters / frame). We find that Flow4D (Waymo) nearly matches Flow4D (AV2)'s overall Dynamic Mean EPE on AV2 (0.199 vs. 0.192, shown in green), and even {\em outperforms} it on fast-moving objects (0.098 vs. 0.103, shown in gray). This suggests that Flow4D (Waymo) can \textit{already} generalize across datasets. Flow4D (nuScenes) performs considerably worse on out-of-distribution datasets (shown in red) due its limited number of training examples.}
\label{tab:metrics_analysis}
\resizebox{\linewidth}{!}{
\begin{tabular}{lc|c|c|cccc}
\toprule
\rowcolor{gray!10}
\textbf{Train} & \textbf{Test} & \textbf{FD (cm) $\downarrow$} & \textbf{Dyn.~Mean $\downarrow$} & \textbf{[0, 0.5)} & \textbf{[0.5, 1.0)} & \textbf{[1.0, 2.0)} & \textbf{[2.0, $\infty$)} \\
\midrule
\multirow{3}{*}{AV2}
& AV2      & 8.55  & \cellcolor{darkgreen}0.192 & 0.020 & 0.119 & 0.114 & \cellcolor{lightgray}0.103 \\
& Waymo    & 8.31  & 0.236 & 0.016 & 0.077 & 0.079 & 0.350 \\
& nuScenes & 20.40 & 0.463 & 0.017 & 0.249 & 0.334 & 0.610 \\
\midrule
\multirow{3}{*}{Waymo}
& AV2      & 8.90  & \cellcolor{darkgreen}0.199 & 0.021 & 0.119 & 0.125 & \cellcolor{lightgray}0.098 \\
& Waymo    & 4.58  & 0.215 & 0.007 & 0.064 & 0.051 & 0.130 \\
& nuScenes & 16.04 & 0.408 & 0.017 & 0.217 & 0.319 & 0.540 \\
\midrule
\multirow{3}{*}{nuScenes}
& AV2      & 16.08 & \cellcolor{red}0.357 & 0.039 & 0.239 & 0.272 & 0.395 \\
& Waymo    & 17.24 & \cellcolor{red}0.364 & 0.022 & 0.222 & 0.209 & 0.598 \\
& nuScenes & 7.97  & 0.230 & 0.018 & 0.100 & 0.109 & 0.370 \\
\bottomrule
\end{tabular}
}
\end{table}
}

 \textbf{Towards Zero-Shot LiDAR Scene Flow.} We argue that scaling up LiDAR-based scene flow methods will be a key enabler for 3D motion understanding and dynamic reconstruction in diverse environments. To this end, we retrain state-of-the-art methods {\em without modification} on supervised data from AV2, Waymo, and nuScenes and demonstrate significant improvements for both in-distribution and out-of-distribution generalization. We denote models trained with multiple datasets as \method. \textit{To the best of our knowledge, \method \ is the first to demonstrate the benefits of cross-domain training for LiDAR scene flow.} Further, \method \ models demonstrate remarkable zero-shot accuracy on TruckScenes \cite{fent2024man} (an autonomous truck dataset collected at highway speeds) and AEVAScenes (a small-scale dataset with a novel FMCW LiDAR sensor), outperforming prior dataset-specific models by 30.1\% (Table \ref{tab:truckscenes-val}) and 22.5\% (Table~\ref{tab:aeva-zero-shot}), respectively. Lastly, we find that cross-domain training on near-range (i.e $\le$ 35 meters from the ego-vehicle) points improves scene flow generalization to unseen far-range points (Table \ref{tab:longrange}). We hypothesize that sparse LiDAR points near the ego-vehicle mimic the distribution of dense LiDAR points far away from the vehicle \cite{peri2023empirical}.

 \textbf{Why Does Multi-Dataset Training Work?} Our experiments suggest that multi-dataset training improves LiDAR scene flow performance because scene flow is inherently class-agnostic. This allows us to avoid complications arising from dataset-specific label definitions (Table \ref{tab:semantic-head}). For example, nuScenes defines {\tt bicycle} as excluding the rider, while Waymo includes the rider \cite{madan2023revisiting, robicheaux2025roboflow100}. Such label ambiguity poses a significant challenge for semantic tasks, potentially explaining why naively combining datasets for LiDAR-based 3D object detection and semantic segmentation yields worse performance than single-dataset models. 
 Furthermore, we find that with careful augmentations, we can simulate diverse sensor configurations that improve cross-dataset generalization (Figure \ref{fig:scaling}).
 For example, dropping out LiDAR beams from a Waymo point cloud makes the data look more like a nuScenes point cloud, while augmenting point cloud height makes a sensor mounted on a sedan look like it was mounted on a truck. 

 \textbf{Contributions.} We present three major contributions. First, we highlight that dataset-specific scene flow models {\em already} show signs of cross-domain generalization, motivating our investigation into multi-dataset training. Next, we demonstrate that re-training existing scene flow methods on diverse, large-scale data yields state-of-the-art performance on nuScenes and Waymo, improving over our baselines by 35.2\% and 5.1\% respectively. Finally, we show that \method \ generalizes well to unseen sensors, outperforming dataset-specific models by 30.1\% on TruckScenes, and by 22.5\% on AEVAScenes.


\begin{figure}[t]
    \centering
    \setlength{\tabcolsep}{2pt}
    \begin{tabular}{@{}c@{\hspace{0.6em}}c@{}}
        {\figtwosidecaption{Overall}{Performance}} &
        \begin{minipage}{0.94\linewidth}\centering
            \begin{subfigure}{0.32\linewidth}
                \includegraphics[width=\linewidth]{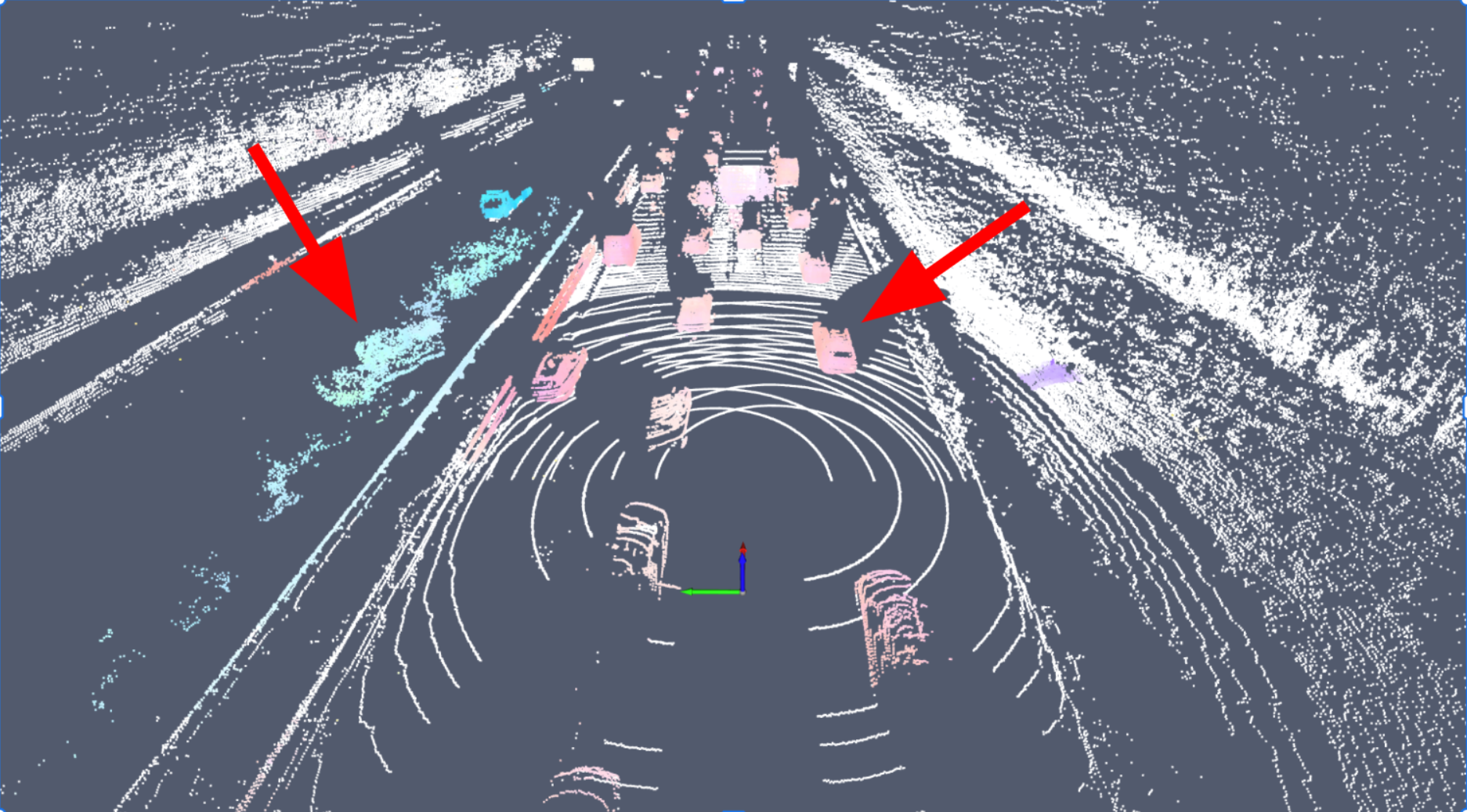}
            \end{subfigure}
            \begin{subfigure}{0.32\linewidth}
                \includegraphics[width=\linewidth]{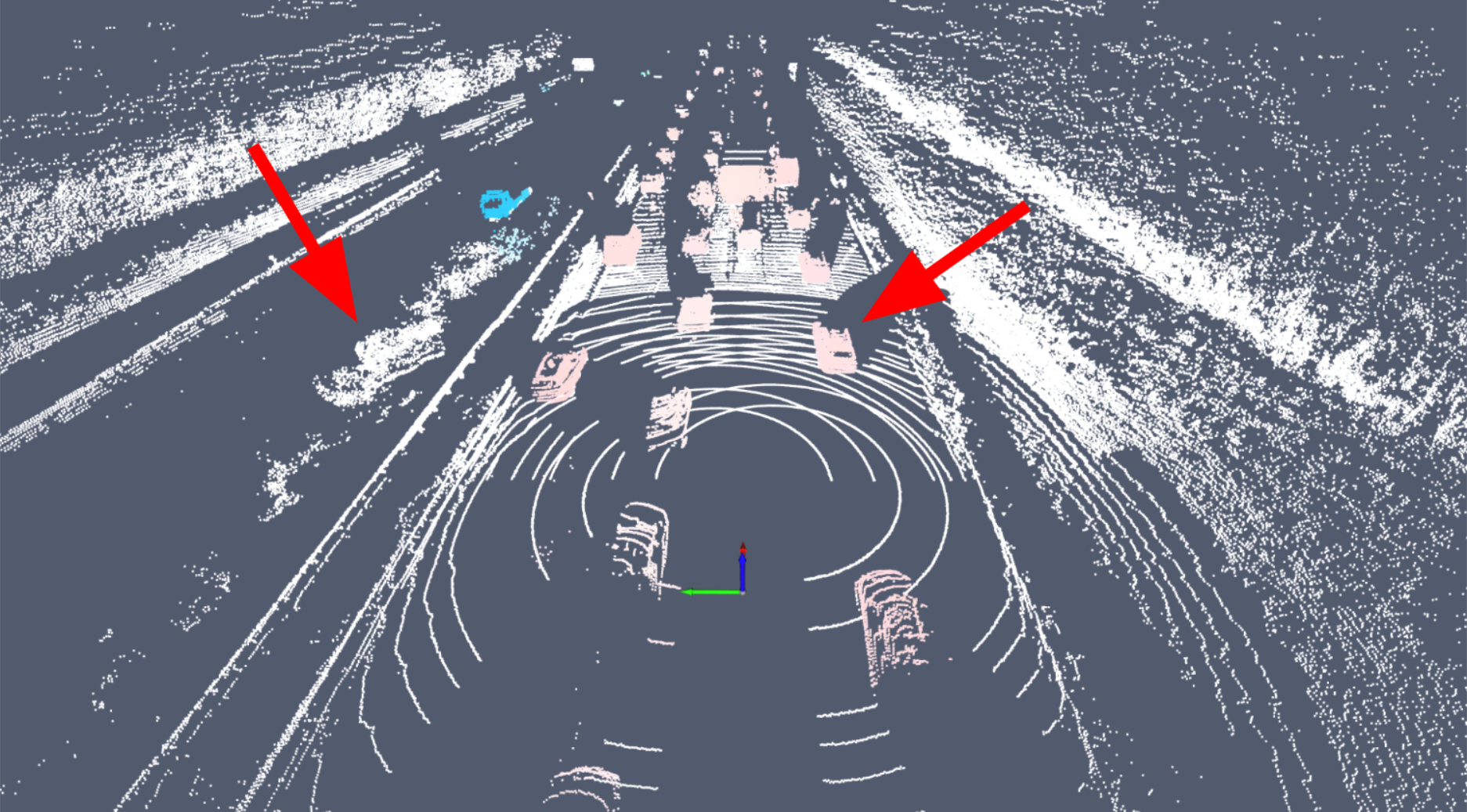}
            \end{subfigure}
            \begin{subfigure}{0.32\linewidth}
                \includegraphics[width=\linewidth]{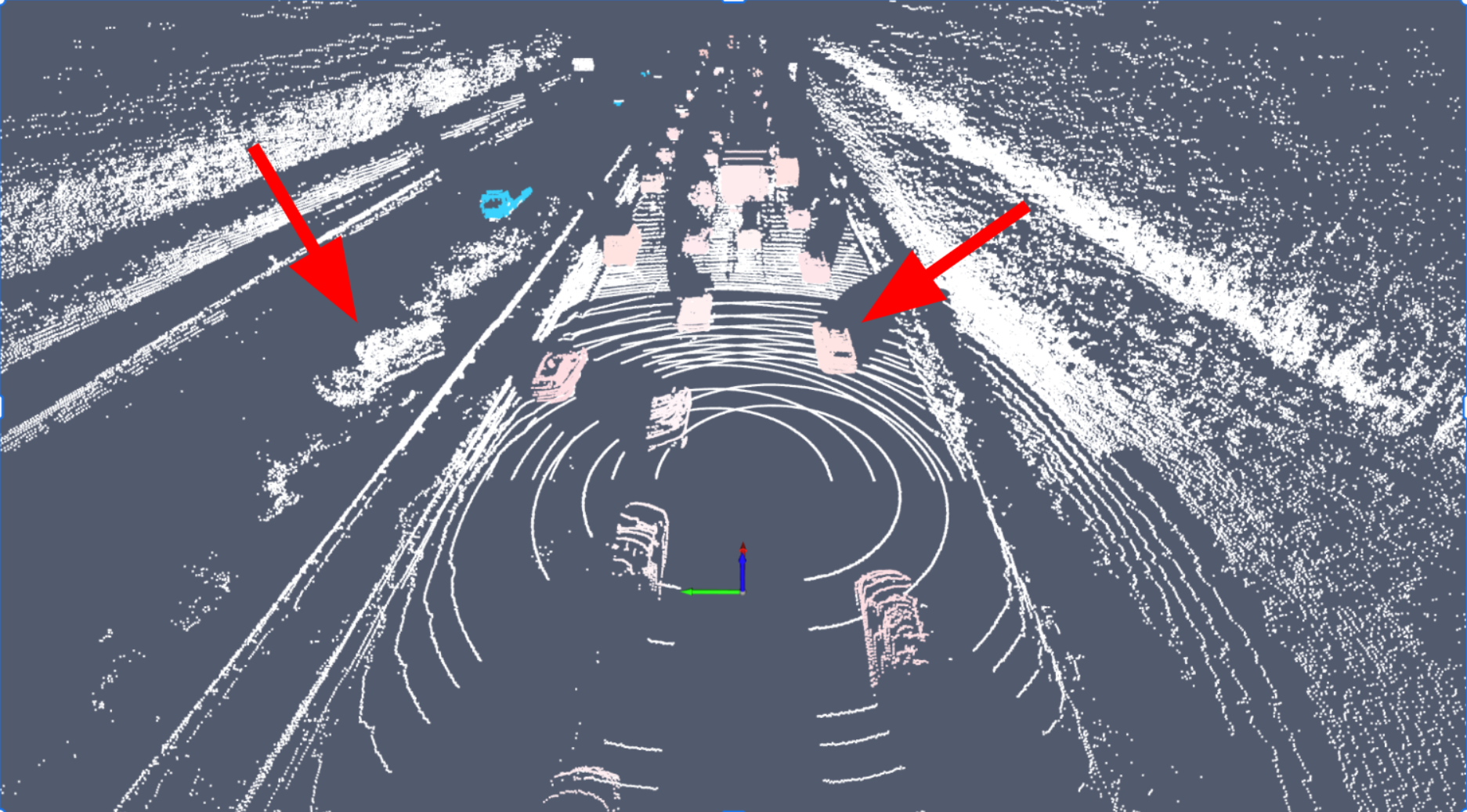}
            \end{subfigure}
        \end{minipage} \\
        \noalign{\vskip 0.4em}
        {\figtwosidecaption{Rare}{Vehicles}} &
        \begin{minipage}{0.94\linewidth}\centering
            \begin{subfigure}{0.32\linewidth}
                \includegraphics[width=\linewidth]{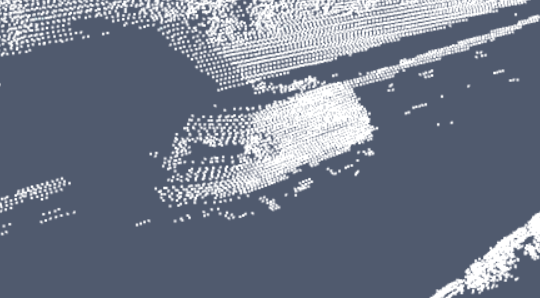}
            \end{subfigure}
            \begin{subfigure}{0.32\linewidth}
                \includegraphics[width=\linewidth]{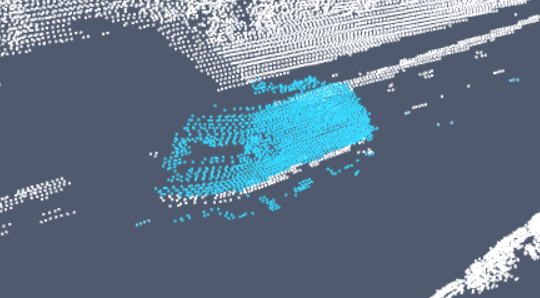}
            \end{subfigure}
            \begin{subfigure}{0.32\linewidth}
                \includegraphics[width=\linewidth]{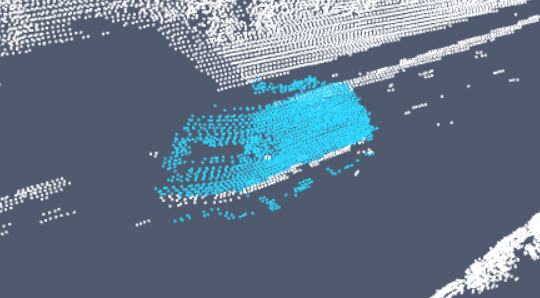}
            \end{subfigure}
        \end{minipage} \\
        \noalign{\vskip 0.4em}
        {\figtwosidecaption{Long Range}{(\(\sim35\!-\!75\) m)}} &
        \begin{minipage}{0.94\linewidth}\centering
            \begin{subfigure}{0.32\linewidth}
                \includegraphics[width=\linewidth]{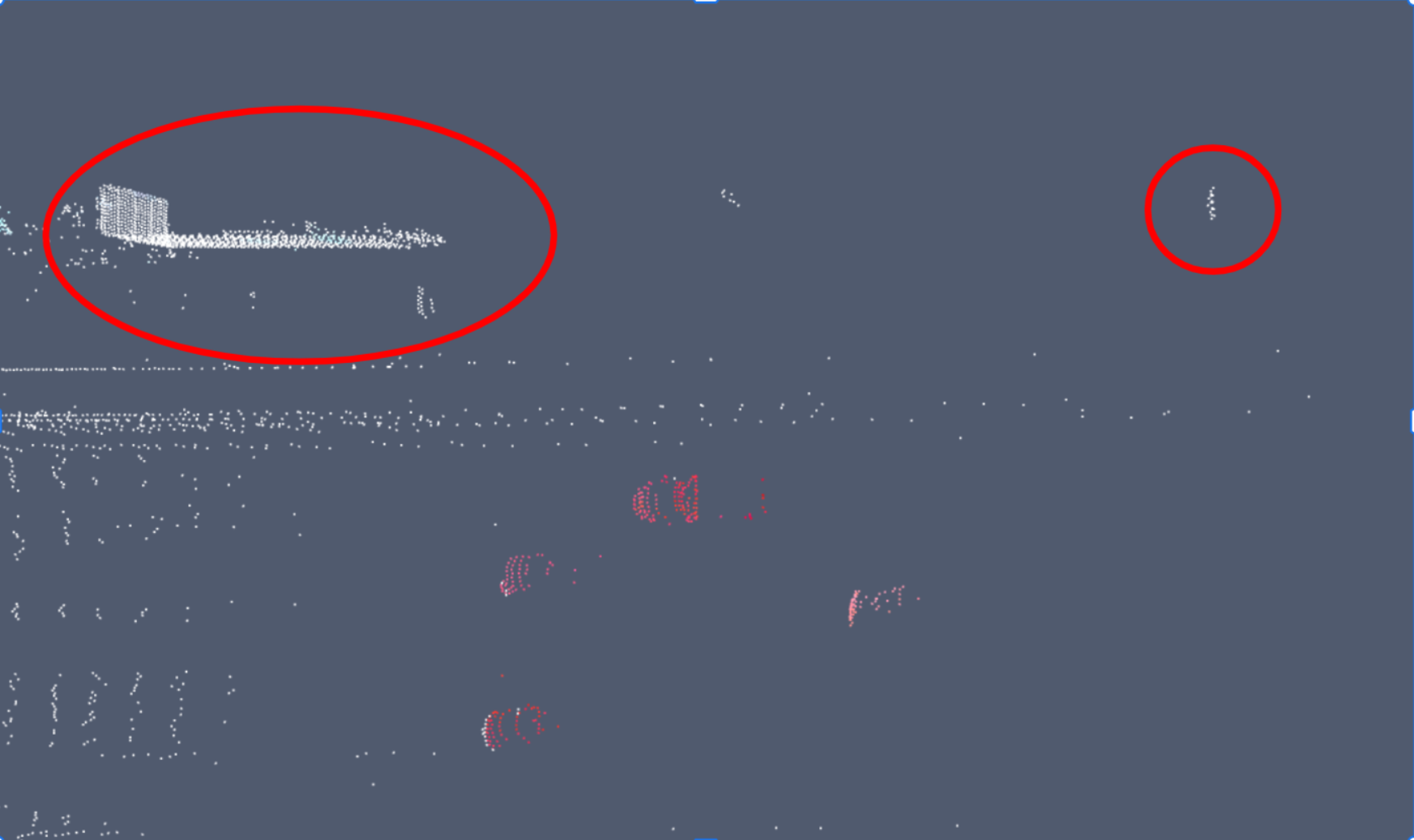}
                \caption*{\textbf{(a) \(\Delta\)Flow}}
            \end{subfigure}
            \begin{subfigure}{0.32\linewidth}
                \includegraphics[width=\linewidth]{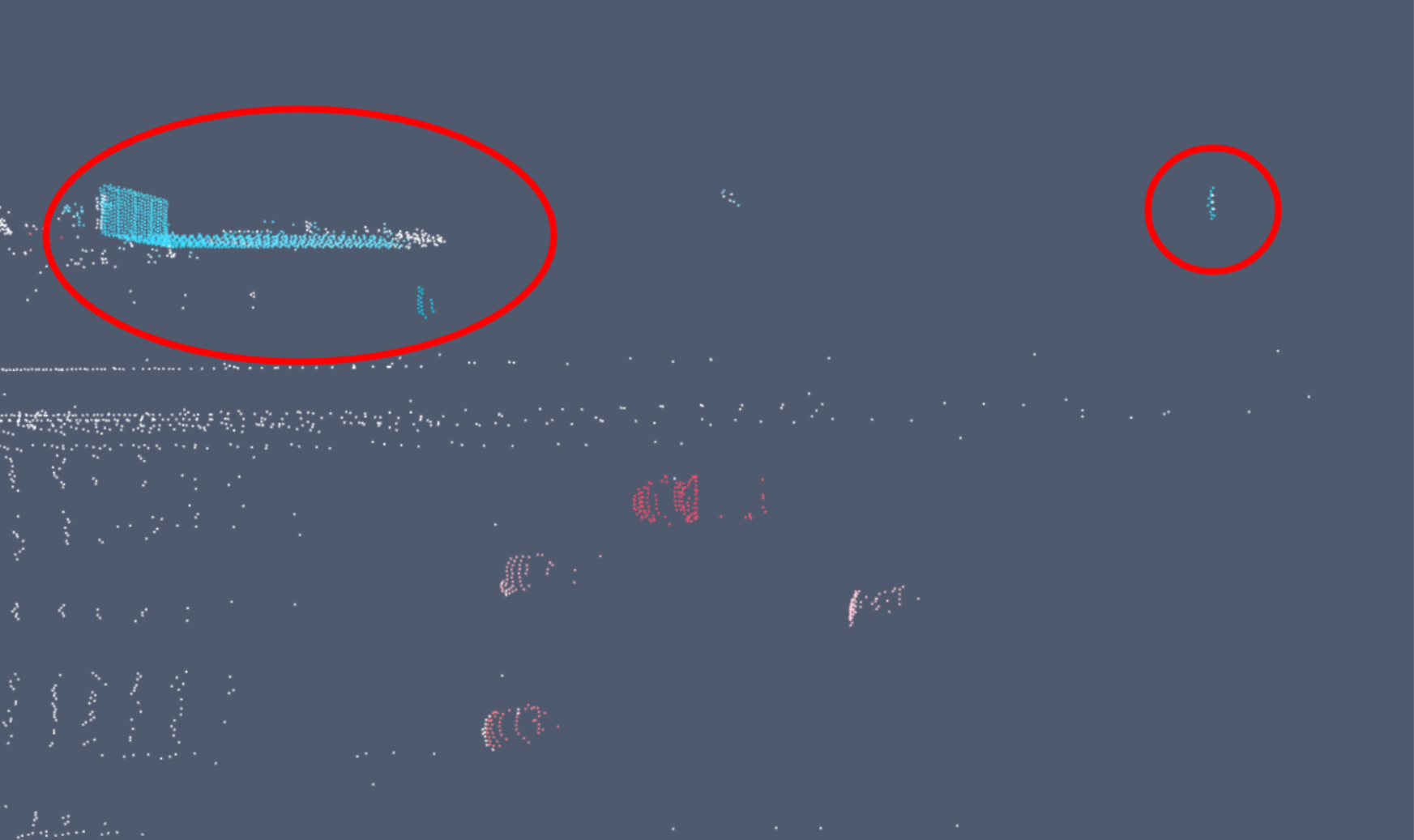}
                \caption*{\textbf{(b) \(\Delta\)Flow (\method)} }
            \end{subfigure}
            \begin{subfigure}{0.32\linewidth}
                \includegraphics[width=\linewidth]{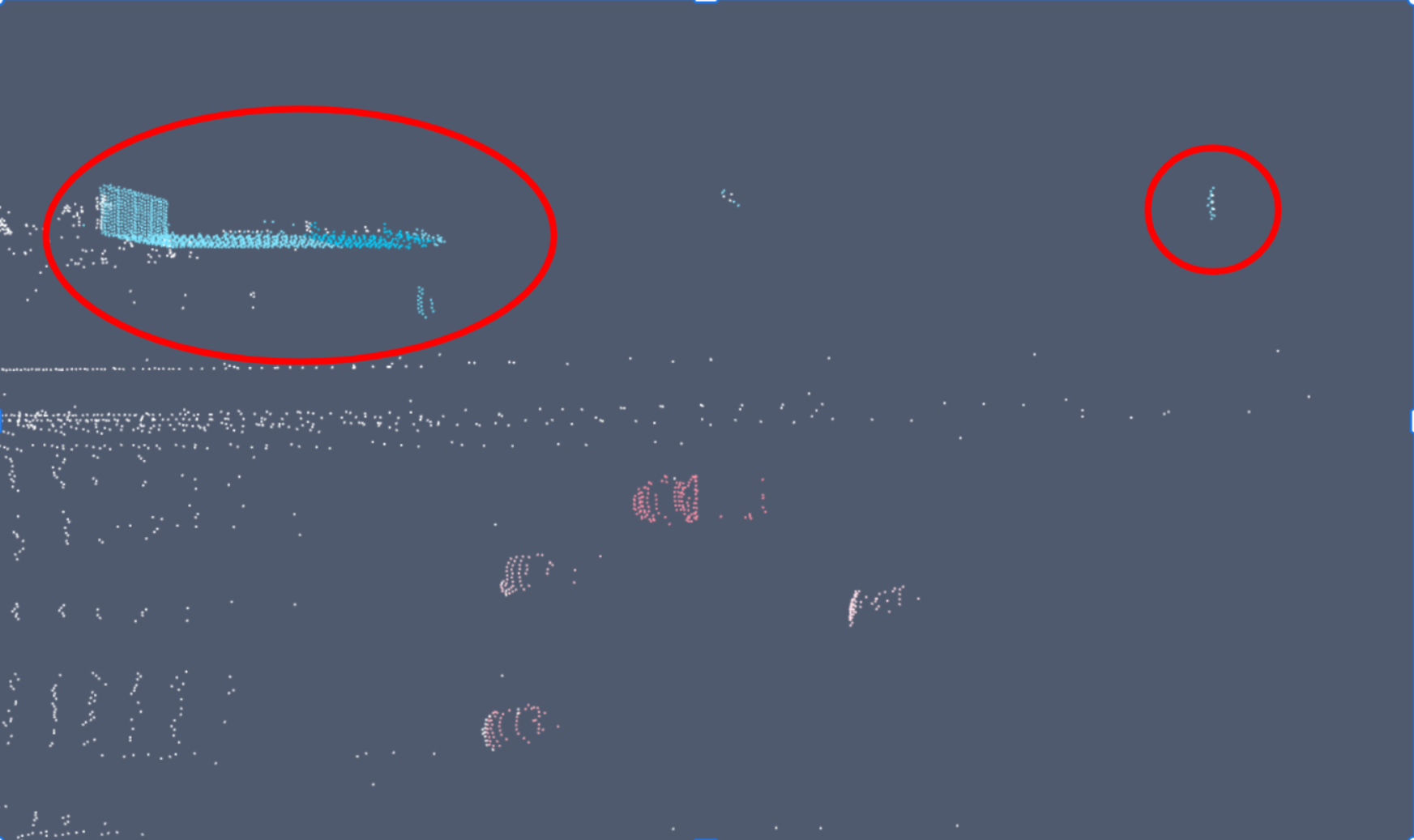}
                \caption*{\textbf{(c) Ground truth}}
            \end{subfigure}
        \end{minipage}
    \end{tabular}
    \caption{\textbf{Zero-Shot Generalization on TruckScenes.} We qualitatively compare the dataset-specific \(\Delta\)Flow model and our \(\Delta\)Flow (\method) model above. \(\Delta\)Flow (UniFlow) produces more accurate vehicle motion estimates in general, and avoids falsely predicting motion for rain artifacts (on the top left)  as seen in \(\Delta\)Flow (top row). Next, \(\Delta\)Flow (UniFlow) correctly estimates the motion of the van since it has been trained on more examples of rare classes (middle row). Lastly, \(\Delta\)Flow (UniFlow) generalizes significantly better at long-range, accurately estimating the truck's motion at $\sim$35\,m and the car's motion at $\sim$70\,m, despite only being trained on LiDAR points up to $35$m (bottom row).}
    \label{fig:qualitative_results}
\end{figure}

\section{Related Works}
 \textbf{Scene Flow Estimation} is the task of describing the 3D motion field between temporally successive point clouds~\cite{vedula2005three, khatri2024can,zhang2025deltaflow}, and is an important primitive for 3D motion understanding and dynamic scene reconstruction \cite{wang2025fluxd}. Early approaches~\cite{wei2020pv,scoop,wang2023dpvraft,zhang2024gmsf} learned point-wise features to estimate per-point flow but struggled to scale to large outdoor environments~\cite{zhang2024deflow,zeroflow}. More recent work~\cite{khoche2025ssf,kim2024flow4d,mambaflow,hoffmann2025floxels, zeroflow} jointly estimates flow for all points in a scene. Contemporary methods can be broadly classified into feedforward models and optimization-based methods. Feedforward models directly learn a mapping between point cloud pairs and flow fields, but require large-scale human annotations~\cite{fastflow3d,kim2024flow4d}. For real world
datasets (typically from the autonomous vehicle domain), these human annotations are
provided in the form of 3D bounding boxes and tracks for every object in the
scene. In contrast, optimization-based methods do not require labeled data, and instead optimize a learned representation per-scene ~\cite{li2021neural,li2023fast}. This per-scene optimization is prohibitively expensive, making it difficult to scale to large datasets. For example, EulerFlow~\cite{vedder2024neural} achieves state-of-the-art unsupervised accuracy on many popular AV datasets, but takes over 24 hours to optimize each scene. In summary, feedforward models achieve efficient inference but demonstrate limited generalization beyond their training data, while optimization-based methods produce high quality flow for diverse scenes, but are too slow for real-time applications. Our work aims to reconcile this trade-off by training a single feedforward model that achieves robust generalization across diverse datasets and sensors.

 \textbf{Cross-Domain Generalization for LiDAR} is a long-standing challenge in LiDAR-based perception. Prior works in 3D object detection~\cite{malic2025gblobs,hegde2025multimodal} and semantic segmentation~\cite{kim2024rethinking,caunes2025seg_3d_by_pc2d,xu2025superflowpp} show that naively training models on multiple LiDAR datasets often performs worse than models trained on a single dataset. This performance drop can be largely attributed to diverse sensor hardware (e.g., number of beams, scan pattern, point cloud density), environmental conditions (e.g., weather, geography), and sensor placement across datasets. Often, such methods also design bespoke architectures to address cross-domain generalization. In addition, several methods have introduced data augmentation strategies~\cite{sun2024empirical}, such as random point dropping~\cite{wang2021pointaugmenting} and object scaling~\cite{cen2022open}, while others~\cite{Liu_2024_CVPR,michele2024saluda} have proposed unsupervised domain adaptation techniques that align feature distributions across datasets. However, such data augmentation strategies primarily address the \textit{geometric domain gap} between datasets. For motion-centric tasks such as scene flow, we argue that understanding the distribution of object velocities between datasets represents a critical yet overlooked axis for addressing cross-domain generalization. To this end, our work presents the first systematic study of this \emph{velocity domain gap}.

\section{\method: Towards Zero-Shot LiDAR Scene Flow for Autonomous Vehicles}
\method \ aims to learn general motion priors that transfer to diverse and unseen LiDAR sensors by training a single scene flow model across diverse autonomous driving datasets. In this section, we describe the process of unifying five popular AV datasets and our training strategy. 

\textbf{Dataset Unification.} We unify five widely used AV datasets to facilitate multi-dataset training. Importantly, we standardize LiDAR sensor frame rates and annotation frequencies to ensure that the time interval between LiDAR sweeps is consistent across datasets. This step is crucial because state-of-the-art scene flow methods parameterize motion as the displacement between two consecutive LiDAR frames. Specifically, we down-sample high-frequency datasets like nuScenes (which provides LiDAR sweeps at 20 Hz) to 10 Hz to match AV2 and Waymo. For datasets with sparse scene flow annotations (e.g., nuScenes annotates tracks at 2 Hz), we train solely on the annotated frames and their real adjacent scans. This ensures that we do not introduce any label noise or motion artifacts due to interpolation. Lastly, we apply LineFit~\cite{himmelsbach2010fast} for ground point removal for all datasets.

\textbf{Unified Multi-Dataset Training.}
We retrain existing state-of-the-art scene flow architectures (SSF \cite{khoche2025ssf}, Flow4D \cite{kim2024flow4d}, and \(\Delta\)Flow~\cite{zhang2025deltaflow}) 
on the unified dataset mixture of nuScenes, AV2, and Waymo.
Importantly, we do not re-weight dataset frequencies or apply dataset-specific augmentations. Instead, we employ a minimal yet effective augmentation set including height randomization to account for sensor placement ($z \pm \text{0.5 - 2m}$, $p=0.8$) and LiDAR ray dropping to simulate varying LiDAR point densities (alternating odd/even rays, $p=0.35$). Ray dropping is not performed on nuScenes due to its inherent sparsity. Further, we do not discard any scans and use dataset provided ego-motion for motion compensation. We do not temporally align sensor measurements between datasets beyond frame rate. This unified training paradigm encourages models to learn domain-agnostic geometric priors rather than memorizing dataset-specific motion patterns, resulting in significant zero-shot generalization improvements. See Appendix \ref{sec:impl_details} for more details.

\begin{table}[t] 
\label{tab:datasets}
\centering
\caption{\textbf{Dataset Statistics.} We summarize the five datasets used for training and evaluation below. We include both the (\# Total Frames / \# Annotated Frames) for nuScenes and TruckScenes.}
\label{tab:dataset_stats}
\resizebox{\linewidth}{!}{
\begin{tabular}{lcccccc}
\toprule
\rowcolor{gray!10}
\textbf{Dataset} & \textbf{Split} & \textbf{\# Scenes} & \textbf{\# Frames} & \textbf{LiDAR Setup} & \textbf{Ego-Vehicle} & \textbf{Scenario} \\
\midrule
\multirow{2}{*}{Argoverse 2} & Train & 700 & 110,071 & \multirow{2}{*}{2 x 32-beam} & \multirow{2}{*}{Car} & \multirow{2}{*}{Urban / City} \\
 & Val & 150 & 23,547 & & & \\
\midrule
\multirow{2}{*}{Waymo} & Train & 798 & 155,000 & \multirow{2}{*}{1 x 64-beam} & \multirow{2}{*}{Car} & \multirow{2}{*}{Urban / Suburban} \\
 & Val & 202 & 40,000 & & & \\
\midrule
\multirow{2}{*}{nuScenes} & Train & 700 & 137,575 / 27,392 & \multirow{2}{*}{1 x 32-beam} & \multirow{2}{*}{Car} & \multirow{2}{*}{Urban / City} \\
 & Val & 150 & 29,126 / 5,798 & & & \\
\midrule
\multirow{2}{*}{TruckScenes} & Train & 524 & 101,902 / 20,380 & \multirow{2}{*}{2 x 64-beam} & \multirow{2}{*}{Truck} & \multirow{2}{*}{Highway} \\
 & Val & 75 & 14,625 / 2,925 & & & \\
\midrule
AEVAScenes & Val & 65 & 6,500 & 6$\times$ FMCW & Car & Urban / Highway
\\
\bottomrule
\end{tabular}
}
\end{table}

\section{Experiments}
In this section, we evaluate \method \ on Argoverse 2, Waymo, nuScenes, TruckScenes, and AEVAScenes. We ablate the impact of dataset characteristics on model performance, and identify scaling trends. We encourage the reader to see the supplement for additional analysis and visuals. 

{
\setlength{\tabcolsep}{0.75em}
\begin{table}[t]
\centering
\caption{\textbf{Argoverse 2 Test Set.} We compare UniFlow  with recent methods on the Argoverse 2 test set. Notably, Flow4D-XL (\method) outperforms Flow4D by 9.0\%, while SSF (\method) outperforms SSF by 13.8\% Dynamic Mean EPE. Notably, \(\Delta\)Flow (\method) achieves slightly worse mean Dynamic EPE than \(\Delta\)Flow. This suggests that performance on AV2 may be saturating, and further improvements may be overfitting to label noise.
}
\label{tab:av2}
\resizebox{\linewidth}{!}{
\begin{tabular}{l|cccc|ccccc}
\toprule
\multirow{2}{*}{\textbf{Methods}} & \multicolumn{4}{c|}{\textbf{Three-way EPE (cm) $\downarrow$}} & \multicolumn{5}{c}{\textbf{Dynamic Bucket-Normalized $\downarrow$}} \\
 & \textbf{Mean} & \textbf{FD} & \textbf{FS} & \textbf{BS} & \textbf{Mean} & \textbf{Car} & \textbf{Other} & \textbf{Pedestr.} & \textbf{VRU} \\
\midrule
\rowcolor{gray!10}
\textit{Unsupervised} & \multicolumn{4}{c|}{} & \multicolumn{5}{c}{} \\
NSFP        & 6.06 & 11.58 & 3.16 & 3.44 & 0.422 & 0.251 & 0.331 & 0.722 & 0.383 \\
FastNSF     & 11.18 & 16.34 & 8.14 & 9.07 & 0.383 & 0.296 & 0.413 & 0.500 & 0.322 \\
SeFlow      & 4.86 & 12.14 & 1.84 & 0.60 & 0.309 & 0.214 & 0.291 & 0.464 & 0.265 \\
ICP Flow    & 6.50 & 13.69 & 3.32 & 2.50 & 0.331 & 0.195 & 0.331 & 0.435 & 0.363 \\
EulerFlow   & 4.23 & 4.98 & 2.45 & 5.25 & 0.130 & 0.093 & 0.141 & 0.195 & \textbf{0.093} \\
\midrule
\rowcolor{gray!10}
\textit{Supervised} & \multicolumn{4}{c|}{} & \multicolumn{5}{c}{} \\
TrackFlow   & 4.73 & 10.30 & 3.65 & 0.24 & 0.269 & 0.182 & 0.305 & 0.358 & 0.230 \\
DeFlow      & 3.43 & 7.32 & 2.51 & 0.46 & 0.276 & 0.113 & 0.228 & 0.496 & 0.266 \\
SSF         & 2.73 & 5.72 & 1.76 & 0.72 & 0.181 & 0.099 & 0.162 & 0.292 & 0.169 \\
Flow4D      & 2.24 & 4.94 & 1.31 & 0.47 & 0.145 & 0.087 & 0.150 & 0.216 & 0.127 \\
\(\Delta\)Flow  & 2.11 & \textbf{4.33} & 1.37 & 0.64 & \textbf{0.113} & 0.077 & \textbf{0.129} & \textbf{0.149} & 0.096 \\
SSF (\method, Ours) & 2.23 & 4.89 & 1.31 & 0.51 & 0.156 & 0.102 & 0.147 & 0.241 & 0.134 \\
Flow4D-XL (\method, Ours) & 2.07 & 4.50 & 1.31 & \textbf{0.40} & 0.132 & \textbf{0.072} & 0.131 & 0.218 & 0.107 \\
\(\Delta\)Flow (\method, Ours) & \textbf{2.07} & {4.41} & \textbf{1.26} & 0.54 & 0.118 & 0.073 & 0.136 & 0.164 & 0.098 \\
\bottomrule
\end{tabular}
}
\end{table}
}

{
\setlength{\tabcolsep}{0.85em}
\begin{table}[t!]
\centering
\caption{\textbf{Waymo Validation Set.} We compare UniFlow with recent methods on the Waymo val set. Notably, supervised methods achieve lower EPE than unsupervised methods, with \(\Delta\)Flow (\method) beating SeFlow by 62.0\% on Foreground Dynamic.}
\label{tab:waymo-val}
\resizebox{\linewidth}{!}{
\begin{tabular}{l|cccc|cccc}
\toprule
\multirow{2}{*}{\textbf{Methods}} & \multicolumn{4}{c|}{\textbf{Three-way EPE (cm) $\downarrow$}} & \multicolumn{4}{c}{\textbf{Dynamic Bucket-Normalized $\downarrow$}} \\
 & \textbf{Mean} & \textbf{FD} & \textbf{FS} & \textbf{BS} & \textbf{Mean} & \textbf{Car} & \textbf{Pedestr.} & \textbf{VRU} \\
\midrule
\rowcolor{gray!10}
\textit{Unsupervised} & \multicolumn{4}{c|}{} & \multicolumn{4}{c}{} \\
NSFP & 10.05 & 17.12 & 10.81 & 2.21 & 0.574 & 0.315 & 0.823 & 0.584 \\
ZeroFlow     & 8.64 & 22.40 & 1.57 & 1.96 & 0.770 & 0.444 & 0.982 & 0.884 \\
SeFlow & 8.50 & 20.81 & 2.14 & 2.57 & 0.328 & 0.305 & 0.485 & 0.185 \\
SeFlow++ & 3.63 & 9.30 & 0.87 & 0.71 & 0.232 & 0.201 & 0.521 & 0.247 \\
\midrule
\rowcolor{gray!10}
\textit{Supervised} & \multicolumn{4}{c|}{} & \multicolumn{4}{c}{} \\
DeFlow          & 4.46 & 9.80  & 2.59 & 0.98 & - & - & - & - \\
SSF             & 2.19 & 5.62  & 0.74 & 0.23 & 0.264 & 0.097 & 0.469 & 0.225 \\
Flow4D          & 1.82 & 4.58  & 0.61 & 0.28 & 0.215 & 0.074 & 0.423 & 0.146 \\
\(\Delta\)Flow  & 1.40 & 3.27 & 0.58 & 0.35 & 0.198 & 0.073 & 0.407 & 0.114 \\
SSF (\method, Ours)     & 1.85 & 4.74  & \textbf{0.58} & \textbf{0.22} & 0.234 & 0.075 & 0.434 & 0.195 \\
Flow4D-XL (\method, Ours)  & 1.60 & 3.83  & 0.68 & 0.30 & 0.191 & \textbf{0.064} & \textbf{0.400} & 0.110 \\
\(\Delta\)Flow (\method, Ours) & \textbf{1.38} & \textbf{3.07} &  0.61& 0.46 & \textbf{0.188} & 0.065 & 0.403 & \textbf{0.096} \\
\bottomrule
\end{tabular}
}
\end{table}
}

{
\setlength{\tabcolsep}{0.75em}
\begin{table}[t]
\centering
\caption{\textbf{nuScenes Validation Set.} We compare UniFlow with recent methods on the nuScenes val set. Interestingly, SSF (\method) outperforms Flow4D-XL (\method) by 26.53\% Dynamic Mean EPE, suggesting that SSF's architecture more effectively learns to estimate flow from sparse point clouds.}
\label{tab:nuscenes-val}
\resizebox{\linewidth}{!}{
\begin{tabular}{l|cccc|ccccc}
\toprule
\multirow{2}{*}{\textbf{Methods}} & \multicolumn{4}{c|}{\textbf{Three-way EPE (cm) $\downarrow$}} & \multicolumn{5}{c}{\textbf{Dynamic Bucket-Normalized $\downarrow$}} \\
 & \textbf{Mean} & \textbf{FD} & \textbf{FS} & \textbf{BS} & \textbf{Mean} & \textbf{Car} & \textbf{Other} & \textbf{Pedestr.} & \textbf{VRU} \\
\midrule
\rowcolor{gray!10}
\textit{Unsupervised} & \multicolumn{4}{c|}{} & \multicolumn{5}{c}{} \\
NSFP        & 10.79 & 20.26 & 4.88 & 7.23 & 0.602 & 0.463 & 0.456 & 0.829 & 0.662 \\
FastNSF        & 12.16 & 18.20 & 6.11 & 12.18 & 0.560 & 0.436 & 0.523 & 0.737 & 0.543 \\
SeFlow      & 8.19  & 16.15 & 3.97 & 4.45 & 0.554 & 0.396 & 0.635 & 0.726 & 0.419 \\
\midrule
\rowcolor{gray!10}
\textit{Supervised} & \multicolumn{4}{c|}{} & \multicolumn{5}{c}{} \\
DeFlow          & 3.98 & 6.99 & 3.45 & 1.50 & 0.314 & 0.163 & 0.286 & 0.533 & 0.275 \\
SSF             & 3.00 & 6.55 & 2.04 & 0.41 & 0.220 & 0.142 & 0.197 & 0.398 & 0.144 \\
Flow4D          & 3.46 & 7.97 & 1.85 & 0.55 & 0.230 & 0.160 & 0.241 & 0.345 & 0.176 \\
\(\Delta\)Flow  &2.33 & 4.83 & 1.37 & 0.79 & 0.216 & 0.138 & 0.219 & 0.327& 0.181\\
SSF (\method, Ours)     & 1.97 & 4.33 & 1.38 & \textbf{0.20} & 0.144 & \textbf{0.081} & \textbf{0.131} & {0.267} & {0.097} \\
Flow4D-XL (\method, Ours)  & 3.01 & 7.28 & 1.47 & 0.28 & 0.196 & 0.137 & 0.219 & 0.272 & 0.157 \\
\(\Delta\)Flow (\method, Ours) & \textbf{1.63} & \textbf{3.19} & \textbf{1.29} & 0.42 & \textbf{0.140} & 0.098 & 0.137 & \textbf{0.232} & \textbf{0.091} \\
\bottomrule
\end{tabular}
}
\end{table}
}

\begin{table}[t!]
\centering
\caption{\textbf{TruckScenes Validation Set.} We compare UniFlow with recent methods on the TruckScenes val set. According to three-way EPE, Flow4D achieves the best performance due to low Foreground Dynamic error. However, Dynamic Bucket Normalized EPE highlights that Flow4D-XL (\method) outperforms it across all moving object categories, reinforcing that three-way EPE is a biased metric \cite{khatri2024can}.}
\label{tab:truckscenes-val}
\resizebox{\linewidth}{!}{
\begin{tabular}{l|cccc|ccccc}
\toprule
\multirow{2}{*}{\textbf{Methods}} & \multicolumn{4}{c|}{\textbf{Three-way EPE (cm) $\downarrow$}} & \multicolumn{5}{c}{\textbf{Dynamic Bucket-Normalized $\downarrow$}} \\
 & \textbf{Mean} & \textbf{FD} & \textbf{FS} & \textbf{BS} & \textbf{Mean} & \textbf{Car} & \textbf{Other} & \textbf{Pedestr.} & \textbf{VRU} \\
\midrule
\rowcolor{gray!10}
\textit{Unsupervised} & \multicolumn{4}{c|}{} & \multicolumn{5}{c}{} \\
NSFP        & 45.63 & 120.45 & 2.07 & 14.38 & 0.658 & 0.303 & 0.350 & 1.221 & 0.758 \\
FastNSF     & 30.72 & 59.44  & 3.35 & 29.38 & 0.588 & 0.218 & 0.376 & 1.124 & 0.635 \\
SeFlow      & 37.41   & 103.97  & 1.56   & 6.68   & 0.681  & 0.494   & 0.752   & 0.886   & 0.591   \\
ICP Flow    & 58.82 & 169.91 & 1.43 & 5.12  & 0.472 & 0.302 & 0.614 & 0.596 & 0.376 \\
\midrule
\rowcolor{gray!10}
\textit{Supervised} & \multicolumn{4}{c|}{} & \multicolumn{5}{c}{} \\
DeFlow      & 7.30  & 16.47 & 1.67 & 3.77 & 0.570 & 0.180 & 0.410 & 0.970 & 0.730 \\
SSF         & 6.20  & 15.17  & \textbf{1.01}  & 2.44  & 0.453  & 0.215  & 0.308  & 0.875  & 0.414  \\
Flow4D      & 16.14 & 44.87 & 1.71 & 1.85 & 0.456 & 0.176 & 0.351 & 0.885 & 0.413 \\
\(\Delta\)Flow   & \textbf{7.28} & \textbf{16.26}  & 1.36  & 4.52  &  0.402 & 0.196  &  0.400 & 0.690  & 0.323 \\
\midrule
\rowcolor{gray!10}
\textit{Zero-shot} & \multicolumn{4}{c|}{} & \multicolumn{5}{c}{} \\
SSF (\method, Ours) & 35.23 & 103.72 & 1.69 & \textbf{0.27} & 0.435 & 0.149 & 0.455 & 0.669 & 0.466 \\
Flow4D-XL (\method, Ours) & 23.59 & 68.41 & 1.78 & 0.57 & \textbf{0.281} & \textbf{0.088} & \textbf{0.277} & 0.530 & 0.230 \\
\(\Delta\)Flow (\method, Ours) & {16.04} & 45.76 & 1.02 & 1.32 & 0.283 & 0.101 & 0.293 & \textbf{0.513} & \textbf{0.226} \\
\bottomrule
\end{tabular}
}
\end{table}

\begin{table}[t]
\centering
\caption{
\textbf{AEVAScenes Validation Set.} We compare UniFlow with recent methods on the AEVAScenes val set. Notably, due to the limited number of logs in the dataset, all methods are evaluated zero-shot. \method \ models consistently outperform single-dataset models.
}
\label{tab:aeva-zero-shot}
\resizebox{\linewidth}{!}{
\begin{tabular}{l|cccc|ccccc}
\toprule
\multirow{2}{*}{\textbf{Method}} &
\multicolumn{4}{c|}{\textbf{Three-way EPE (cm) $\downarrow$}} &
\multicolumn{5}{c}{\textbf{Dynamic Bucket-Norm. $\downarrow$}} \\
 & \textbf{Mean} & \textbf{FD} & \textbf{FS} & \textbf{BS} &
\textbf{Mean} & \textbf{Car} & \textbf{Other} & \textbf{Pedestr.} & \textbf{VRU} \\
\midrule
SSF (AV2) & 32.13 & 92.92 & 2.09 & 1.39 & 0.822 & 0.822 & 0.851 & 0.858 & 0.859 \\
SSF (Waymo) & 20.59 & 59.38 & 1.86 & 0.53 & 0.811 & 0.513 & 0.957 & 0.877 & 0.899 \\
SSF (nuScenes) & 31.04 & 88.68 & 3.08 & 1.36 & 0.772 & 0.689 & 0.865 & 0.775 & 0.759 \\
SSF (\method, Ours) & 10.80 & 29.76 & 2.05 & 0.60 & 0.544 & 0.239 & 0.686 & 0.612 & 0.639 \\
\midrule
Flow4D (AV2) & 16.13 & 43.20 & 3.19 & 2.01 & 0.419 & 0.317 & 0.434 & 0.491 & 0.433 \\
Flow4D (Waymo) & 8.14 & 18.89 & 4.03 & 1.50 & 0.408 & 0.152 & 0.560 & 0.491 & 0.560 \\
Flow4D (nuScenes) & 25.61 & 70.06 & 3.86 & 2.93 & 0.535 & 0.512 & 0.678 & 0.459 & 0.489 \\
Flow4D-XL (\method, Ours) & \textbf{6.14} & 13.16 & 3.89 & 1.37 & 0.338 & 0.114 & \textbf{0.312} & 0.480 & 0.448 \\
\midrule
\(\Delta\)Flow (AV2) & 15.63 & 43.37 & 2.73 & 0.79 & 0.444 & 0.302 & 0.560 & 0.460 & 0.453 \\
\(\Delta\)Flow (Waymo) & 7.01 & 16.27 & 3.59 & 1.19 & 0.391 & 0.123 & 0.648 & 0.426 & 0.426 \\
\(\Delta\)Flow (nuScenes) & 21.80 & 49.91 & 4.12 & 1.56 & 0.457 & 0.396 & 0.596 & 0.435 & 0.402 \\
\(\Delta\)Flow (\method, Ours) & 6.65 & 13.17 & 5.16 & 1.62 & \textbf{0.303} & \textbf{0.112} & 0.359 & \textbf{0.398} & \textbf{0.344} \\
\bottomrule
\end{tabular}
}
\end{table}

\textbf{Datasets.} 
We train and evaluate \method \ on five large-scale autonomous driving datasets to rigorously evaluate cross-domain generalization to diverse sensor configurations, ego-vehicle platforms, and driving scenarios (Table \ref{tab:dataset_stats}). AV2 and nuScenes collect data in dense urban environments with 32-beam LiDARs. Waymo uses a custom LiDAR sensor to collect denser point clouds from urban and suburban environments with more varied driving dynamics. We evaluate zero-shot generalization using TruckScenes~\cite{fent2024man} since it collects data in different environments (high-speed highway scenes vs. urban and suburban scenes), with different sensors (32-beam LiDARs vs. 64-beam LiDARs), using a different vehicle type (truck vs. sedan). We also evaluate zero-shot generalization using AEVAScenes since it uses a novel FMCW LiDAR. These five datasets allow us to systematically analyze the impact of point density (32-beam vs. 64-beam), ego-vehicle perspective (car vs. truck), and velocity (urban vs. highway) on cross-domain generalization.

\textbf{Metrics.}
We use \textit{Dynamic Bucket-Normalized EPE}~\cite{khatri2024can} as our primary metric to evaluate cross-domain generalization, particularly across diverse velocity distributions. This protocol is specifically designed to be both class-aware and speed-normalized. By normalizing motion estimation error by an object's speed, Dynamic Bucket-Normalized EPE measures the fraction of motion \textit{not} described, enabling fair comparison between slow-moving pedestrians and fast-moving vehicles. We also report three-way EPE~\cite{Chodosh_2024_WACV} for completeness, but note that it is a biased estimator of performance.

\textbf{Baselines.}
We compare UniFlow against both self-supervised and fully-supervised methods: NSFP~\cite{li2021neural}, FastNSF~\cite{li2023fast}, ICPFlow~\cite{lin2024icp}, SeFlow~\cite{zhang2024seflow}, SeFlow++~\cite{zhang2025himo}, EulerFlow~\cite{vedder2024neural}, VoteFlow~\cite{lin2025voteflow}, TrackFlow~\cite{khatri2024can}, DeFlow~\cite{zhang2024deflow}, SSF~\cite{khoche2025ssf}, Flow4D~\cite{kim2024flow4d} and \(\Delta\)Flow~\cite{zhang2025deltaflow}. 
We design a new model called Flow4D-XL, which simply adds more parameters to Flow4D to address model underfitting and improve its scaling behavior (Appendix \ref{sec:impl_details}). To ensure fairness, Argoverse 2 test-set results are obtained directly from the public leaderboard. We use Waymo, nuScenes and TruckScenes baselines as implemented in OpenSceneFlow\footnote{\url{https://github.com/KTH-RPL/OpenSceneFlow}} with standard training hyperparameters. We include additional implementation details in the supplement.

\textbf{Comparison to State-of-the-Art Methods.} We demonstrate the effectiveness of the \method~framework by retraining SSF, Flow4D, and \(\Delta\)Flow on multiple datasets and compare their performance against recent approaches on popular scene flow benchmarks (Tables~\ref{tab:av2}, \ref{tab:waymo-val}, and \ref{tab:nuscenes-val}). On the Argoverse 2 test set, Flow4D-XL (\method) improves over Flow4D by 9.0\% and SSF (\method) improves over SSF by 13.8\% in Dynamic Mean EPE. Moreover, Flow4D-XL (\method) achieves near-parity with EulerFlow at a fraction of the runtime. We see similar improvements between Flow4D and SSF and their \method \ counterparts on the Waymo (\Cref{tab:waymo-val}) and nuScenes (\Cref{tab:nuscenes-val}) validation sets. Interestingly, SSF (\method) outperforms Flow4D-XL (\method) on nuScenes, suggesting that the SSF architecture is naturally better suited for processing sparse point clouds.

\textbf{TruckScenes Zero-Shot Performance.} We demonstrate \method's strong zero-shot generalization capabilities with extensive evaluations on TruckScenes (Table~\ref{tab:truckscenes-val}). All baseline methods (except \method) were trained on TruckScenes. Despite never being explicitly trained on high speed highway scenes, Flow4D-XL~(\method) significantly outperforms the dataset-specific Flow4D on Dynamic Mean EPE. Notably, Flow4D-XL (\method) was only trained on urban driving scenes collected from a sedan, indicating that our approach can successfully generalize across significant domain gaps. Table~\ref{tab:truckscenes-val} also highlight the importance of the speed normalized and bucketed EPE metric \cite{khatri2024can}, as three-way EPE results don't adequately measure Flow4D-XL's (\method) significant improvements on pedestrian and VRU performance. We ablate this further in Appendix \ref{sec:truckscenes_breakdown}.

\begin{table}[t]
\centering
\caption{
\textbf{Ablation on Long-Range Generalization.}
All $^{\mathrm{LR}}$ results are obtained by evaluating models at extended voxel ranges while keeping the voxel size fixed and reusing the original trained weights. $^{\mathrm{ZS}}$ denotes zero-shot evaluation, where the model has never seen the target domain during training. We report Dynamic Mean EPE for each distance interval from the ego-vehicle. \(\Delta\)Flow (UniFlow) demonstrates strong zero-shot generalization to extended spatial ranges for both in- and out-of-domain datasets.
}
\label{tab:longrange}
\resizebox{\linewidth}{!}{
\begin{tabular}{l l l l l l l l}
\toprule
\rowcolor{gray!10}
Dataset & Method & Mean & 0--35 & 35--50 & 50--75 & 75--100 & 100--inf \\
\midrule
\multirow{3}{*}{AV2}
& \(\Delta\)Flow & 0.526 & 0.117 & 0.492 & 0.745 & 0.708 & 0.568 \\
& \(\Delta\)Flow$^{\mathrm{LR}}$ & 0.259 & 0.117 & 0.205 & 0.240 & 0.325 & 0.409 \\
& \(\Delta\)Flow$^{\mathrm{LR}}$ (\method, Ours) & 
\textbf{0.226}\tabblue{12.7\%} & 
0.114\tabblue{2.6\%} & 
0.198\tabblue{3.4\%} & 
0.219\tabblue{8.8\%} & 
0.260\tabblue{20.0\%} & 
0.339\tabblue{17.1\%} \\
\midrule
\multirow{3}{*}{Waymo}
& \(\Delta\)Flow & 0.379 & 0.055 & 0.455 & 0.628 & - & -  \\
& \(\Delta\)Flow$^{\mathrm{LR}}$ & 0.115 & 0.068 & 0.112 & 0.166 & - & - \\
& \(\Delta\)Flow$^{\mathrm{LR}}$ (\method, Ours) & 
\textbf{0.105}\tabblue{9.1\%} & 
0.063\tabblue{7.4\%} & 
0.104\tabblue{7.1\%} & 
0.149\tabblue{10.2\%} & 
- &
- \\
\midrule
\multirow{3}{*}{nuScenes}
& \(\Delta\)Flow & 0.389 & 0.172 & 0.380 & 0.633 & 0.606 & - \\
& \(\Delta\)Flow$^{\mathrm{LR}}$ & 0.267 & 0.181 & 0.242 & 0.318 & 0.447 & - \\
& \(\Delta\)Flow$^{\mathrm{LR}}$ (\method, Ours) & 
\textbf{0.248}\tabblue{7.1\%} & 
0.166\tabblue{7.9\%} & 
0.209\tabblue{13.6\%} & 
0.301\tabblue{5.3\%} & 
0.405\tabblue{9.4\%} & 
- \\
\midrule
\multirow{3}{*}{TruckScenes}
& \(\Delta\)Flow & 1.679 & 1.211 & 1.514 & 1.901 & 1.835 & 1.936 \\
& \(\Delta\)Flow$^{\mathrm{LR}}$ & 1.479 & 1.184 & 1.311 & 1.487 & 1.582 & 1.831 \\
& \(\Delta\)Flow$^{\mathrm{LR, ZS}}$ (\method, Ours) & 
\textbf{0.880}\tabblue{40.6\%} & 
0.786\tabblue{33.6\%} & 
0.826\tabblue{37.0\%} & 
0.694\tabblue{53.3\%} & 
0.779\tabblue{50.8\%} & 
1.317\tabblue{28.1\%} \\
\bottomrule
\end{tabular}
}
\end{table}

\begin{table}[t]
\centering
\caption{\textbf{Ablation on Semantic Cross-Domain Generalization.} We validate our claim that semantics are less generalizable than motion by retraining Flow4D-XL with an additional per-point semantic classification head. Notably, training Flow4D-XL on the unified dataset improves zero-shot scene flow estimation on TruckScenes, but hinders mIoU on out-of-distribution datasets. }
\label{tab:semantic-head}
\resizebox{\linewidth}{!}{
\begin{tabular}{cc|cc|c|c|ccccc}
\toprule
\multicolumn{2}{c|}{\textbf{Dataset}} &
\multicolumn{2}{c|}{\textbf{Scene Flow}$ \downarrow$} &
\multicolumn{6}{c}{\textbf{Semantic Classification} $\uparrow$} \\
\textbf{Train} & \textbf{Test} &
\textbf{Three-Way (cm)} & \textbf{Dyn. Norm.} &
\textbf{mIoU} & \textbf{Accuracy} & \textbf{CAR} & \textbf{OTHER} & \textbf{PED} & \textbf{VRU} \\
\midrule
AV2         & AV2         & 3.46  & 0.204 & 75.3\% & 93.9\% & 90.9\% & 86.2\% & 88.2\% & 36.0\% \\
Waymo       & Waymo       & 1.74  & 0.202 & 79.1\% & 98.9\% & 99.0\% & -  & 87.1\% & 51.3\% \\
nuScenes    & nuScenes    & 2.99  & 0.214 & 70.5\% & 89.8\% & 82.8\% & 83.3\% & 80.5\% & 35.4\% \\
TruckScenes & TruckScenes & 15.56 & 0.462 & 64.6\% & 95.3\% & 91.0\% & 92.9\% & 32.6\% & 42.1\%  \\
\midrule
Unified     & AV2         & 3.26  & 0.178 & 78.8\% & 94.2\% & 91.4\% & 85.8\% & 90.0\% & 47.9\% \\
Unified     & Waymo       & 1.57  & 0.190 & 64.0\% & 99.4\% & 99.4\% & 0.0\%  & 93.1\% & 63.5\% \\
Unified     & nuScenes    & 3.19  & 0.215 & 79.9\% & 93.4\% & 88.0\% & 89.1\% & 82.3\% & 60.3\% \\
Unified     & TruckScenes & 24.69 & 0.288 & 40.5\% & 66.1\% & 53.5\% & 46.4\% & 20.1\% & 41.9\% \\
\bottomrule
\end{tabular}}
\end{table}

\begin{table}[t]
\centering
\caption{\textbf{Ablation on TruckScenes.}
We ablate the impact of multi-dataset training, data augmentation, and model scaling on zero-shot TruckScenes performance.
All the model architectures we test benefit from multi-dataset training, while data augmentation continues to provide notable improvements even with 3$\times$ larger training sets, suggesting that scaling up \method \ may yield further improvements.}
\label{tab:ablation}
\resizebox{\linewidth}{!}{
\begin{tabular}{lllllll}
\toprule
\rowcolor{gray!10}
\textbf{Method} &
\textbf{FD (cm) $\downarrow$} & 
\textbf{Dynamic EPE $\downarrow$} & 
\textbf{[0, 0.5)} & 
\textbf{[0.5, 1.0)} & 
\textbf{[1.0, 2.0)} & 
\textbf{[2.0, $\infty$)} \\
\midrule
Flow4D (Waymo)
 & 116.01
 & 0.336
 & 0.071
 & 0.129
 & 0.215
 & 0.586 \\

\quad + Unified Dataset
 & 93.76  \tabblueabs{22.25}
 & 0.310  \tabblueabs{0.026}
 & 0.040  \tabblueabs{0.031}
 & 0.116  \tabblueabs{0.013}
 & 0.170  \tabblueabs{0.045}
 & 0.594  \tabredabs{0.008} \\

\quad + Augmentation
 & 68.32  \tabblueabs{25.44}
 & 0.301  \tabblueabs{0.009}
 & 0.047  \tabredabs{0.007}
 & 0.111  \tabblueabs{0.005}
 & 0.141  \tabblueabs{0.029}
 & 0.407  \tabblueabs{0.187} \\

\quad + XL Backbone
 & 68.41  \tabredabs{0.09}
 & 0.281  \tabblueabs{0.020}
 & 0.033  \tabblueabs{0.014}
 & 0.097  \tabblueabs{0.014}
 & 0.119  \tabblueabs{0.022}
 & 0.389  \tabblueabs{0.018} \\
\midrule

SSF (Waymo)
 & 170.65
 & 0.737
 & 0.049
 & 0.555
 & 0.740
 & 0.971 \\

\quad + Unified Dataset
 & 133.04  \tabblueabs{37.61}
 & 0.577  \tabblueabs{0.160}
 & 0.042  \tabblueabs{0.007}
 & 0.257  \tabblueabs{0.298}
 & 0.514  \tabblueabs{0.226}
 & 0.827  \tabblueabs{0.144} \\

\quad + Augmentation
 & 103.72  \tabblueabs{29.32}
 & 0.435  \tabblueabs{0.142}
 & 0.036  \tabblueabs{0.006}
 & 0.149  \tabblueabs{0.108}
 & 0.285  \tabblueabs{0.229}
 & 0.637  \tabblueabs{0.190} \\
\midrule

$\Delta$Flow (Waymo)
 & 117.83
 & 0.357
 & 0.067
 & 0.114
 & 0.302
 & 0.714 \\

\quad + Unified Dataset
 & 84.16  \tabblueabs{33.67}
 & 0.331  \tabblueabs{0.026}
 & 0.036  \tabblueabs{0.031}
 & 0.119  \tabredabs{0.005}
 & 0.268  \tabblueabs{0.034}
 & 0.659  \tabblueabs{0.055} \\

\quad + Augmentation
 & 16.04  \tabblueabs{68.40}
 & 0.283  \tabblueabs{0.048}
 & 0.039  \tabredabs{0.003}
 & 0.094  \tabblueabs{0.025}
 & 0.127  \tabblueabs{0.141}
 & 0.429  \tabblueabs{0.230} \\
\bottomrule
\end{tabular}
}
\end{table}

\begin{figure*}[t]
    \centering
    \includegraphics[width=\linewidth]{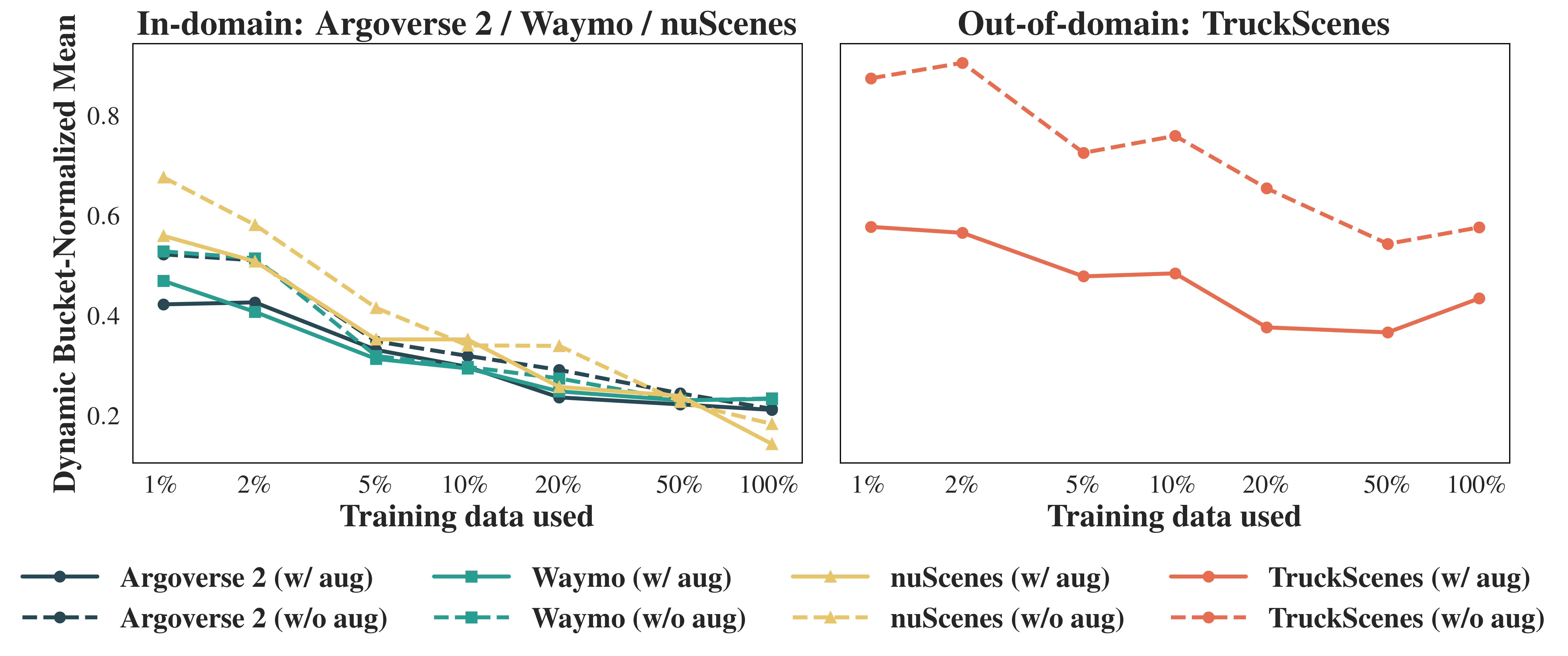}
    \caption{\textbf{Scaling Laws.} We evaluate both the in-distribution (on AV2, nuScenes, and Waymo) and out-of-distribution (on TruckScenes) performance of Flow4D-XL  (\method) with different amounts of training data. Unsurprisingly, increasing data reduces Dynamic Mean EPE. However, we find that data augmentation is significantly more important for out-of-distribution performance, and has minimal impact on in-distribution performance. Lower is better. }
    \label{fig:scaling}
\end{figure*}

\begin{table}[t!]
\centering
\caption{\textbf{Ablation on Frame-Rate.} We evaluate Flow4D-XL (\method),  SSF (\method) and \(\Delta\)Flow's (\method) generalization to different frame rates as a stress test outside standard operating conditions. Note that slower frame-rates effectively mimic a slower ego-vehicle, while faster-frame rates mimic a faster ego-vehicle. Although we train all models on 10 Hz annotations, multi-dataset training yields better results than single-dataset models, particularly at slower frame rates. Interestingly, faster frame rates do not negatively impact performance nearly as much as slower frame-rates.}
\label{tab:frame_rate}
\resizebox{\linewidth}{!}{
\begin{tabular}{l|cccc|ccccc}
\hline
\textbf{Method} & \multicolumn{4}{c|}{\textbf{Three-way EPE (cm)}} & \multicolumn{5}{c}{\textbf{Dynamic Bucket-Normalized}} \\
 & \textbf{Mean $\downarrow$} & \textbf{FD $\downarrow$} & \textbf{FS $\downarrow$} & \textbf{BS $\downarrow$} & \textbf{Dyn.~Mean $\downarrow$} & \textbf{Car $\downarrow$} & \textbf{Other $\downarrow$} & \textbf{Pedestr. $\downarrow$} & \textbf{VRU $\downarrow$} \\
\hline
\rowcolor{gray!10}
\multicolumn{10}{l}{\textit{2 Hz}} \\
Flow4D              & 30.79 & 88.10 & 3.26 & 1.02 & 0.495 & 0.364 & 0.494 & 0.613 & 0.509 \\
SSF                 & 31.16 & 89.82 & 3.07 & 0.59 & 0.520 & 0.363 & 0.498 & 0.684 & 0.535 \\
\(\Delta\)Flow   & 26.41 & 74.43 & 3.12 & 1.66 & 0.457 & 0.370 & 0.471 & 0.545 & 0.440 \\
Flow4D-XL (UniFlow, Ours) & {27.52} & {79.54} & \textbf{2.36} & 0.65 & {0.444} & 0.297 & \textbf{0.415} & {0.587} & \textbf{0.476} \\
SSF (UniFlow, Ours)   & 28.67 & 83.11 & 2.50 & \textbf{0.39} & 0.503 & \textbf{0.296} & 0.432 & 0.671 & 0.611 \\
\(\Delta\)Flow (\method, Ours)   & \textbf{23.87} & \textbf{68.26} & 2.46 & 0.88 & \textbf{0.398} & 0.306 & 0.420 & \textbf{0.494} & 0.494  \\
\hline
\rowcolor{gray!10}
\multicolumn{10}{l}{\textit{5 Hz}} \\
Flow4D              & 9.58 & 25.64 & 2.45 & 0.65 & 0.323 & 0.215 & 0.312 & 0.432 & 0.333 \\
SSF                 & 8.74 & 23.02 & 2.74 & 0.47 & 0.311 & 0.179 & 0.286 & 0.489 & 0.291 \\
\(\Delta\)Flow   & 6.43 & 16.54 & 1.98 & 0.76 & 0.264 & 0.173 & 0.271 & 0.369 & 0.242\\
Flow4D-XL (\method, Ours) & 8.02 & 21.78 & {1.92} & 0.37 & 0.290 & 0.190 & 0.265 & {0.405} & {0.302} \\
SSF (\method, Ours)   & {6.75} & {17.96} & 2.06 & \textbf{0.22} & {0.285} & \textbf{0.135} & \textbf{0.218} & 0.454 & 0.333 \\
\(\Delta\)Flow (\method, Ours)   & \textbf{5.26} & \textbf{13.44} & \textbf{1.81} & 0.58 & \textbf{0.232} & 0.150 & 0.220 & \textbf{0.343} & \textbf{0.214} \\
\hline
\rowcolor{gray!10}
\multicolumn{10}{l}{\textit{10 Hz (Standard Frame Rate)}} \\
Flow4D              & 3.46 & 7.97 & 1.85 & 0.55 & 0.230 & 0.160 & 0.241 & {0.241} & 0.176 \\
SSF                 & 3.00 & 6.55 & 2.04 & 0.41 & 0.220 & 0.142 & 0.197 & 0.398 & 0.144 \\
\(\Delta\)Flow   & 2.05 & 4.19 & 1.36 & 0.59 &  0.173 & 0.132 & 0.184 & 0.261 & 0.117 \\
Flow4D-XL (\method, Ours) & 3.01 & 7.28 & 1.47 & 0.28 & 0.196 & 0.137 & 0.219 & 0.272 & 0.157 \\
SSF (\method, Ours)   & {1.97} & {4.33} & {1.38} & \textbf{0.20} & {0.144} & \textbf{0.081} & \textbf{0.131} & 0.267 & {0.097} \\
\(\Delta\)Flow (\method, Ours)   & \textbf{1.63} & \textbf{3.18} & \textbf{1.29} & 0.42 & \textbf{0.140} & 0.098 & 0.137 & \textbf{0.232} & \textbf{0.091} \\
\hline
\rowcolor{gray!10}
\multicolumn{10}{l}{\textit{20 Hz}} \\
Flow4D              & 2.55 & 5.79 & 1.35 & 0.50 & 0.272 & 0.196 & 0.315 & 0.377 & 0.200 \\
SSF                 & 2.94 & 6.72 & 1.73 & 0.38 & 0.316 & 0.177 & 0.303 & 0.533 & 0.250 \\
\(\Delta\)Flow   & 1.60 & 3.28 & 0.48 & 0.48 & 0.220 & 0.167 & 0.231 & 0.327 & 0.156 \\
Flow4D-XL (\method, Ours) & {1.95} & {4.65} & {1.00} & {0.21} & {0.230} & {0.171} & {0.269} & {0.315} & {0.163} \\
SSF (\method, Ours)   & 2.13 & 5.16 & 1.11 & \textbf{0.12} & 0.246 & \textbf{0.127} & 0.275 & 0.395 & 0.189 \\
\(\Delta\)Flow (\method, Ours)   & \textbf{1.30} & \textbf{2.60} & \textbf{0.96} & 0.34 & \textbf{0.196} & 0.153 & \textbf{0.215} & \textbf{0.277} & \textbf{0.138} \\
\hline
\end{tabular}
}
\end{table}

\textbf{AEVAScenes Zero-Shot Performance. }
To further evaluate zero-shot generalization to unseen LiDARs, we benchmark our models on the AEVAScenes dataset~\footnote{\url{https://scenes.aeva.com/}}, which features high-resolution FMCW LiDARs that differs substantially from rotating LiDAR sensors used in existing benchmarks. Due to the limited scale of AEVAScenes, all methods are evaluated zero-shot.  
Notably, \method \ variants consistently outperform their respective best single-dataset counterparts on AEVAScenes, with SSF (\method) improving Dynamic Mean EPE by 33.8\%, Flow4D-XL (\method) by 17.2\%, and $\Delta$Flow (\method) by 22.5\%. We include visuals of zero-shot performance in Appendix \ref{sec:aeva_visuals}.

\textbf{Analysis of Long-Range Generalization.} 
To evaluate whether unified training produces range-robust representations, we evaluate how models trained on short-range LiDAR data transfer to longer-range settings \textit{without  retraining}. Specifically, we extend the voxelization range during inference while keeping the voxel size identical to the original trained model. This change alters only the spatial coverage of the dynamic voxelization module, requiring a simple weight transfer step (i.e., model weights trained for short-range scene flow are loaded into a model with identical voxel size but a larger voxel grid for long-range processing) to adapt to the new grid dimensions while preserving all learned features. As shown in Table~\ref{tab:longrange}, \(\Delta\)Flow$^{\mathrm{LR, ZS}}$ (\method) consistently outperforms \(\Delta\)Flow$^{\mathrm{LR}}$ across all distance intervals on AV2, 
reducing Dynamic Mean EPE by 12.7\% on average and up to 20.0\% in the 75--100\,m range. 
This indicates that unified multi-dataset training produces representations that remain robust even when the perception range is significantly expanded. On TruckScenes, \(\Delta\)Flow$^{\mathrm{LR, ZS}}$ (\method) achieves a remarkable 40.6\% reduction in overall error compared to \(\Delta\)Flow$^{\mathrm{LR}}$, with improvements exceeding 50\% in the 50--100\,m range, despite never being trained on this dataset. These results highlight that the representations learned through unified multi-dataset training naturally scale to larger spatial extents.

\textbf{Analysis of Semantic Cross-Domain Generalization.}
We evaluate the benefits of unified training on both geometric and semantic understanding by extending Flow4D-XL with a multi-task semantic head. This new Flow4D-XL variant simultaneously predicts scene flow and per-point semantics using the same backbone features. Since all datasets follow different taxonomies, we map their annotations to four coarse categories (e.g. {\tt car}, {\tt other}, {\tt pedestrian}, and {\tt VRU}) and jointly train the model with an additional class-balanced cross-entropy loss. As shown in Table~\ref{tab:semantic-head}, adding the semantic head has almost no effect on scene flow accuracy, indicating that the auxiliary task does not interfere with motion estimation. Consistent with our original findings, unified multi-dataset training continues to improve both in-domain and zero-shot performance for scene flow. However, training the semantic prediction head on the unified dataset yields mixed results. Models trained on unified datasets show small in-domain gains, but perform substantially worse than dataset-specific model when evaluated on unseen domains. This supports our hypothesis that dataset unification primarily benefits low-level geometric representations, while high-level semantics remain sensitive to dataset-specific biases.

\textbf{Analysis of Training Recipe.} We ablate the impact of multi-dataset training, data augmentation, and model scaling on zero-shot TruckScenes performance in Table \ref{tab:ablation}. All baselines are trained on Waymo only to ensure fair zero-shot evaluation. All model architectures show clear gains from multi-dataset training. Notably, despite training on a dataset three times larger, data augmentation continues to show substantial improvements, indicating that further scaling may yield additional performance gains.

\textbf{Analysis on Scaling Laws.} We evaluate Flow4D-XL (\method) with varying amounts of training data on both in-distribution benchmarks (AV2, nuScenes, and Waymo) and an out-of-distribution benchmark (TruckScenes) in Fig~\ref{fig:scaling}. As expected, larger training sets reduce Dynamic Mean EPE. However, data augmentation proves far more critical for out-of-distribution performance, while its effect on in-distribution performance remains minimal.

\textbf{Analysis on Frame-Rate.} We evaluate the generalization of Flow4D-XL (\method), SSF (\method) and \(\Delta\)Flow (\method) across different frame rates in Table~\ref{tab:frame_rate}. These settings serve as a stress test of temporal generalization outside standard operating conditions: slower frame rates approximate a slower-moving ego-vehicle, while faster frame rates approximate a faster one. Although all models are trained using 10 Hz annotations, multi-dataset training consistently improves performance over single-dataset models, especially at slower frame rates. Interestingly, higher frame rates have a smaller negative effect on performance, likely because models only need to estimate smaller displacement vectors for each timestep.


\section{Conclusion}
In this paper, we introduce \method, a frustratingly simple approach for scalable LiDAR scene flow that re-trains off-the-shelf models with diverse data from multiple datasets. Although prior work in LiDAR-based 3D object detection and segmentation suggest that naively training on multiple LiDAR datasets yields worse results than single-dataset trained models, we posit that ``low-level'' tasks such as motion estimation may be less sensitive to sensor configuration. Notably, our model establishes a new state-of-the-art on Waymo and nuScenes, improving over prior work by 5.1\% and 35.2\% respectively. Although \method is primarily trained on urban city driving scenarios and standard spinning LiDARs, we demonstrate that it generalizes surprisingly well to highway truck driving and novel FMCW sensors \textit{without any dataset-specific fine-tuning}. Our extensive analysis shows that such improvements are model-agnostic, suggesting that future scene flow methods should adopt this multi-dataset training strategy.

\clearpage
\bibliographystyle{splncs04}
\bibliography{main}

\clearpage
\newpage

\appendix

\section{Implementation Details}
\label{sec:impl_details}
\textbf{Models.} We train and evaluate SSF and $\Delta$Flow using their official implementations without modification. As shown in Table \ref{tab:ablation}, Flow4D benefits from a larger backbone when trained on the unified dataset. We denote this model as Flow4D-XL, which preserves the published architecture and increases all channel widths in the 4D Voxel Network by a uniform factor of three, with the Voxel Feature Encoders and Point Head adjusted accordingly to match the widened input/output dimensions. The full configuration is shown in Table~\ref{tab:flow4d_xl}. All dfataset-specific Flow4D models use the original backbone, while all Flow4D models trained on the unified dataset use Flow4D-XL unless otherwise noted. We train all models on 8x A6000 GPUs. Training takes 12 hours for baselines and 2 days for \method. Runtime evaluations are conducted on a single A6000 GPU. Inference latency is identical for both approaches. 

\begin{table}[h]
\centering
\caption{\textbf{Flow4D-XL 4D Voxel Network Architecture.} }
\label{tab:flow4d_xl}
\setlength{\tabcolsep}{10pt}
\resizebox{\linewidth}{!}{
\begin{tabular}{l c c c}
\hline
\rowcolor{gray!10}
\textbf{Stage} & \textbf{STDB-P (Filters)} & \textbf{Pool \& Up (Kernel, Stride)} & \textbf{Output shape \([W\times L\times H\times T\times ch]\)} \\
\hline
Input & -- & -- & \(512\times512\times32\times5\times \mathbf{48}\) \\
\hline
\multirow{2}{*}{1} & (48, 96) & \multirow{2}{*}{Pool (2,2,2,1)} & \multirow{2}{*}{\(256\times256\times16\times5\times \mathbf{96}\)} \\
 & (96, 96) & & \\
\hline
\multirow{2}{*}{2} & (96, 192) & \multirow{2}{*}{Pool (2,2,2,1)} & \multirow{2}{*}{\(128\times128\times 8\times5\times \mathbf{192}\)} \\
 & (192, 192) & & \\
\hline
\multirow{2}{*}{3} & (192, 192) & \multirow{2}{*}{Pool (2,2,2,1)} & \multirow{2}{*}{\(64\times64\times4\times5\times \mathbf{192}\)} \\
 & (192, 192) & & \\
\hline
\multirow{2}{*}{4} & (192, 192) & \multirow{2}{*}{Pool (2,2,1,1)} & \multirow{2}{*}{\(32\times32\times4\times5\times \mathbf{192}\)} \\
 & (192, 192) & & \\
\hline
\multirow{2}{*}{5} & (192, 192) & \multirow{2}{*}{Up (2,2,1,1)} & \multirow{2}{*}{\(32\times32\times4\times5\times \mathbf{192}\)} \\
 & (192, 192) & & \\
\hline
\multirow{2}{*}{6} & (192, 192) & \multirow{2}{*}{Up (2,2,2,1)} & \multirow{2}{*}{\(64\times64\times4\times5\times \mathbf{192}\)} \\
 & (192, 192) & & \\
\hline
\multirow{2}{*}{7} & (192, 192) & \multirow{2}{*}{Up (2,2,2,1)} & \multirow{2}{*}{\(128\times128\times8\times5\times \mathbf{192}\)} \\
 & (192, 192) & & \\
\hline
\multirow{2}{*}{8} & (192, 192) & \multirow{2}{*}{Up (2,2,2,1)} & \multirow{2}{*}{\(256\times256\times16\times5\times \mathbf{192}\)} \\
 & (192, 192) & & \\
\hline
9 & (96, 48) & -- & \(512\times512\times32\times5\times \mathbf{48}\) \\
\hline
\end{tabular}
}
\end{table}

\begin{table}[h!]
\centering
\caption{
\textbf{Ablation on Dataset Weighting Strategies.}
We re-train Flow4D-XL (\method) while varying the sampling frequency of data from each domain. Unified training consistently outperforms single-domain baselines, even if the in-domain dataset is only sampled 5\% of the time. Somewhat unsurprisingly, one can maximize in-domain performance by slightly biasing sampling towards in-domain data. 
}
\label{tab:weight_ablation}
\resizebox{\linewidth}{!}{
\begin{tabular}{l l ccccc}
\toprule
\textbf{Weight Strategy} & \textbf{Dataset Weights} & \textbf{Mean} & \textbf{FD} & \textbf{FS} & \textbf{BS} & \textbf{Dynamic M} \\
\midrule

\rowcolor{gray!10}\multicolumn{7}{l}{\textbf{TruckScenes}} \\
Proportional & 60\%W, 25\%A, 15\%N & \textbf{23.59} & \textbf{68.41} & 1.78 & 0.57 & \textbf{0.281} \\
Uniform & 33\%W, 33\%A, 33\%N & 30.44 & 89.01 & 1.73 & 0.58 & 0.307 \\
Waymo Heavy & 90\%W, 5\%A, 5\%N & 33.25 & 97.48 & 1.72 & \textbf{0.56} & 0.312 \\
AV2 Heavy & 90\%A, 5\%W, 5\%N & 29.82 & 87.08 & \textbf{1.70} & 0.67 & 0.310 \\
nuScenes Heavy & 90\%N, 5\%A, 5\%W & 31.26 & 91.11 & 1.70 & 0.96 & 0.332 \\
\midrule

\rowcolor{gray!10}\multicolumn{7}{l}{\textbf{Argoverse 2}} \\
Proportional & 60\%W, 25\%A, 15\%N & 3.24 & 7.83 & 1.26 & 0.63 & 0.176 \\
Uniform & 33\%W, 33\%A, 33\%N & \textbf{3.22} & \textbf{7.79} & \textbf{1.25} & {0.63} & \textbf{0.176} \\
Waymo Heavy & 90\%W, 5\%A, 5\%N & 3.28 & 7.87 & 1.28 & 0.69 & 0.189 \\
AV2 Heavy & 90\%A, 5\%W, 5\%N & {3.21} & 7.80 & 1.27 & \textbf{0.56} & 0.183 \\
nuScenes Heavy & 90\%N, 5\%A, 5\%W & 3.51 & 8.42 & 1.37 & 0.75 & 0.192 \\
AV2-only & 100\%A & 3.62 & 8.55 & 1.62 & 0.69 & 0.192 \\
\midrule

\rowcolor{gray!10}\multicolumn{7}{l}{\textbf{nuScenes}} \\
Proportional & 60\%W, 25\%A, 15\%N & 3.01 & 7.28 & 1.47 & 0.28 & 0.196 \\
Uniform & 33\%W, 33\%A, 33\%N & 2.07 & {4.70} & 1.34 & 0.16 & 0.134 \\
Waymo Heavy & 90\%W, 5\%A, 5\%N & 2.87 & 6.88 & 1.48 & 0.25 & 0.199 \\
AV2 Heavy & 90\%A, 5\%W, 5\%N & 2.93 & 6.93 & 1.55 & 0.30 & 0.222 \\
nuScenes Heavy & 90\%N, 5\%A, 5\%W & \textbf{1.75} & \textbf{3.81} & \textbf{1.29} & \textbf{0.15} & \textbf{0.109} \\
nuScenes-only & 100\%N & 3.46 & 7.97 & 1.85 & 0.55 & 0.230 \\
\midrule

\rowcolor{gray!10}\multicolumn{7}{l}{\textbf{Waymo}} \\
Proportional & 60\%W, 25\%A, 15\%N & 1.60 & 3.83 & 0.68 & 0.30 & \textbf{0.191} \\
Uniform & 33\%W, 33\%A, 33\%N & 1.59 & {3.89} & 0.61 & 0.27 & {0.195} \\
Waymo Heavy & 90\%W, 5\%A, 5\%N & \textbf{1.50} & \textbf{3.75} & \textbf{0.55} & \textbf{0.21} & 0.203 \\
AV2 Heavy & 90\%A, 5\%W, 5\%N & 1.81 & 4.41 & 0.70 & 0.32 & 0.213 \\
nuScenes Heavy & 90\%N, 5\%A, 5\%W & 1.90 & 4.76 & 0.61 & 0.32 & 0.215 \\
Waymo-only & 100\%W & 1.82 & 4.58 & 0.61 & 0.28 & 0.215 \\
\bottomrule
\end{tabular}}
\end{table}

\textbf{Datasets.} We retain all scans from Waymo, Argoverse 2, nuScenes, and TruckScenes when constructing our unified dataset. For AEVAScenes, we manually filter the dataset to remove scenes with missing packets or unreliable ground-truth annotations, retaining 65 valid scenes out of 100. We use dataset provided ego motion for motion compensation. 

\textbf{Augmentations.} We apply two dataset-agnostic augmentations. Following $\Delta$Flow, we apply height jitter with probability 0.8 by translating the point cloud along the sensor's z-axis using a uniform offset in the range of [-0.5, 2.0]m. To improve cross-dataset density generalization, we also apply LiDAR beam dropout with probability 0.35, removing every other LiDAR beam for all Waymo and Argoverse 2 sweeps to approximates a 32-beam sensor.

\textbf{Training Settings.}
We follow the standard evaluation protocol and report test-set performance for the Argoverse 2 benchmark using the official public leaderboard for fair comparison. For all other datasets, baseline methods and their \method\ variants are trained and evaluated on the prescribed train-val splits. Unless otherwise noted, \method\ models use the same architectures as their corresponding baselines, and all models are trained until convergence. \(\Delta\)Flow and its \method\ variants are trained on a \(76.8 \times 76.8\) m spatial grid with voxel resolution \((0.15, 0.15, 0.15)\). Flow4D is trained on a \(51.2 \times 51.2\) m spatial range with voxel resolution \((0.2, 0.2, 0.2)\), while SSF is trained on a \(51.2 \times 51.2\) m spatial range with voxel resolution \((0.2, 0.2, 6)\). For the long-range generalization experiments in Table~\ref{tab:longrange}, we extend all evaluated models to a \(153.6 \times 153.6\) m range for each model while keeping each model's voxel size fixed.

\section{Analysis on Dataset Weighting Strategies.}
\label{sec:weight_strats}

We further examine the impact of per-dataset sampling frequencies on our unified training approach. All models are trained on the unified dataset for a fixed number of iterations, but we vary the per-domain sampling frequency to control how often samples from each distribution are drawn. As shown in Table~\ref{tab:weight_ablation}, unified training consistently outperforms the single-dataset baselines, even under extreme weighting schemes. 
For example, on nuScenes, models trained with only 5\% exposure to the target domain and 90\% exposure to a very different distribution (Waymo) still surpass the in-domain baseline by 13.5\% on Dynamic Mean EPE, indicating that even limited cross-domain exposure provides considerable benefits. 
Unsurprisingly, one can maximize in-domain performance by slightly biasing the model twoards in-domain data. Lastly, we find that  uniformly sampling all datasets with equal probability yields the best out-of-distribution performance on TruckScenes.

\begin{table*}[!t]
\centering
\caption{
\textbf{Ablation on Synthetic Sparse LiDARs.}
We evaluate methods on synthetically sparsified LiDAR sweeps, created by removing every other beam from the original sensors. 
Colored numbers indicate the change relative to the original dense setting, with red showing degradation and blue showing improvement. Lower is better.
}
\label{tab:sparsity}
\resizebox{\linewidth}{!}{
\begin{tabular}{l|cccc|ccccc}
\toprule
\multirow{2}{*}{\textbf{Method}} &
\multicolumn{4}{c|}{\textbf{Three-way EPE (cm) $\downarrow$}} &
\multicolumn{5}{c}{\textbf{Dynamic Bucket-Normalized $\downarrow$}} \\
 &
\textbf{Mean} & \textbf{FD} & \textbf{FS} & \textbf{BS} &
\textbf{Dynamic M} & \textbf{CAR} & \textbf{OTHER} & \textbf{PED} & \textbf{VRU} \\
\midrule

\rowcolor{gray!10}
\multicolumn{10}{l}{\textit{Sparse AV2}} \\
SSF
  & 4.51  \tabredabs{0.44}
  & 10.79 \tabredabs{1.10}
  & 1.96  \tabredabs{0.21}
  & 0.80  \tabredabs{0.01}
  & 0.289 \tabredabs{0.041}
  & 0.183 \tabredabs{0.016}
  & 0.329 \tabredabs{0.061}
  & 0.336 \tabredabs{0.054}
  & 0.309 \tabredabs{0.032} \\
SSF (\method, Ours)
  & 3.73 \tabredabs{0.24}
  & 9.29 \tabredabs{0.62}
  & 1.38 \tabredabs{0.11}
  & 0.53 \tabredabs{0.01}
  & 0.240 \tabredabs{0.018}
  & 0.160 \tabredabs{0.009}
  & 0.168 \tabredabs{0.003}
  & 0.328 \tabredabs{0.040}
  & 0.304 \tabredabs{0.019} \\
Flow4D
  & 3.99 \tabredabs{0.37}
  & 9.58 \tabredabs{1.03}
  & 1.70 \tabredabs{0.08}
  & 0.69 \tabredabs{0.00}
  & 0.209 \tabredabs{0.017}
  & 0.158 \tabredabs{0.012}
  & 0.183 \tabredabs{0.019}
  & 0.238 \tabredabs{0.017}
  & 0.255 \tabredabs{0.020} \\
Flow4D-XL (\method, Ours)
  & 3.44 \tabredabs{0.20}
  & 8.27 \tabredabs{0.99}
  & 1.39 \tabredabs{0.13}
  & 0.65 \tabredabs{0.02}
  & 0.181 \tabredabs{0.005}
  & 0.138 \tabredabs{0.005}
  & 0.151 \tabredabs{0.002}
  & 0.198 \tabblueabs{0.005}
  & 0.237 \tabredabs{0.017} \\
\(\Delta\)Flow
  & 3.30 \tabredabs{0.26}
  & 8.00 \tabredabs{0.67}
  & 1.22 \tabredabs{0.09}
  & 0.67 \tabredabs{0.03}
  & 0.218 \tabredabs{0.042}
  & 0.149 \tabredabs{0.013}
  & 0.169 \tabredabs{0.008}
  & 0.230 \tabredabs{0.036}
  & 0.218 \tabredabs{0.005} \\
\(\Delta\)Flow (\method, Ours)
  & 3.19 \tabredabs{0.17}
  & 7.53 \tabredabs{0.40}
  & 1.25 \tabredabs{0.07}
  & 0.77 \tabredabs{0.03}
  & 0.182 \tabredabs{0.010}
  & 0.141 \tabredabs{0.008}
  & 0.157 \tabredabs{0.002}
  & 0.219 \tabredabs{0.025}
  & 0.213 \tabredabs{0.004} \\
\midrule

\rowcolor{gray!10}
\multicolumn{10}{l}{\textit{Sparse Waymo}} \\
SSF
  & 2.45 \tabredabs{0.24}
  & 6.12 \tabredabs{0.50}
  & 0.92 \tabredabs{0.18}
  & 0.30 \tabredabs{0.05}
  & 0.217 \tabblueabs{0.040}
  & 0.100 \tabredabs{0.003}
  & {-}
  & 0.342 \tabblueabs{0.127}
  & 0.210 \tabredabs{0.003} \\
SSF (\method, Ours)
  & 2.02 \tabredabs{0.17}
  & 5.22 \tabredabs{0.36}
  & 0.65 \tabredabs{0.10}
  & 0.18 \tabredabs{0.05}
  & 0.199 \tabblueabs{0.042}
  & 0.085 \tabredabs{0.004}
  & {-}
  & 0.318 \tabblueabs{0.139}
  & 0.193 \tabredabs{0.010} \\
Flow4D
  & 2.00 \tabredabs{0.18}
  & 4.92 \tabredabs{0.34}
  & 0.76 \tabredabs{0.15}
  & 0.33 \tabredabs{0.05}
  & 0.196 \tabblueabs{0.018}
  & 0.076 \tabredabs{0.001}
  & {-}
  & 0.363 \tabblueabs{0.059}
  & 0.150 \tabredabs{0.004} \\
Flow4D-XL (\method, Ours)
  & 1.66 \tabredabs{0.05}
  & 3.92 \tabredabs{0.09}
  & 0.73 \tabredabs{0.05}
  & 0.31 \tabblueabs{0.02}
  & 0.143 \tabblueabs{0.046}
  & 0.063 \tabblueabs{0.001}
  & {-}
  & 0.249 \tabblueabs{0.151}
  & 0.119 \tabredabs{0.015} \\
\(\Delta\)Flow
  & 1.53 \tabredabs{0.09}
  & 3.42 \tabredabs{0.12}
  & 0.68 \tabredabs{0.08}
  & 0.50 \tabredabs{0.09}
  & 0.146 \tabblueabs{0.050}
  & 0.072 \tabredabs{0.002}
  & {-}
  & 0.257 \tabblueabs{0.154}
  & 0.107 \tabredabs{0.000} \\
\(\Delta\)Flow (\method, Ours)
  & 1.43 \tabredabs{0.05}
  & 3.12 \tabredabs{0.05}
  & 0.66 \tabredabs{0.05}
  & 0.52 \tabredabs{0.06}
  & 0.138 \tabblueabs{0.050}
  & 0.066 \tabredabs{0.001}
  & {-}
  & 0.246 \tabblueabs{0.157}
  & 0.102 \tabredabs{0.006} \\
\midrule

\rowcolor{gray!10}
\multicolumn{10}{l}{\textit{Sparse nuScenes}} \\
SSF
  & 3.96 \tabredabs{0.96}
  & 9.24 \tabredabs{2.69}
  & 2.18 \tabredabs{0.14}
  & 0.45 \tabredabs{0.05}
  & 0.298 \tabredabs{0.078}
  & 0.173 \tabredabs{0.031}
  & 0.223 \tabredabs{0.027}
  & 0.556 \tabredabs{0.158}
  & 0.241 \tabredabs{0.096} \\
SSF (\method, Ours)
  & 2.94 \tabredabs{1.05}
  & 7.11 \tabredabs{3.01}
  & 1.56 \tabredabs{0.13}
  & 0.14 \tabredabs{0.01}
  & 0.222 \tabredabs{0.089}
  & 0.094 \tabredabs{0.013}
  & 0.165 \tabredabs{0.034}
  & 0.454 \tabredabs{0.225}
  & 0.173 \tabredabs{0.083} \\
Flow4D
  & 4.31 \tabredabs{0.86}
  & 10.41 \tabredabs{2.44}
  & 1.92 \tabredabs{0.07}
  & 0.61 \tabredabs{0.06}
  & 0.309 \tabredabs{0.078}
  & 0.227 \tabredabs{0.067}
  & 0.283 \tabredabs{0.042}
  & 0.439 \tabredabs{0.094}
  & 0.287 \tabredabs{0.056} \\
Flow4D-XL (\method, Ours)
  & 3.72 \tabredabs{0.71}
  & 9.31 \tabredabs{2.03}
  & 1.52 \tabredabs{0.05}
  & 0.31 \tabredabs{0.03}
  & 0.279 \tabredabs{0.083}
  & 0.140 \tabredabs{0.003}
  & 0.266 \tabredabs{0.047}
  & 0.430 \tabredabs{0.158}
  & 0.279 \tabredabs{0.122} \\
\(\Delta\)Flow
  & 2.76 \tabredabs{0.71}
  & 6.06 \tabredabs{1.87}
  & 1.46 \tabredabs{0.10}
  & 0.75 \tabredabs{0.16}
  & 0.255 \tabredabs{0.081}
  & 0.179 \tabredabs{0.047}
  & 0.217 \tabredabs{0.033}
  & 0.409 \tabredabs{0.148}
  & 0.214 \tabredabs{0.097} \\
\(\Delta\)Flow (\method, Ours)
  & 2.25 \tabredabs{0.62}
  & 4.81 \tabredabs{1.63}
  & 1.43 \tabredabs{0.14}
  & 0.51 \tabredabs{0.09}
  & 0.206 \tabredabs{0.067}
  & 0.126 \tabredabs{0.028}
  & 0.170 \tabredabs{0.033}
  & 0.365 \tabredabs{0.133}
  & 0.165 \tabredabs{0.074} \\
\bottomrule

\end{tabular}
}
\end{table*}

\section{Ablation on Synthetic Sparse LiDARs}

We stress-test the performance of \method \ models on synthetically sparsified LiDAR point clouds to investigate the impact of LiDAR sparsity on cross-domain generalization. We do not train any new models; all results are from models trained on the original datasets. Sparse datasets are created by removing every other beam from each LiDAR sweep. As shown in Table~\ref{tab:sparsity}, the colored deltas indicate the change in performance relative to each model's performance on the standard datasets (higher EPE indicates worse performance). Across datasets, degradation correlates strongly with dataset sparsity: models evaluated on sparse nuScenes perform the worst due to its low (16-beam) density, while models evaluated on sparse Waymo are affected the least due to its substantially higher point density. Notably, \method\ models are consistently more robust than dataset-specific baselines, exhibiting smaller performance drops across all datasets. This suggests that multi-dataset training provides useful regularization even under extreme sparsity. 

\begin{figure*}[t!]
  \centering
  \includegraphics[width=0.98\linewidth]{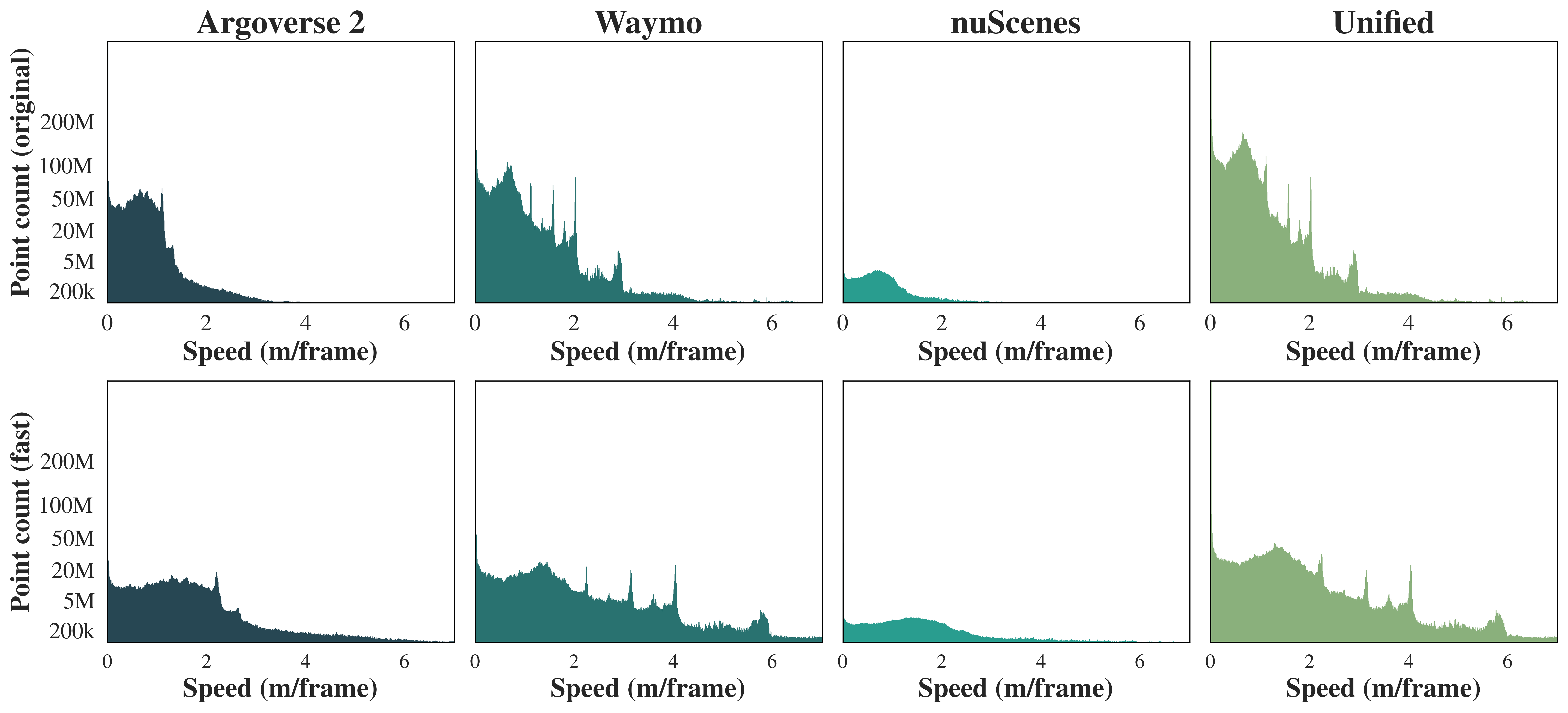}
  \caption{
    \textbf{Comparing Original and Downsampled Velocity Distributions.}
    We plot the velocity distributions of the original AV2, Waymo, and nuScenes training sets (top row), their corresponding down-sampled ``fast'' versions (bottom row), and the unified distribution that combine the three datasets (right column).
    }
  \label{fig:velocity_fast_unified}
\end{figure*}

\begin{table*}[t!]
\centering
\setlength{\tabcolsep}{0.8em}
\caption{\textbf{Velocity Augmentation.}
We perform velocity augmentation by adding fast versions of each dataset to the unified dataset, and compare the results to Flow4D and Flow4D-XL (\method) trained on the original 10\,Hz data. We denote zero-shot evaluation with $^{\mathrm{ZS}}$ for cases where the model has never seen the target domain during training.}
\label{tab:velocity-ablation}
\resizebox{\linewidth}{!}{
\begin{tabular}{l|ccccc|cccc}
\toprule
\multirow{2}{*}{\textbf{Method}} &
\multicolumn{5}{c|}{\textbf{Dynamic Bucket-Normalized $\downarrow$}} &
\multicolumn{4}{c}{\textbf{Speed Bucket-Normalized $\downarrow$}} \\
 & \textbf{Dynamic M} & \textbf{CAR} & \textbf{OTHER} & \textbf{PED} & \textbf{VRU}
 & \textbf{[0,0.5]} & \textbf{(0.5,1.0]} & \textbf{(1.0,2.0]} & \textbf{(2.0,$\infty$]} \\
\midrule
\rowcolor{gray!10}
\multicolumn{10}{l}{\textit{Argoverse 2}} \\
Flow4D       & 0.192 & 0.146 & 0.164 & 0.221 & 0.236 & 0.020 & 0.119 & 0.114 & 0.103 \\
Flow4D (\method, Ours)      & 0.176 & 0.133 & 0.149 & 0.203 & 0.220 & \textbf{0.018} & \textbf{0.107} & \textbf{0.108} & 0.091 \\
Flow4D (Fast-\method, Ours) & 0.179 & 0.134 & 0.154 & 0.204 & 0.224 & 0.019 & 0.114 & 0.111 & \textbf{0.085} \\
\midrule
\rowcolor{gray!10}
\multicolumn{10}{l}{\textit{Waymo}} \\
Flow4D       & 0.214 & 0.074 & {-}   & 0.423 & 0.146 & 0.007 & 0.064 & 0.051 & 0.130 \\
Flow4D-XL (\method, Ours)      & 0.191 & 0.064 & {-}   & 0.400 & 0.110 & \textbf{0.006} & \textbf{0.053} & \textbf{0.042} & 0.128 \\
Flow4D-XL (Fast-\method, Ours) & 0.208 & 0.066 & {-}   & 0.409 & 0.149 & 0.007 & 0.058 & 0.043 & \textbf{0.093} \\
\midrule

\rowcolor{gray!10}
\multicolumn{10}{l}{\textit{nuScenes}} \\
Flow4D       & 0.230 & 0.160 & 0.241 & 0.345 & 0.176 & 0.013 & 0.100 & 0.109 & 0.370 \\
Flow4D-XL (\method, Ours)      & 0.196 & 0.137 & 0.219 & 0.272 & 0.157 & \textbf{0.009} & \textbf{0.082} & \textbf{0.089} & 0.374 \\
Flow4D-XL (Fast-\method, Ours)  & 0.187 & 0.126 & 0.217 & 0.263 & 0.143 & 0.009 & 0.090 & 0.101 & \textbf{0.306} \\
\midrule

\rowcolor{gray!10}
\multicolumn{10}{l}{\textit{TruckScenes}} \\
Flow4D       & 0.456 & 0.176 & 0.351 & 0.885 & 0.413 & 0.054 & 0.192 & 0.171 & \textbf{0.091} \\
Flow4D-XL (\method, Ours) $^{\mathrm{ZS}}$     & 0.281 & 0.088 & 0.277 & 0.530 & 0.230 & \textbf{0.034} & \textbf{0.097} & 0.119 & 0.389 \\
Flow4D-XL (Fast-\method, Ours)$^{\mathrm{ZS}}$& 0.281 & 0.079 & 0.264 & 0.546 & 0.235 & 0.034 & 0.112 & \textbf{0.106} & 0.323 \\
\bottomrule
\end{tabular}
}
\end{table*}

\section{Ablation on Velocity Augmentation}
\label{sec:velocity_aug}

Our velocity-bucket analysis in Figure \ref{fig:velocity} suggests that the velocity domain gap between training datasets is highly correlated with cross-domain generalization. To test whether increasing the proportion of fast motion improves flow estimation on fast moving objects across datasets, we create a ``fast'' version of each dataset by down-sampling each dataset from 10Hz to 5Hz and add these sequences to the unified training set. This effectively doubles the speed of all dynamic objects and shifts the velocity distribution to the right (Figure~\ref{fig:velocity_fast_unified}). As shown in Table~\ref{tab:velocity-ablation}, this improves performance in the highest-speed bucket $(2.0,\infty)$, but consistently degrades accuracy for all other velocity buckets. We attribute this to the aggressiveness of the augmentation: doubling the speed of every dynamic object produces unrealistic trajectories that no longer resemble any real object motion. Although the velocity domain gap significantly contributes to cross-domain generalization, naively down-sampling datasets to increase the proportion of fast moving objects does not improve performance. Future work should consider different strategies for effective velocity augmentation.

\begin{table*}[!t]
\centering
\caption{
\textbf{Performance Breakdown on TruckScenes.}
Three-way EPE heavily penalizes fast-moving objects and is biased towards fast-moving highway scenes. In contrast, Dynamic Bucket-Normalized EPE provides a more fair comparison across object categories and speeds.
}
\label{tab:truckscenes-breakdown}
\resizebox{\linewidth}{!}{
\begin{tabular}{l|cccc|ccccc|cccc}
\toprule
\multirow{2}{*}{\textbf{Method}} &
\multicolumn{4}{c|}{\textbf{Three-way EPE (cm) $\downarrow$}} &
\multicolumn{5}{c|}{\textbf{Dynamic Bucket-Norm. $\downarrow$}} &
\multicolumn{4}{c}{\textbf{Speed Bucket-Norm. $\downarrow$}} \\
 & \textbf{Mean} & \textbf{FD} & \textbf{FS} & \textbf{BS} &
\textbf{Dyn. M} & \textbf{CAR} & \textbf{OTHER} & \textbf{PED} & \textbf{VRU} &
\textbf{[0,0.5]} & \textbf{(0.5,1.0]} & \textbf{(1.0,2.0]} & \textbf{(2.0,$\infty$]} \\
\midrule
SSF &
6.20 & 15.17 & 1.01 & 2.44 &
0.453 & 0.215 & 0.308 & 0.875 & 0.414 &
0.055 & 0.190 & 0.173 & 0.086 \\

SSF (\method, Ours) &
35.23 & 103.72 & 1.69 & 0.27 &
0.435 & 0.149 & 0.455 & 0.669 & 0.466 &
0.036 & 0.148 & 0.285 & 0.637 \\

Flow4D &
16.14 & 44.87 & 1.71 & 1.85 &
0.456 & 0.176 & 0.351 & 0.885 & 0.413 &
0.027 & 0.249 & 0.157 & 0.079 \\

Flow4D-XL (\method, Ours) &
23.59 & 68.41 & 1.78 & 0.57 &
0.281 & 0.088 & 0.277 & 0.530 & 0.230 &
0.034 & 0.097 & 0.119 & 0.389 \\

\(\Delta\)Flow &
7.28 & 16.26 & 1.36 & 4.52 &
0.402 & 0.196 & 0.400 & 0.690 & 0.323 &
0.079 & 0.222 & 0.203 & 0.102 \\

\(\Delta\)Flow (\method, Ours) &
15.76 & 45.76 & 1.02 & 1.32 &
0.283 & 0.101 & 0.293 & 0.513 & 0.226 &
0.039 & 0.094 & 0.127 & 0.429 \\
\bottomrule
\end{tabular}
}
\end{table*}

\section{Performance Breakdown on TruckScenes}
\label{sec:truckscenes_breakdown}
We present a detailed breakdown between three-way EPE and dynamic bucket-normalized EPE performance on TruckScenes to investigate the inconsistent trends found in the main paper. As demonstrated by Khatri et. al. \cite{khatri2024can}, three-way EPE directly measures absolute endpoint error and therefore scales with the magnitude of the ground-truth flow. As a result, underestimating large displacement vectors from fast moving objects in highway scenes incurs disproportionately high penalties, inflating three-way EPE for TruckScenes. In contrast, dynamic bucket-normalized EPE offers a more balanced assessment of model predictions by evaluating error relative to object speed. As shown in Table~\ref{tab:truckscenes-breakdown}, \method \ achieves strong zero-shot generalization compared to in-domain models across common-speed buckets (0 to 2.0 m/frame), indicating that \method \ effectively models the motion patterns dominating most scenes. We find that \method \ models perform worse than in-domain models in the high-speed regime ($>2.0$ m/frame), suggesting that future work should explicitly address high-speed scene flow. 

\section{Impact of Model Size vs. Dataset Size}
\label{sec:mdoel_size_vs_data_size}

In Table \ref{tab:ablation}, we found that scaling up Flow4D's backbone and scaling up training data considerably improved out-of-domain performance on TruckScenes. We explicitly disentangle the impact of model size and dataset size on both in-domain and out-of-domain performance in Table \ref{tab:model_data_size}. We find that training Flow4D on the unified dataset improves performance on all datasets except nuScenes. We note that training Flow4D-XL with the standard training recipe yields worse performance than Flow4D on AV2 and nuScenes. Intuitively, dramatically increasing model capacity without increasing training data size leads to underfitting. Lastly, we find that scaling up model size and training data achieves the best performance, improving performance across all in-domain and out-of-domain datasets. 

\begin{table}[!t]
\centering
\caption{\textbf{Ablation on Model Size vs. Dataset Size.} We evaluate the impact of scaling model size and dataset size on Flow4D. Training Flow4D with more data yields improvements on all datasets except nuScenes. Increasing Flow4D's model capacity and training data size yields improved performance across all datasets. 
}
\label{tab:model_data_size}
\resizebox{\linewidth}{!}{
\begin{tabular}{l|ccccc}
\toprule
\multirow{2}{*}{\textbf{Methods}} &  \multicolumn{5}{c}{\textbf{Dynamic Bucket-Normalized $\downarrow$}} \\
& \textbf{\# Params.} & \textbf{AV2} & \textbf{Waymo} & \textbf{nuScenes} & \textbf{TruckScenes$^{ZS}$} \\
\midrule
Flow4D  & 4.6M  & 0.192 & 0.215 & 0.230 & 0.456  \\
Flow4D (\method, Ours)  & 4.6M & 0.189  & 0.204 & 0.257 & 0.301 \\
Flow4D-XL  & 41.4M & 0.196  & 0.210 & 0.249 & 0.416 \\
Flow4D-XL (\method, Ours) & 41.4M  & 0.176 & 0.191 & 0.196 & 0.281 \\
\bottomrule
\end{tabular}
}
\end{table}

\begin{table}[b]
\centering
\caption{\textbf{Comparison with Domain Adaptation Baselines.}
We compare augmentation-based domain adaptation variants of dataset-specific \(\Delta\)Flow models against \(\Delta\)Flow trained with our unified multi-dataset pipeline on TruckScenes zero-shot evaluation. \(\Delta\)Flow (\method) achieves the best overall Dynamic Bucket-Normalized EPE, outperforming the best DA baseline by 14.5\%. Lower is better.}
\label{tab:da_baselines}
\resizebox{\linewidth}{!}{
\begin{tabular}{l|ccccc}
\toprule
\multirow{2}{*}{\textbf{TruckScenes Zero-Shot}} &
\multicolumn{5}{c}{\textbf{Dynamic Bucket-Norm. $\downarrow$}} \\
& \textbf{Dyn. M} & \textbf{Car} & \textbf{Other} & \textbf{Pedestr.} & \textbf{VRU} \\
\midrule
\(\Delta\)Flow (AV2) + Domain Adaptation & 0.528 & 0.366 & 0.520 & 0.555 & 0.528 \\
\(\Delta\)Flow (Waymo) + Domain Adaptation & 0.331 & 0.157 & 0.411 & 0.549 & \textbf{0.207} \\
\(\Delta\)Flow (nuScenes) + Domain Adaptation & 0.667 & 0.533 & 0.730 & 0.709 & 0.696 \\
\(\Delta\)Flow (\method, Ours) & \textbf{0.283} & \textbf{0.101} & \textbf{0.293} & \textbf{0.513} & 0.226 \\
\bottomrule
\end{tabular}
}
\end{table}

\section{Comparison with Domain Adaptation Baselines}
\label{sec:da_baselines}

We additionally compare \method\ against domain adaptation baselines on TruckScenes zero-shot evaluation. Specifically, we apply the same data augmentation strategies to dataset-specific \(\Delta\)Flow models trained on AV2, Waymo, and nuScenes, and compare them against \(\Delta\)Flow trained with our unified multi-dataset pipeline. As shown in Table~\ref{tab:da_baselines}, \(\Delta\)Flow (\method) achieves the best overall performance, outperforming the best domain adaptation baseline by 14.5\% on Dynamic Bucket-Normalized EPE. These results suggest that unified multi-dataset training is a particularly effective strategy for improving cross-domain generalization.

\section{Zero-Shot Visualizations on AEVAScenes}
\label{sec:aeva_visuals}

\begin{figure}[t]
    \centering
    \setlength{\tabcolsep}{1.5pt}
    \begin{tabular}{@{}c@{\hspace{0.4em}}c@{}}
        \figtwosidecaption{Scene I}{} &
        \begin{minipage}{0.90\linewidth}\centering
            \begin{subfigure}{0.32\linewidth}
                \includegraphics[width=\linewidth]{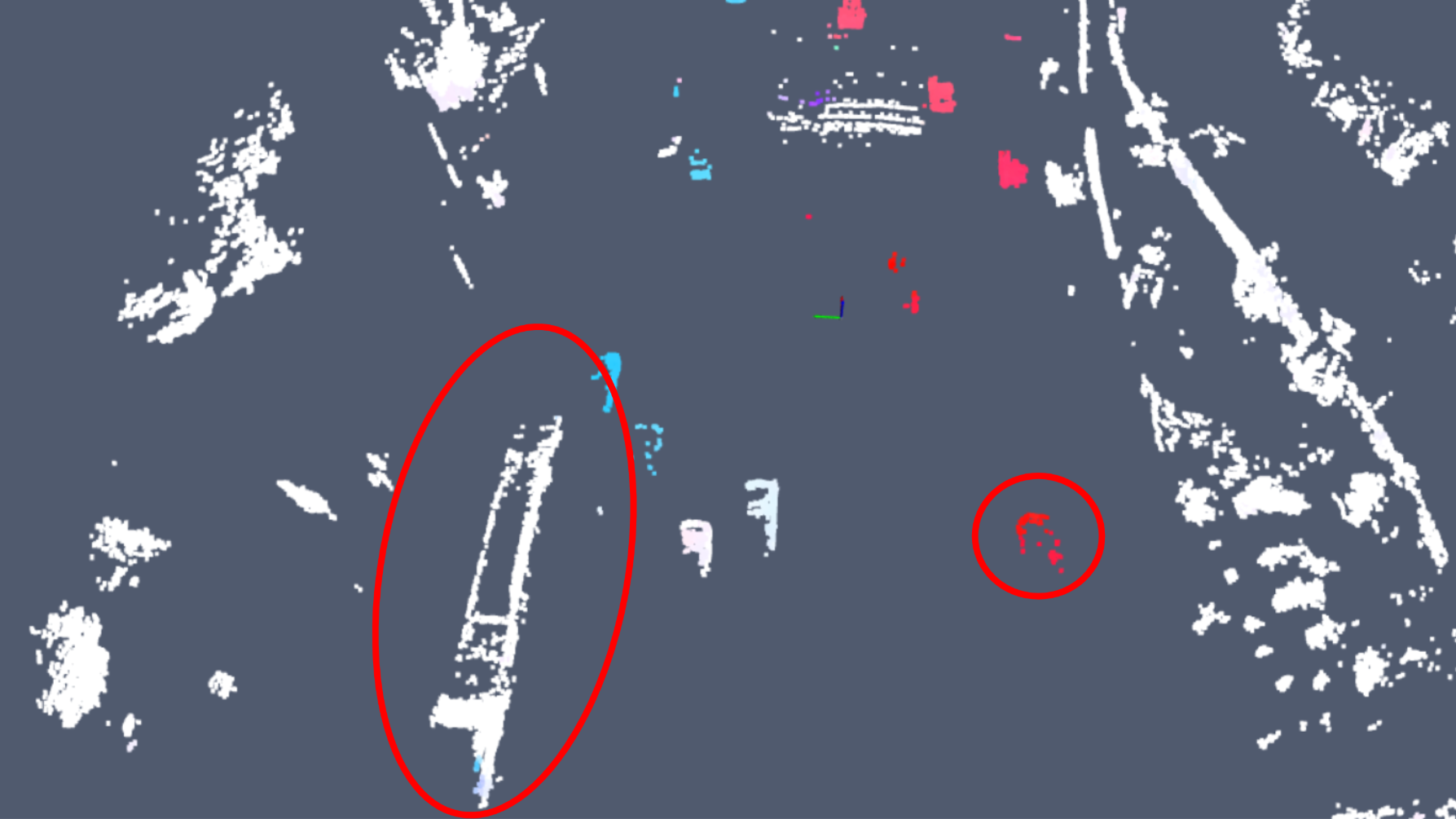}
            \end{subfigure}\hfill
            \begin{subfigure}{0.32\linewidth}
                \includegraphics[width=\linewidth]{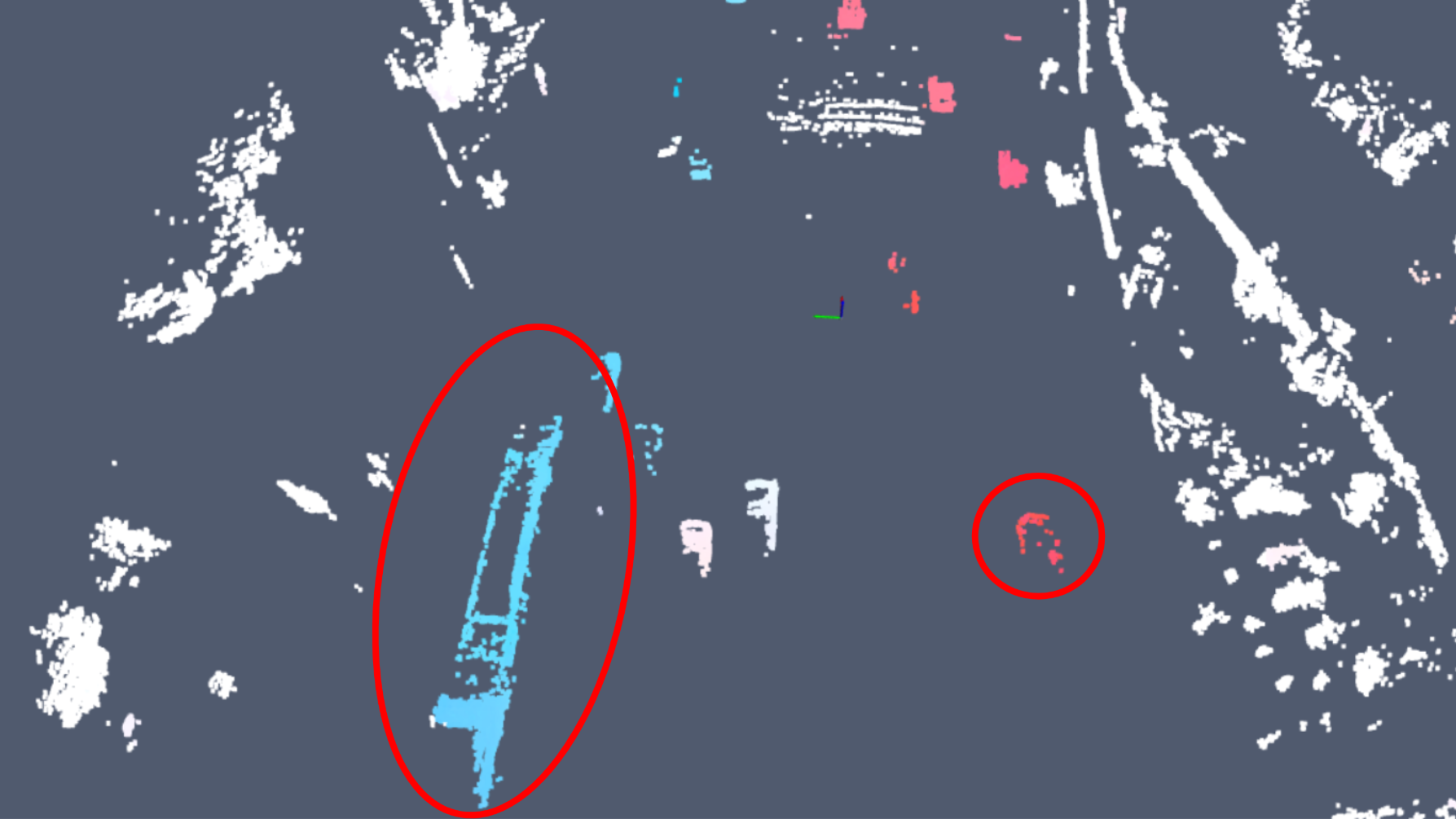}
            \end{subfigure}\hfill
            \begin{subfigure}{0.32\linewidth}
                \includegraphics[width=\linewidth]{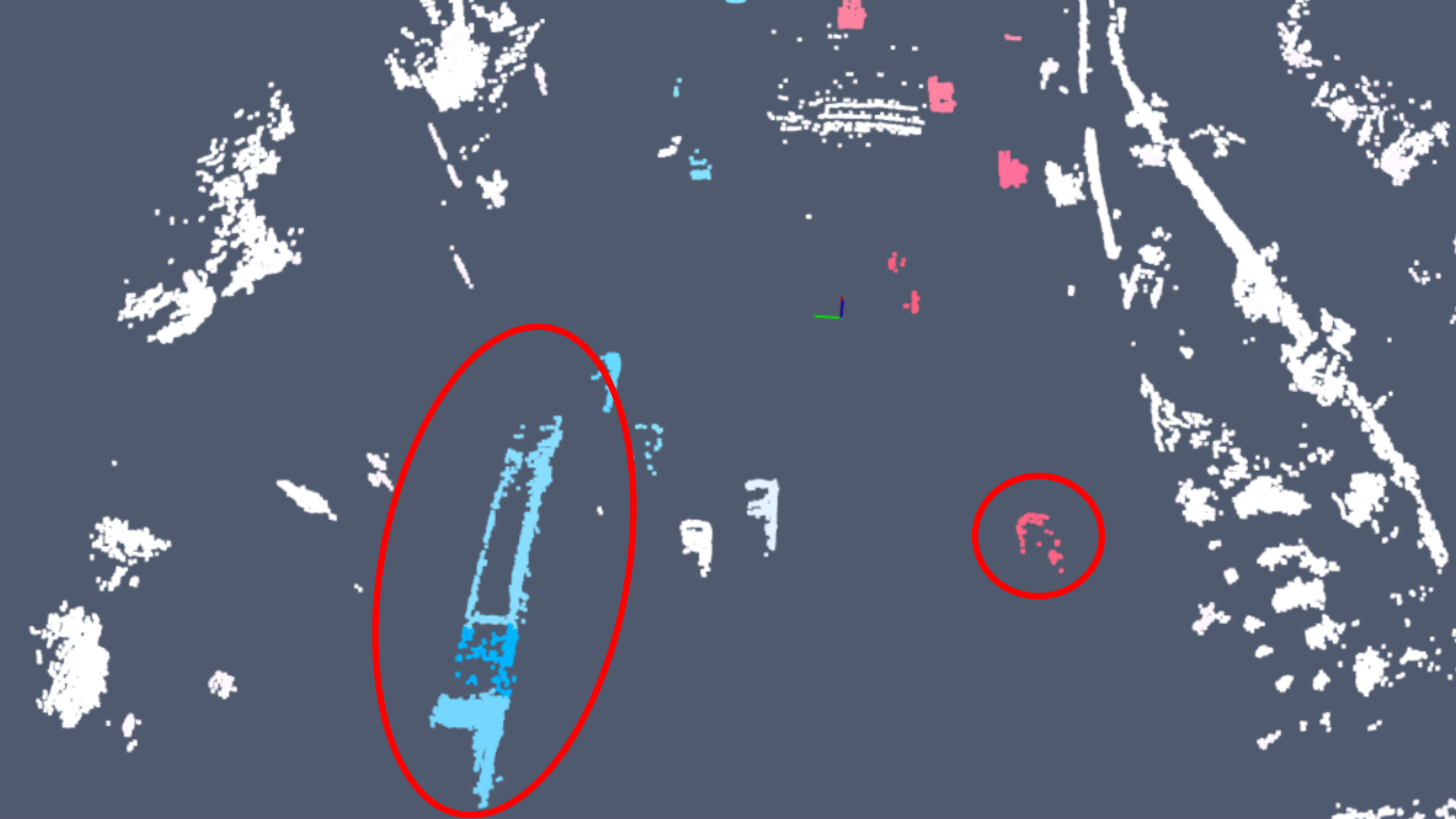}
            \end{subfigure}
        \end{minipage} \\
        \noalign{\vskip 0.4em}
        \figtwosidecaption{Scene II}{} &
        \begin{minipage}{0.90\linewidth}\centering
            \begin{subfigure}{0.32\linewidth}
                \includegraphics[width=\linewidth]{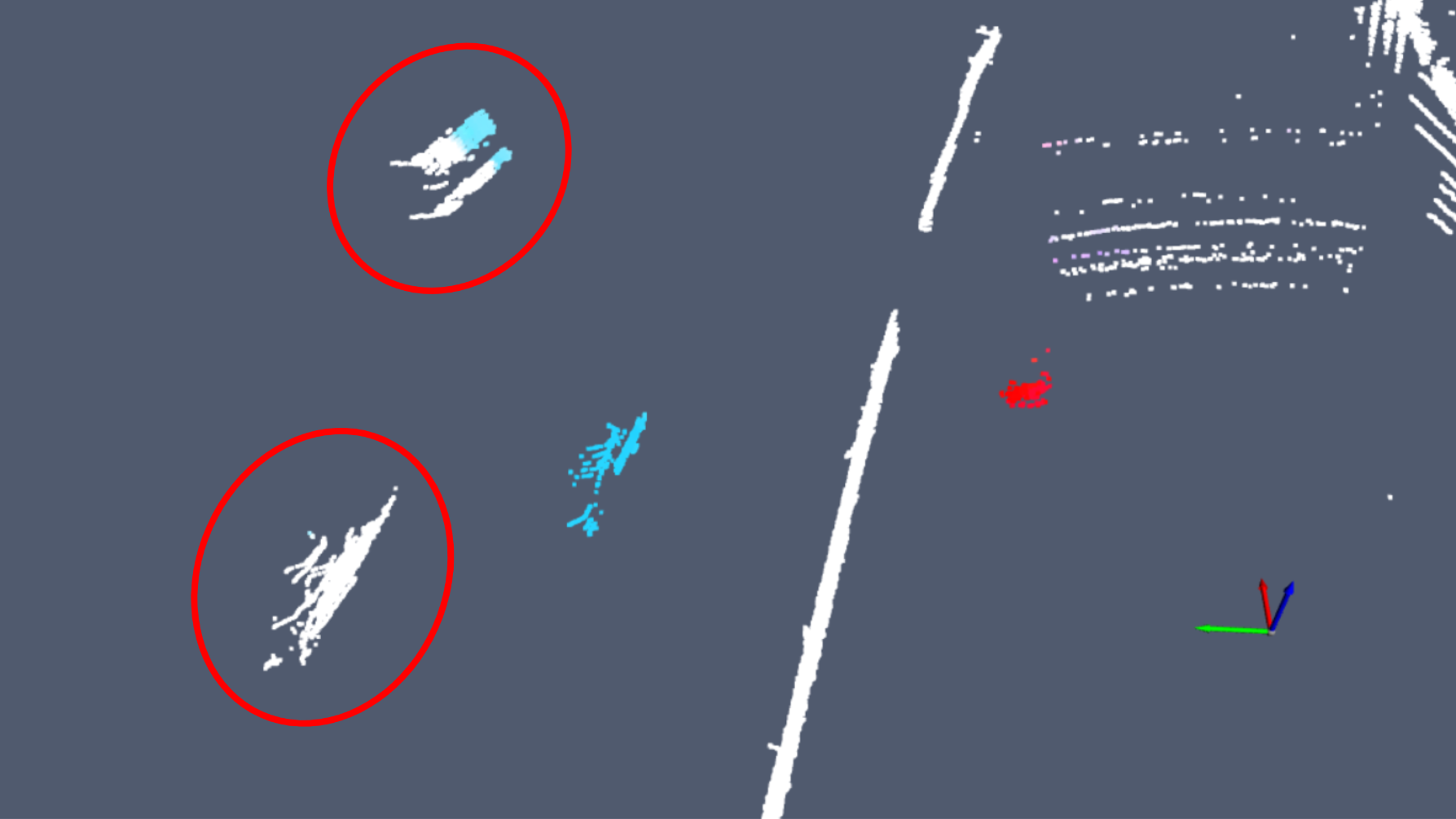}
            \end{subfigure}\hfill
            \begin{subfigure}{0.32\linewidth}
                \includegraphics[width=\linewidth]{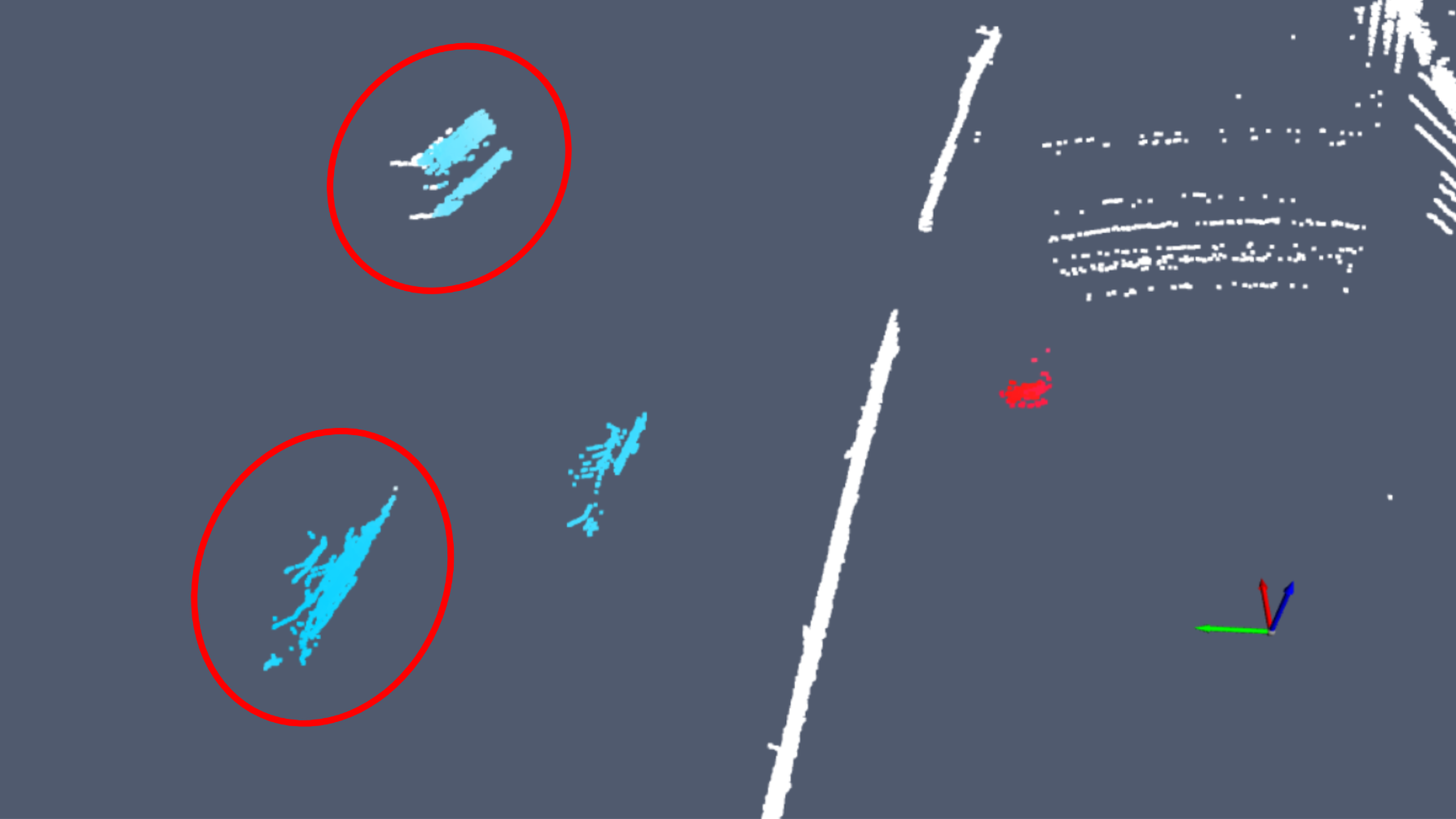}
            \end{subfigure}\hfill
            \begin{subfigure}{0.32\linewidth}
                \includegraphics[width=\linewidth]{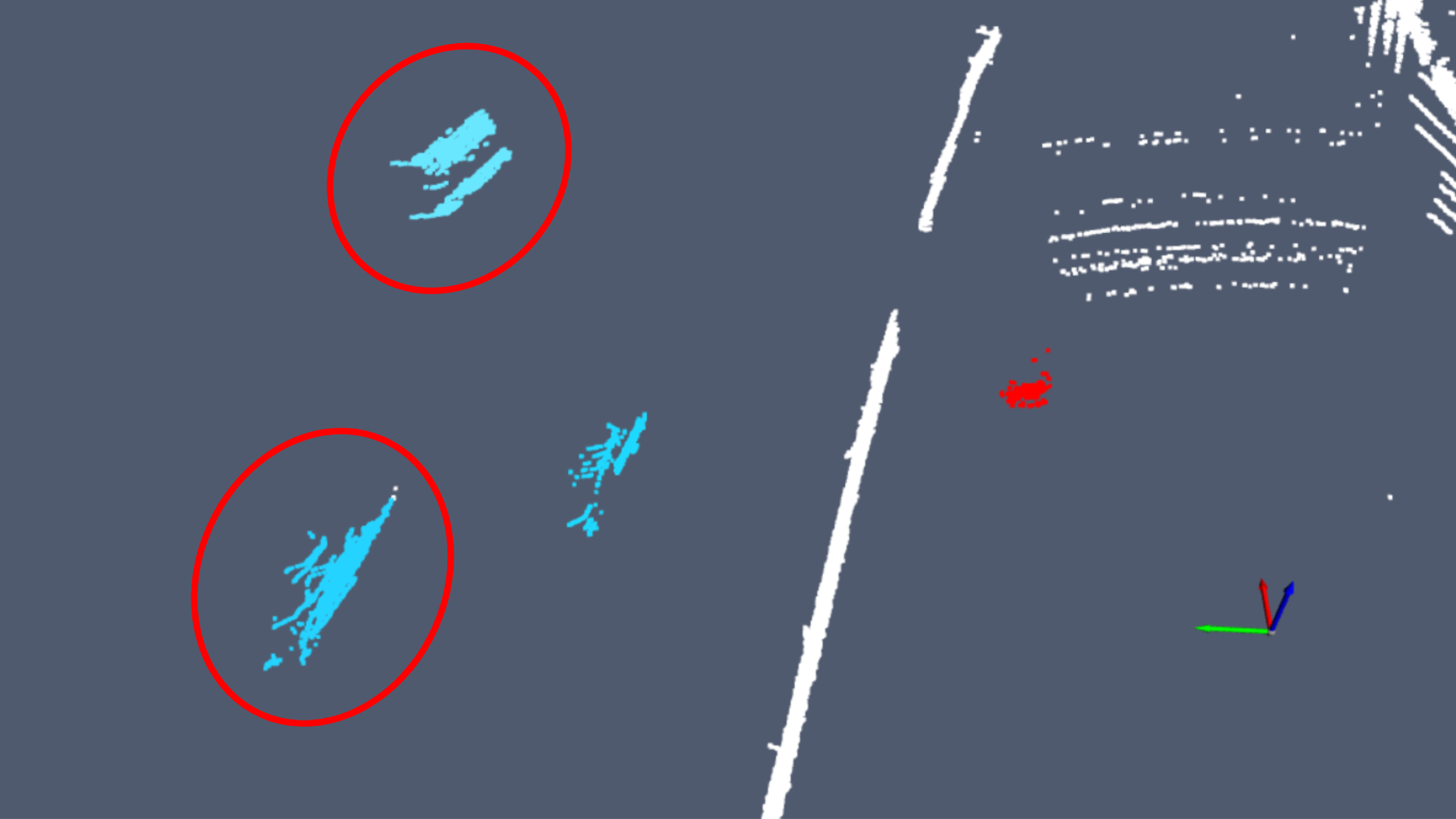}
            \end{subfigure}
        \end{minipage} \\
        \noalign{\vskip 0.4em}
        &
        \begin{minipage}{0.90\linewidth}\centering
            \begin{subfigure}{0.32\linewidth}
                \caption*{\textbf{$\Delta$Flow}}
            \end{subfigure}\hfill
            \begin{subfigure}{0.32\linewidth}
                \caption*{\textbf{$\Delta$Flow (\method)}}
            \end{subfigure}\hfill
            \begin{subfigure}{0.32\linewidth}
                \caption*{\textbf{GT}}
            \end{subfigure}
        \end{minipage}
    \end{tabular}
   \caption{\textbf{AEVAScenes Zero-Shot Results.}
   We compare dataset-specific \(\Delta\)Flow, \(\Delta\)Flow (\method), and ground truth on two representative AEVAScenes examples. In both scenes, \method\ captures the motion of foreground vehicles that the dataset-specific baseline fails to recover, highlighted by the blue vehicles in the red circles. The unified model also produces more accurate motion estimates for the red objects to the right in Scene~I.}
   \label{fig:aeva_qualitative_rebuttal}
\end{figure}

We provide additional qualitative results on AEVAScenes to illustrate zero-shot generalization to a previously unseen LiDAR modality. Figure~\ref{fig:aeva_qualitative_rebuttal} compares dataset-specific \(\Delta\)Flow, \(\Delta\)Flow (\method), and ground truth on two representative scenes. In both examples, \method\ recovers the motion of foreground vehicles that the dataset-specific baseline fails to capture, highlighted by the blue vehicles in the red circles. \method\ also produces more accurate motion estimates on the small foreground object to the right in Scene I.

\begin{figure}[b!]
    \centering
    \setlength{\tabcolsep}{1.5pt}
    \newlength{\failH}
    \setlength{\failH}{0.16\linewidth}

    \begin{tabular}{@{}c@{\hspace{0.4em}}c@{}}
        \figtwosidecaption{Scene I}{} &
        \begin{minipage}{0.90\linewidth}\centering
            \begin{subfigure}{0.32\linewidth}
                \includegraphics[height=\failH,keepaspectratio]{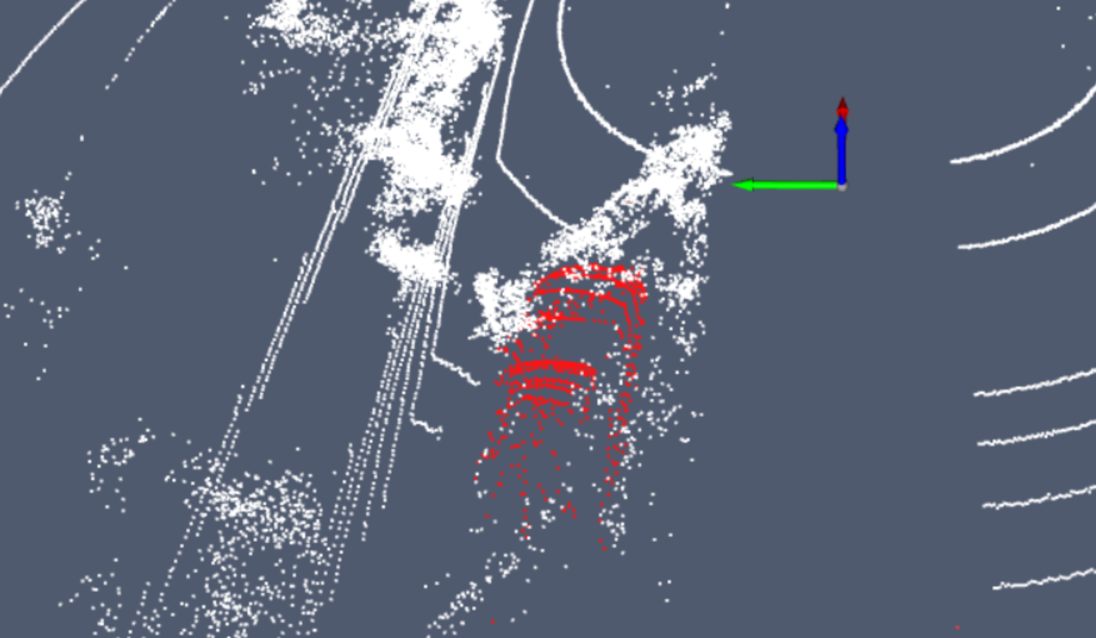}
            \end{subfigure}\hfill
            \begin{subfigure}{0.32\linewidth}
                \includegraphics[height=\failH,keepaspectratio]{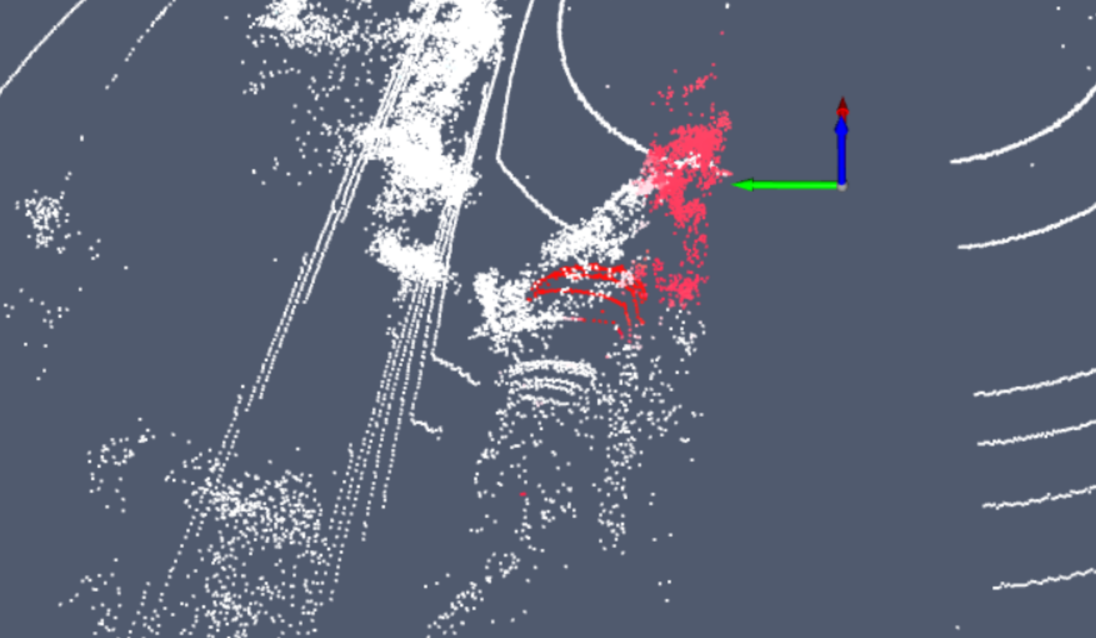}
            \end{subfigure}\hfill
            \begin{subfigure}{0.32\linewidth}
                \includegraphics[height=\failH,keepaspectratio]{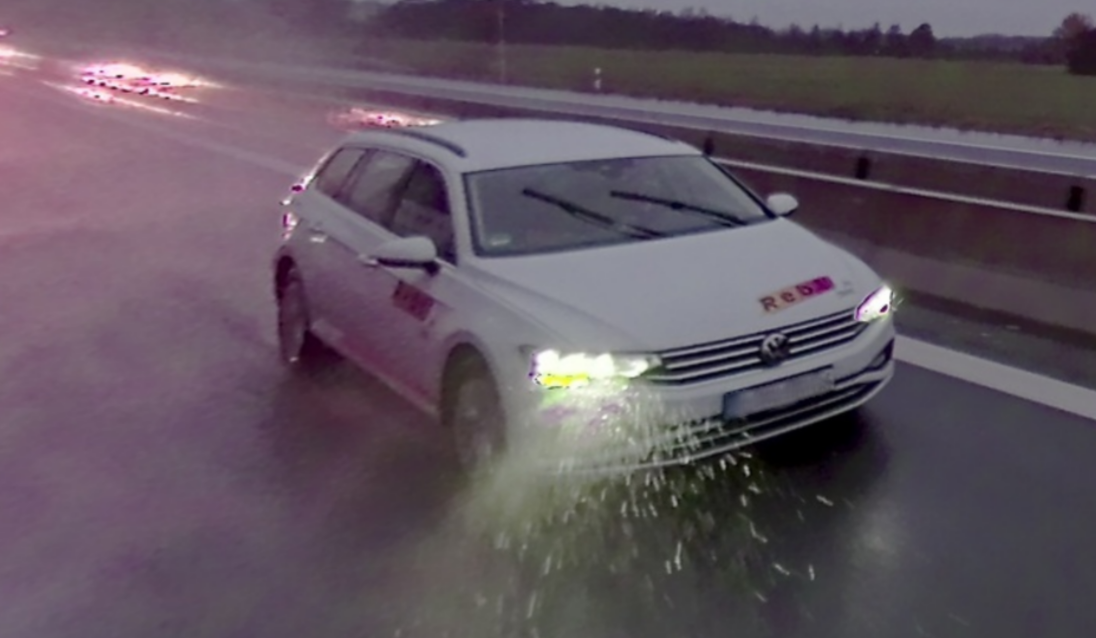}
            \end{subfigure}
        \end{minipage} \\
        \noalign{\vskip 0.4em}
        \figtwosidecaption{Scene II}{} &
        \begin{minipage}{0.90\linewidth}\centering
            \begin{subfigure}{0.32\linewidth}
                \includegraphics[height=\failH,keepaspectratio]{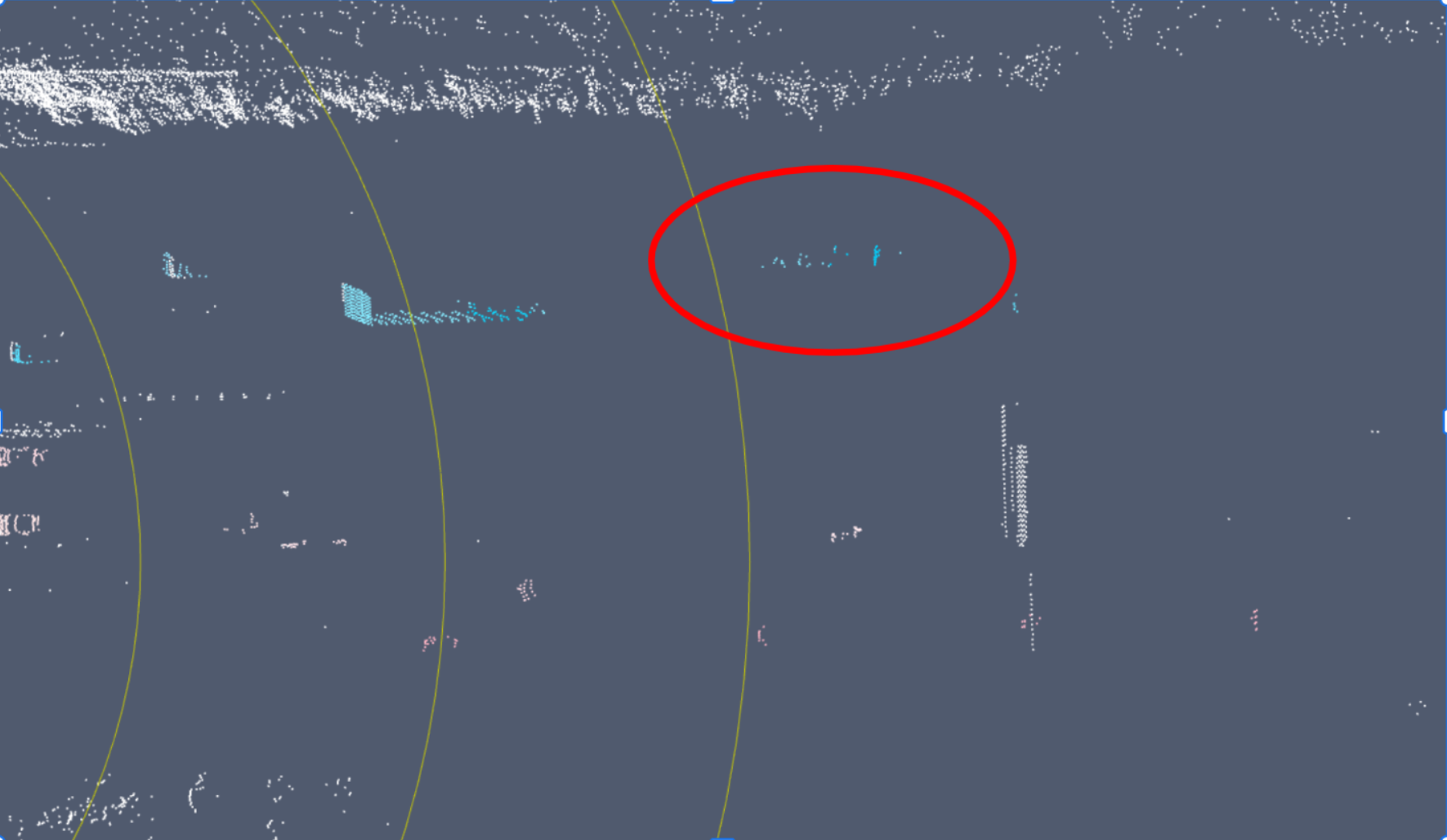}
            \end{subfigure}\hfill
            \begin{subfigure}{0.32\linewidth}
                \includegraphics[height=\failH,keepaspectratio]{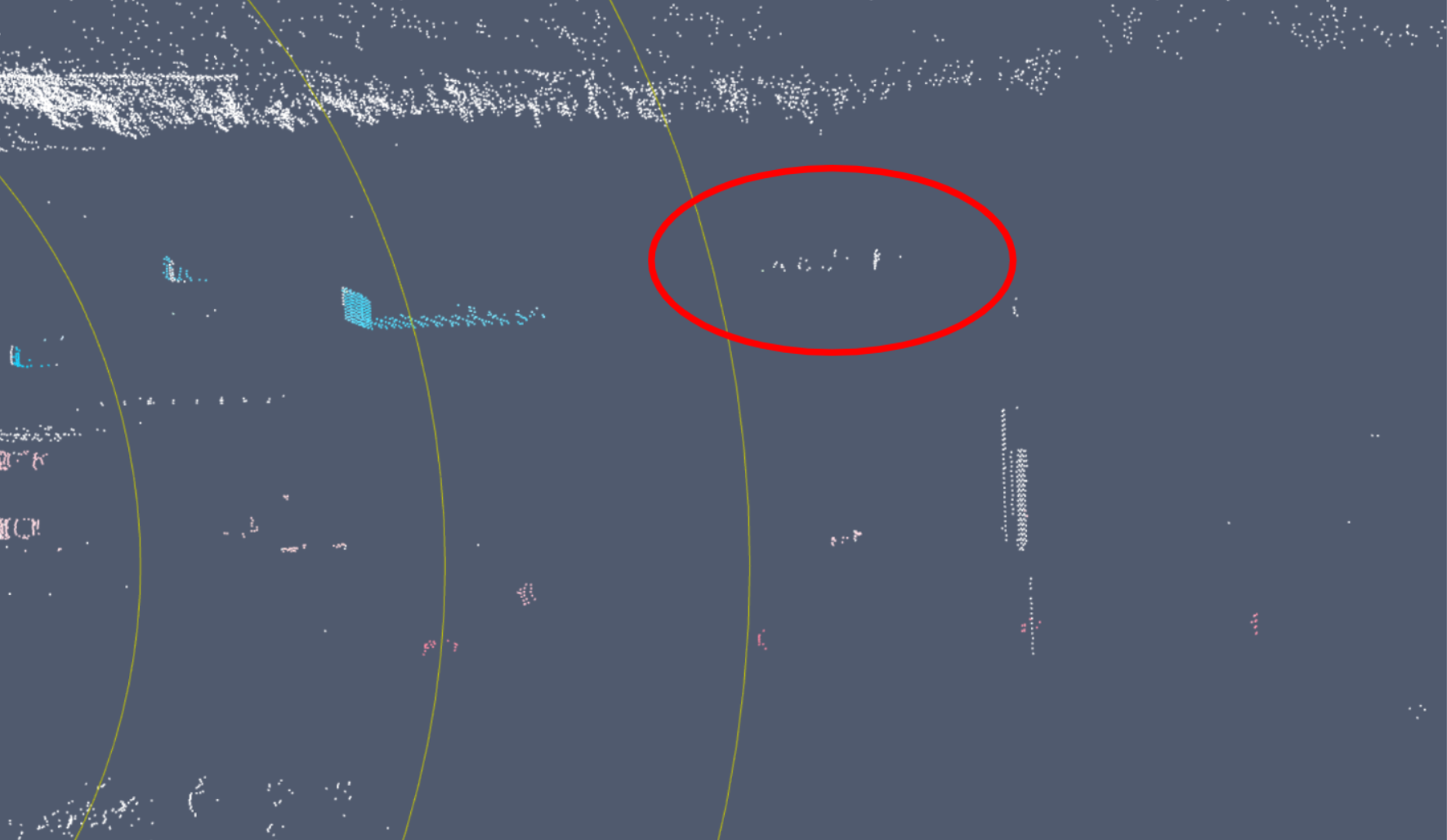}
            \end{subfigure}\hfill
            \begin{subfigure}{0.32\linewidth}
                \includegraphics[height=\failH,keepaspectratio]{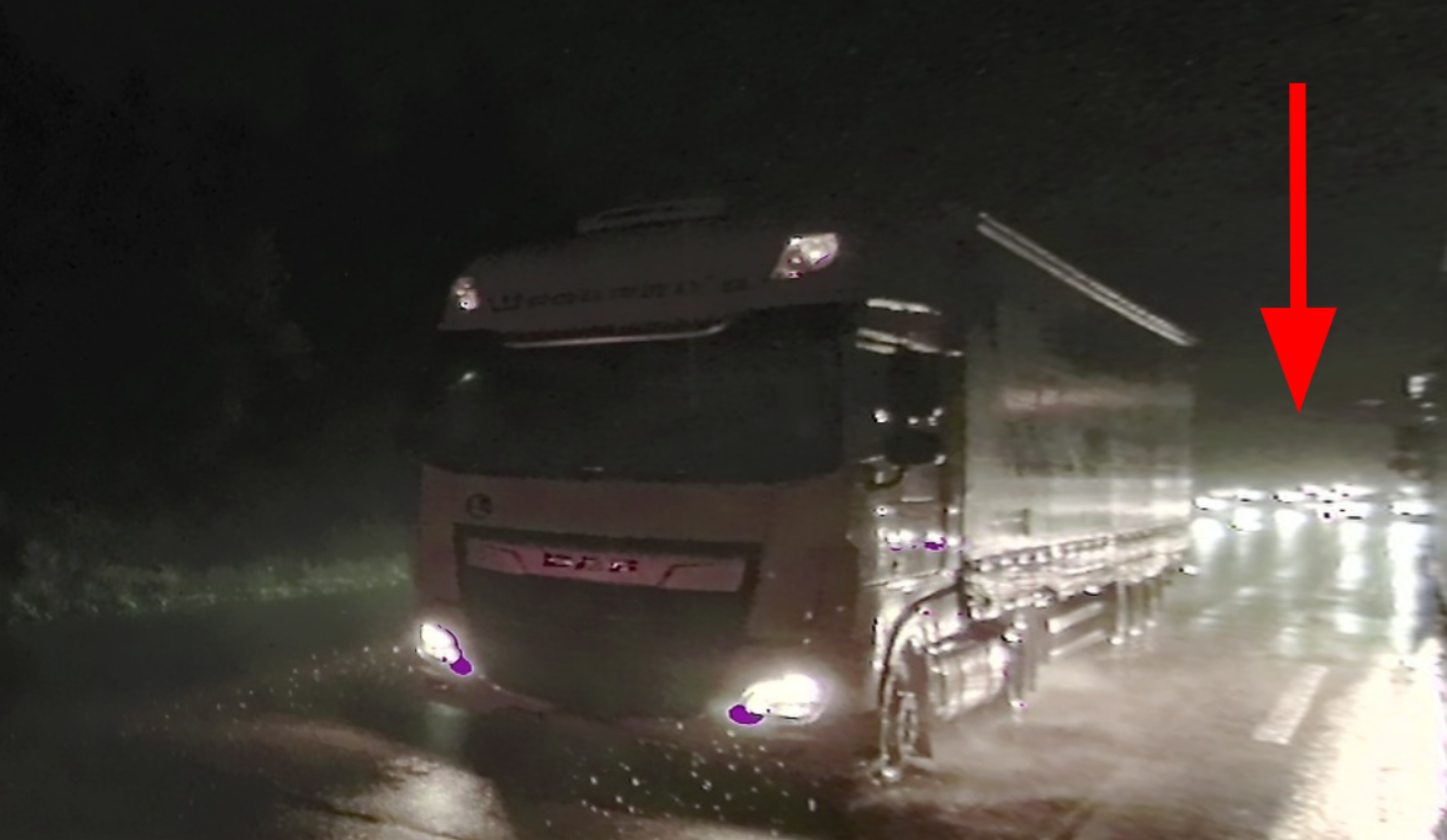}
            \end{subfigure}
        \end{minipage} \\
        \noalign{\vskip 0.4em}
        &
        \begin{minipage}{0.90\linewidth}\centering
            \begin{subfigure}{0.32\linewidth}
                \caption*{\textbf{GT}}
            \end{subfigure}\hfill
            \begin{subfigure}{0.32\linewidth}
                \caption*{\textbf{\method}}
            \end{subfigure}\hfill
            \begin{subfigure}{0.32\linewidth}
                \caption*{\textbf{RGB}}
            \end{subfigure}
        \end{minipage}
    \end{tabular}

    \caption{\textbf{Failure Case Visualizations.}
    We show two failure cases of Flow4D-XL (\method) in TruckScenes. Each example includes the ground-truth flow (left), the model prediction (middle), and the corresponding RGB frame (right).}
    \label{fig:failure_cases}
\end{figure}

\section{Visualizing Failure Cases}
Figure~\ref{fig:failure_cases} illustrates two representative failure modes of \method \ models in TruckScenes. In the first example, adverse weather introduces dense rain streaks that create spurious LiDAR returns, leading the model to produce localized artifacts. In the second example, extreme sparsity at long range causes the model not estimate flow for an object located nearly 100m away.

\end{document}